\documentclass{article} 
\PassOptionsToPackage{table}{xcolor}

\usepackage{graphicx}
\graphicspath{{report/}}
\newcommand{\fitcomparisontable}[2]{\scalebox{#1}{#2}}

\usepackage{fancyhdr}
\usepackage[final]{dlg-agenticmodeling}

\usepackage{microtype}
\usepackage{multicol}
\usepackage{hyperref}
\usepackage{url}
\usepackage{booktabs}
\usepackage{caption}
\usepackage{placeins}
\usepackage{amsmath}
\usepackage{amssymb}
\usepackage[ruled,vlined,linesnumbered]{algorithm2e}
\SetAlgoNlRelativeSize{-1}
\SetAlgoCaptionSeparator{.}
\DontPrintSemicolon
\usepackage{wrapfig}
\makeatletter
\newcommand{\reportfinishwrap}{%
  \par
  \ifvoid\WF@box
    \ifnum\c@WF@wrappedlines>1
      \begingroup
      \@tempcnta=\c@WF@wrappedlines\advance\@tempcnta by -1
      \@tempdima=\baselineskip\multiply\@tempdima by \@tempcnta
      \vspace{\@tempdima}%
      \endgroup
    \fi
  \fi
  \WFclear
}
\makeatother
\usepackage{pifont}
\usepackage{marvosym}
\usepackage{hyperref}
\usepackage{fontawesome5}
\usepackage{enumitem}

\usepackage[most]{tcolorbox}
\definecolor{SoftBlue}{RGB}{232,242,255} 

\newtcolorbox{abstractbox}{
  colback=SoftBlue,
  colframe=SoftBlue, 
  boxrule=0pt,       
  arc=2mm,           
  left=1mm,right=1mm,top=2mm,bottom=2mm
}
\definecolor{TakeawayBlue}{rgb}{0.40,0.80,1.00}
\definecolor{TakeawayBlueBG}{rgb}{0.94,0.97,1.00}
\newtcolorbox{takeaway}{
  colback=TakeawayBlueBG,
  colframe=TakeawayBlue,
  boxrule=0pt,
  leftrule=2pt,
  arc=0pt,
  outer arc=0pt,
  left=8pt,
  right=6pt,
  top=4pt,
  bottom=4pt,
  before skip=5pt,
  after skip=5pt,
}

\usepackage{lineno}
\usepackage{multirow}
\usepackage{booktabs}
\usepackage[T1]{fontenc}
\usepackage{tcolorbox}
\tcbuselibrary{listings,skins,breakable}

\renewcommand{\thefootnote}{\fnsymbol{footnote}}

\definecolor{darkblue}{rgb}{0.0, 0.0, 0.5}
\definecolor{colorbest}{rgb}{0.40, 0.80, 1.00}
\definecolor{colorlowerbest}{rgb}{0.69, 0.89, 0.98}
\hypersetup{colorlinks=true, citecolor=darkblue, linkcolor=darkblue, urlcolor=darkblue}

\newcommand{\ours}{Simplifying and Improving On-Policy Distillation for Few-Step Diffusion}
\newcommand{\shortours}{SMD}
\newcommand{\cb}[1]{\textbf{#1}}
\newcommand{\clb}[1]{\underline{#1}}

\title{\ours}

\author{%
  \textbf{Jiawei Zhang}\textsuperscript{1},
  \textbf{Ziyu Wan}\textsuperscript{1,}\thanks{Corresponding author.}\hspace{0.35em},
  \textbf{Nanye Ma}\textsuperscript{2},
  \textbf{Long Zhao},
  \textbf{Dong Chen}\textsuperscript{1},
  \textbf{Hongyu Liu}\textsuperscript{3} \\
  \textbf{Qifeng Chen}\textsuperscript{3},
  \textbf{Dongdong Chen}\textsuperscript{1} \\
  \textsuperscript{1}Microsoft Research \quad
  \textsuperscript{2}NYU \quad
  \textsuperscript{3}HKUST \\[3pt]
  \href{https://zjwfufu.github.io/SMD-Page/}{\faGlobe\enspace Project Page}\qquad
  \href{https://github.com/zjwfufu/SMD}{\faGithub\enspace Code}\qquad
  \href{https://huggingface.co/collections/zjwfufu/smd}{\faCube\enspace Model}\qquad
  \href{https://huggingface.co/collections/zjwfufu/smd-data}{\faDatabase\enspace Data} \\
}

\newtcblisting{toolfamily}[2][]{%
    enhanced, breakable,
    colback=blue!5!white,
    colframe=blue!55!black,
    colbacktitle=blue!55!black,
    coltitle=white,
    fonttitle=\bfseries,
    title={#2},
    sharp corners=south, arc=3pt,
    boxrule=0.5pt,
    left=8pt, right=8pt, top=4pt, bottom=4pt,
    listing only,
    listing options={
        basicstyle=\ttfamily\footnotesize,
        keywordstyle=\bfseries\color{blue!55!black},
        showstringspaces=false,
        breaklines=true,
        breakatwhitespace=true,
        columns=fullflexible,
        keepspaces=true,
        morekeywords={true,false,null},
    },
    #1
}

\makeatletter
\newif\ifreportappendix
\let\report@addtocontents\addtocontents
\renewcommand{\addtocontents}[2]{%
  \def\report@requestedstream{#1}%
  \def\report@mainstream{toc}%
  \ifreportappendix
    \ifx\report@requestedstream\report@mainstream
      \report@addtocontents{atoc}{#2}%
    \else
      \report@addtocontents{#1}{#2}%
    \fi
  \else
    \report@addtocontents{#1}{#2}%
  \fi
}
\newcommand{\appendixtableofcontents}{%
  \section*{Appendix Contents}%
  \@starttoc{atoc}%
}
\makeatother

\begin{document}

\renewcommand{\thefootnote}{\arabic{footnote}}

\ifcolmsubmission
\linenumbers
\fi

\setcounter{footnote}{1}
\maketitle
\vspace{-18pt}
\raggedbottom
\makeatletter
\setlength{\@fptop}{0pt}
\setlength{\@fpsep}{12pt}
\setlength{\@fpbot}{0pt plus 1fil}
\makeatother

\begin{abstractbox}
\centerline{\normalsize\bfseries Abstract}
\vspace{0.25ex}
{
\noindent
Few-step distillation reduces the sampling cost of diffusion and flow models,
but existing approaches often require auxiliary score models or specialized
architectures. We show that simple on-policy velocity regression already
provides a strong foundation for 4-NFE generation. To represent teacher
velocity variation within each coarse transition, we introduce \textit{Pivot},
a training-only residual proxy that separates local corrections from the
student's interval-mean prediction. We further repel the student from weaker
teacher predictions, providing a lightweight mechanism for avoiding
undesirable modes. The resulting framework, \textit{Simplified Matching
Distillation (SMD)}, remains entirely regression-based, requires neither
online score estimation nor adversarial training, and leaves the deployed
generator unchanged. We evaluate SMD on image generation and editing, 3D
generation, video synthesis, and joint audio--video generation, scaling to
33 billion parameters. At 4-NFE, SMD outperforms strong few-step baselines
and surpasses multi-step teachers on several benchmarks. It also supports
cross-model, multi-teacher, and causal-video distillation.
\par}
\end{abstractbox}

\begingroup
\setlength{\intextsep}{2pt}
\begin{figure}[!h]
\centering
\captionsetup{font=small,skip=2pt}
\includegraphics[width=1.0\textwidth]{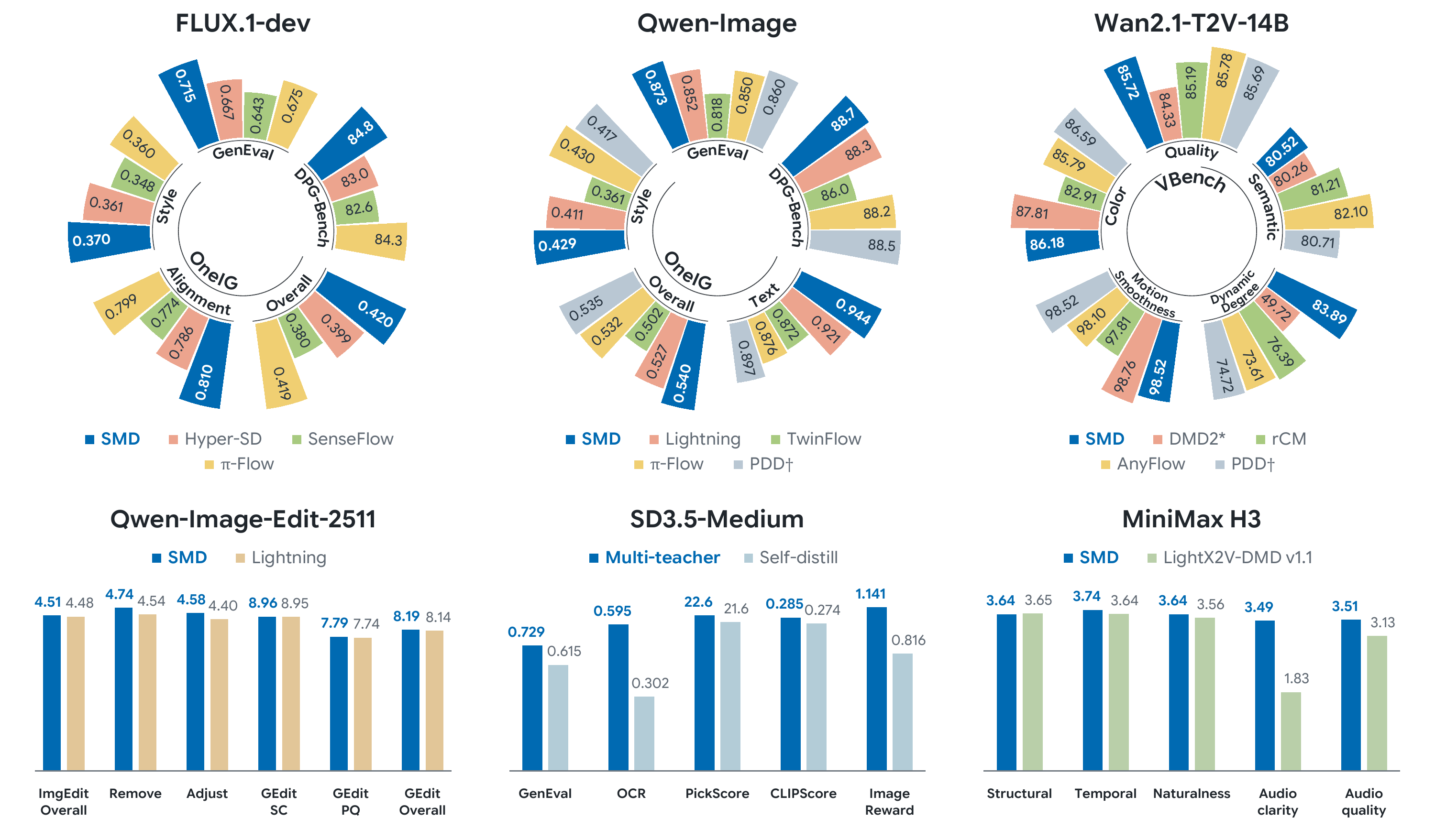}
\caption{\textbf{Simplicity scales.}
SMD at 4-NFE across image, video, editing, multi-teacher, and audio--video generation.
Radial scales are truncated per metric. Column scales start at zero. All metrics are higher-is-better.
SC/PQ denote semantic consistency/perceptual quality.
$^{*}$Our DMD2 reproduction. $^{\dagger}$PDD results~\cite{pdd}.}
\label{fig:benchmark_overview}
\end{figure}
\endgroup

\clearpage
\begingroup
\newlength{\galleryimagewidth}
\setlength{\galleryimagewidth}{0.92\textwidth}
\makeatletter
\setlength{\@fptop}{0pt}
\setlength{\@fpsep}{12pt}
\makeatother

\begin{figure}[p]
\centering
\includegraphics[width=\galleryimagewidth]{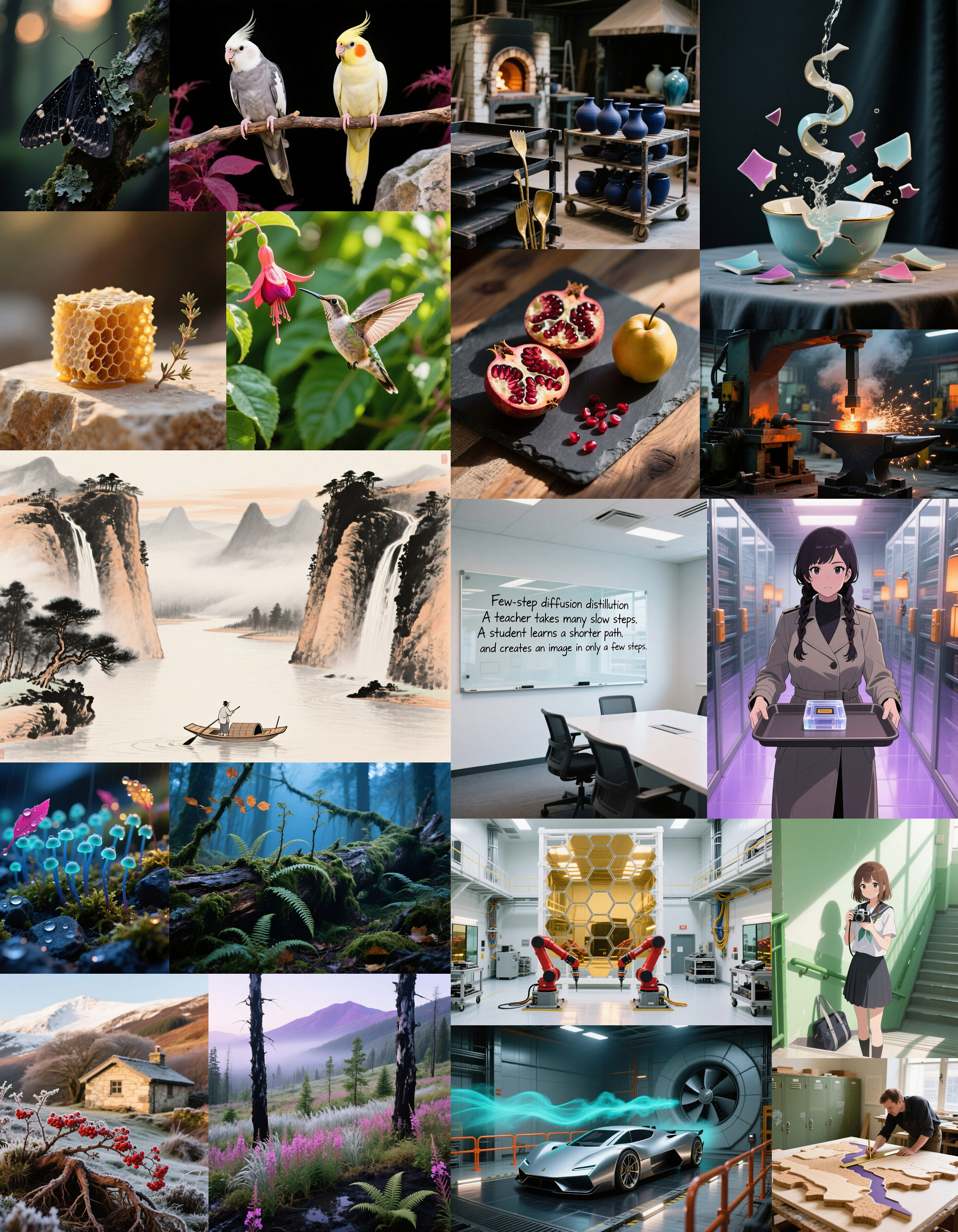}
\caption{\textbf{Few-step image generation with Qwen-Image.}
The SMD-distilled 4-NFE student generates images with varied compositions,
embedded text, and visual styles.}
\label{fig:gallery_qwen_image}
\end{figure}
\clearpage

\begin{figure}[p]
\centering
\includegraphics[width=\galleryimagewidth]{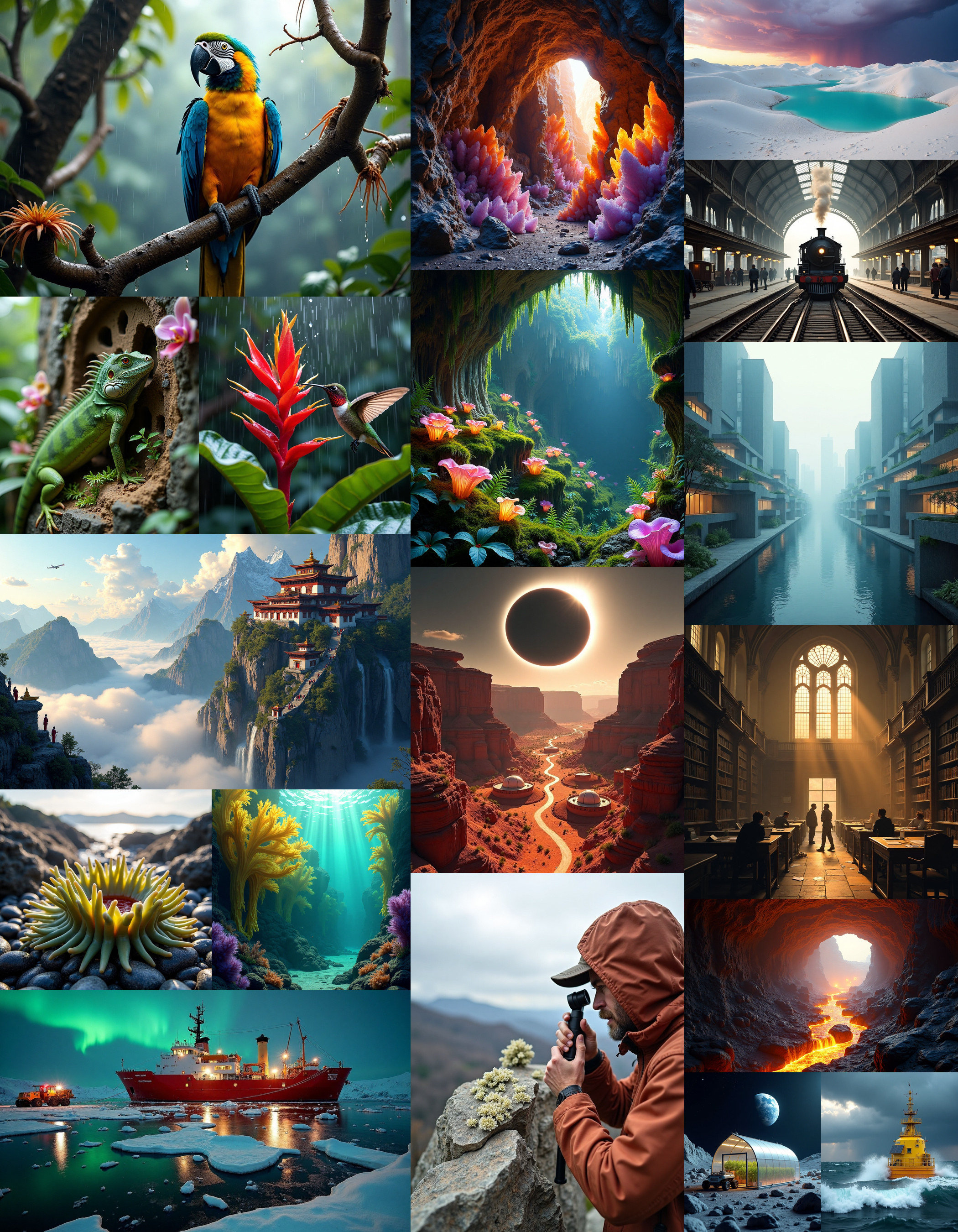}
\caption{\textbf{Few-step image generation with FLUX.1-dev.}
The SMD-distilled 4-NFE student generates animals, plants, architecture,
landscapes, and fantasy scenes.}
\label{fig:gallery_flux1dev}
\end{figure}
\clearpage

\begin{figure}[p]
\centering
\includegraphics[width=\galleryimagewidth]{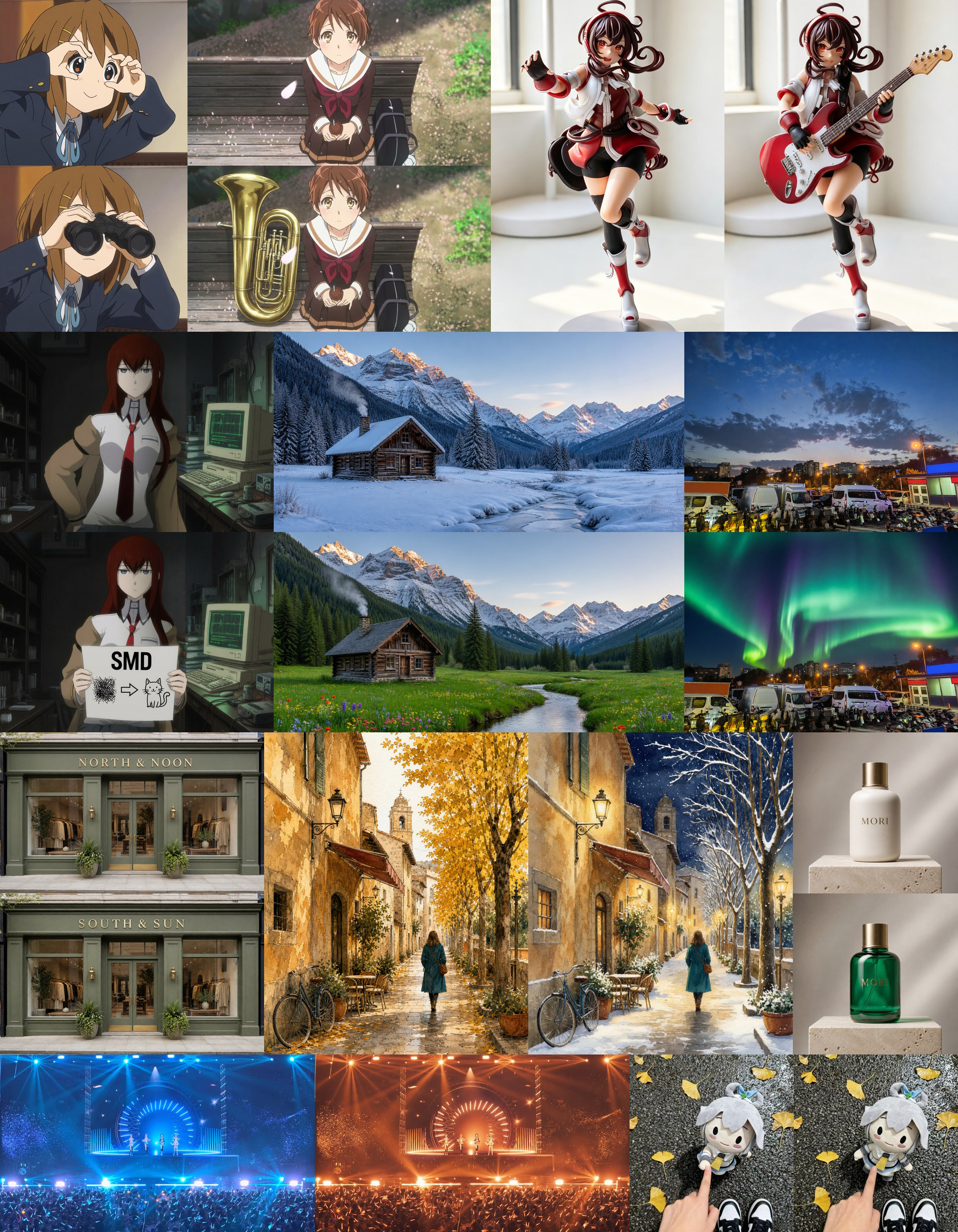}
\caption{\textbf{Few-step image editing with Qwen-Image-Edit-2511.}
At 4-NFE, SMD supports object insertion and replacement, text edits,
and changes in season, lighting, and appearance.}
\label{fig:gallery_qwen_image_edit}
\end{figure}
\clearpage

\begin{figure}[p]
\begin{minipage}[t][0.96\textheight][s]{\textwidth}
\centering

\includegraphics[width=\galleryimagewidth]{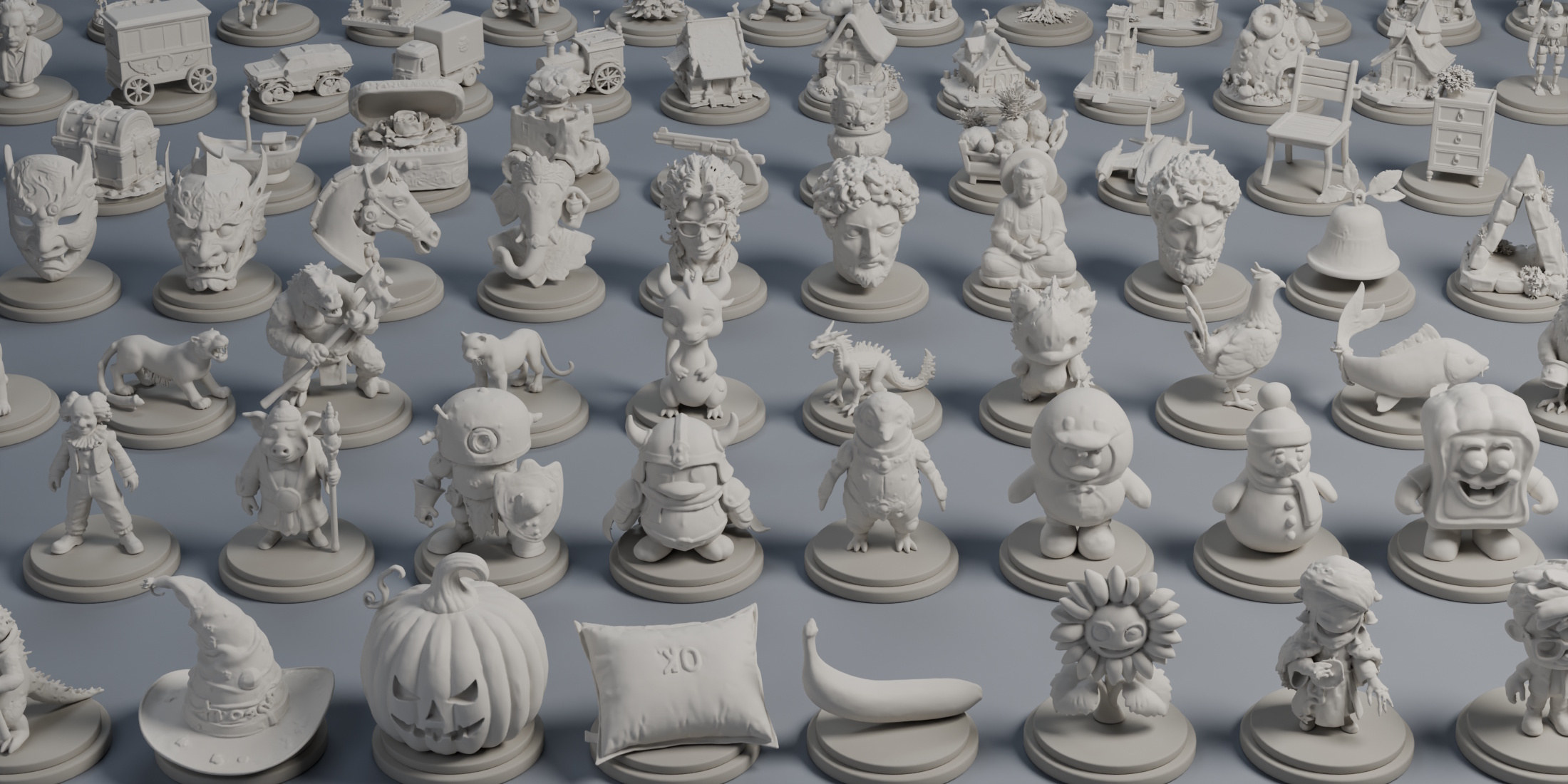}
\caption{\textbf{Few-step shape generation with Hunyuan3D-2.1.}
SMD reduces the shape-generation stage to 4-NFE.
The untextured renders highlight the generated geometry.}
\label{fig:gallery_hunyuan3d}

\vfill

\includegraphics[width=\galleryimagewidth]{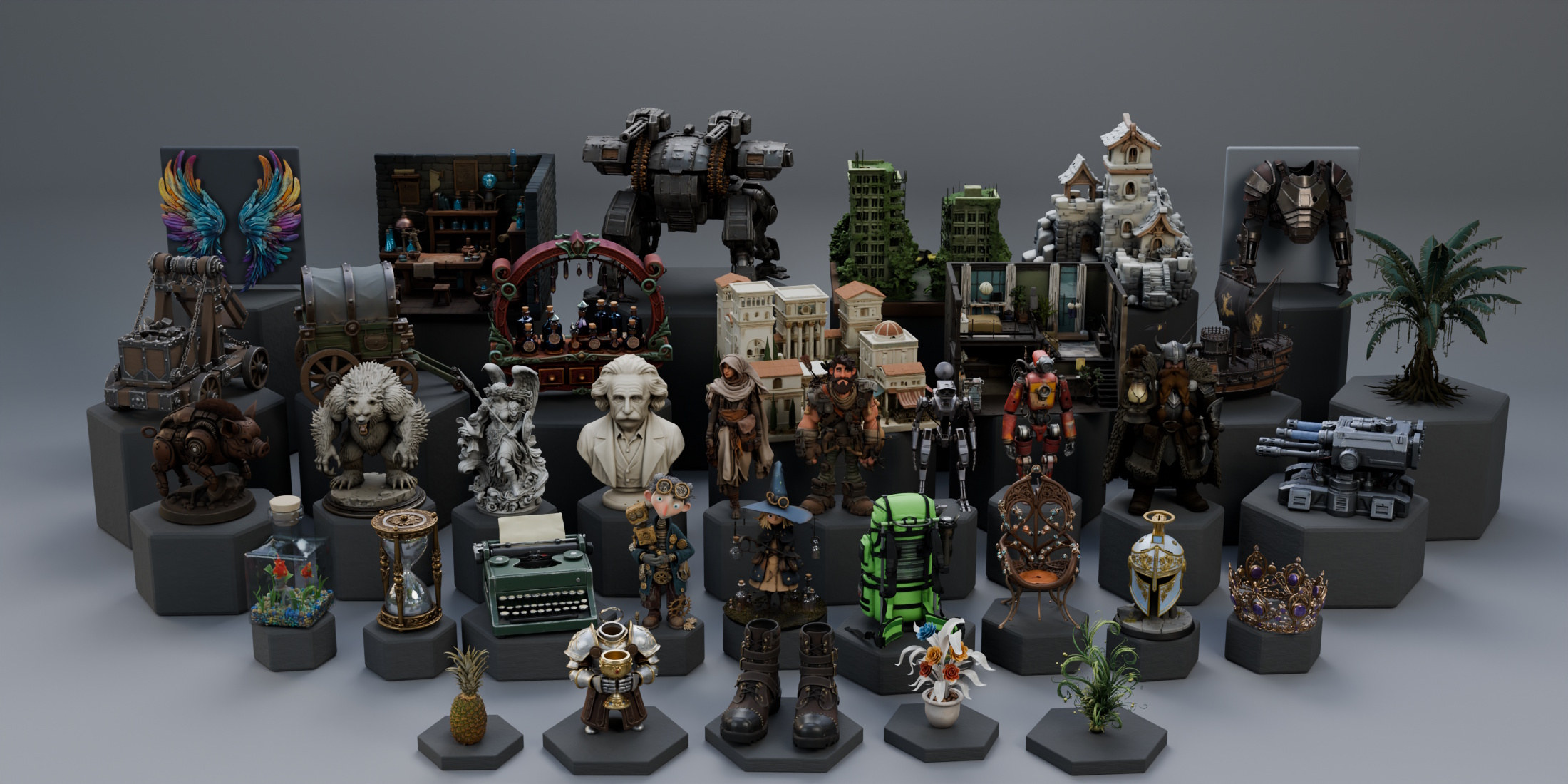}
\caption{\textbf{Few-step shape and texture generation with TRELLIS.2.}
SMD distills the high-resolution shape and texture stages to
4-NFE each, producing textured 3D assets.}
\label{fig:gallery_trellis2}

\end{minipage}
\end{figure}
\clearpage

\begin{figure}[p]
\centering
\includegraphics[width=\galleryimagewidth]{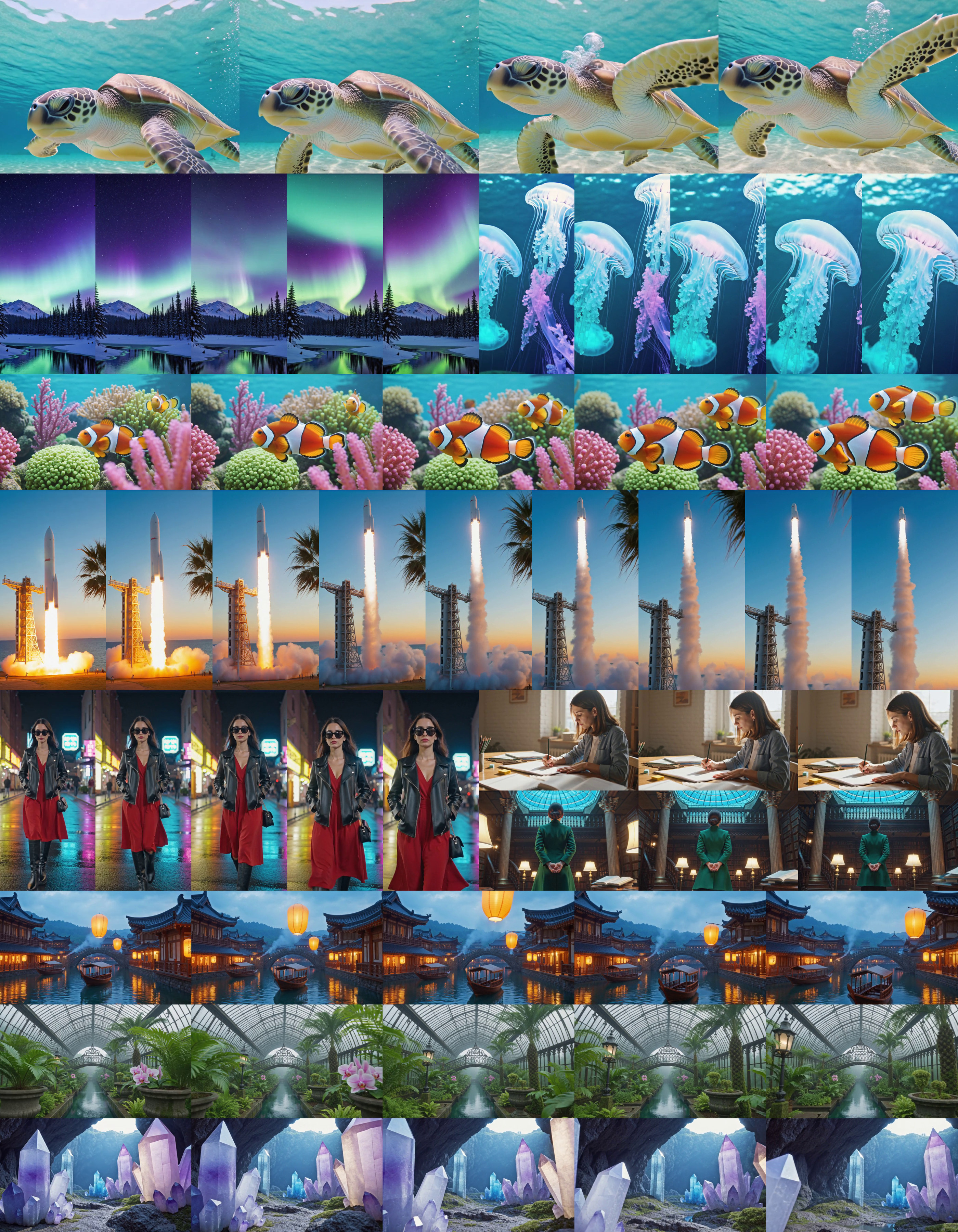}
\caption{\textbf{Few-step video generation with Wan2.1-T2V-14B.}
A 4-NFE student distilled by SMD generates animals, landscapes, human
activities, and a rocket launch.
Frames within each sequence are ordered chronologically from left to right.}
\label{fig:gallery_wan14}
\end{figure}
\clearpage

\begin{figure}[p]
\centering
\includegraphics[width=\galleryimagewidth]{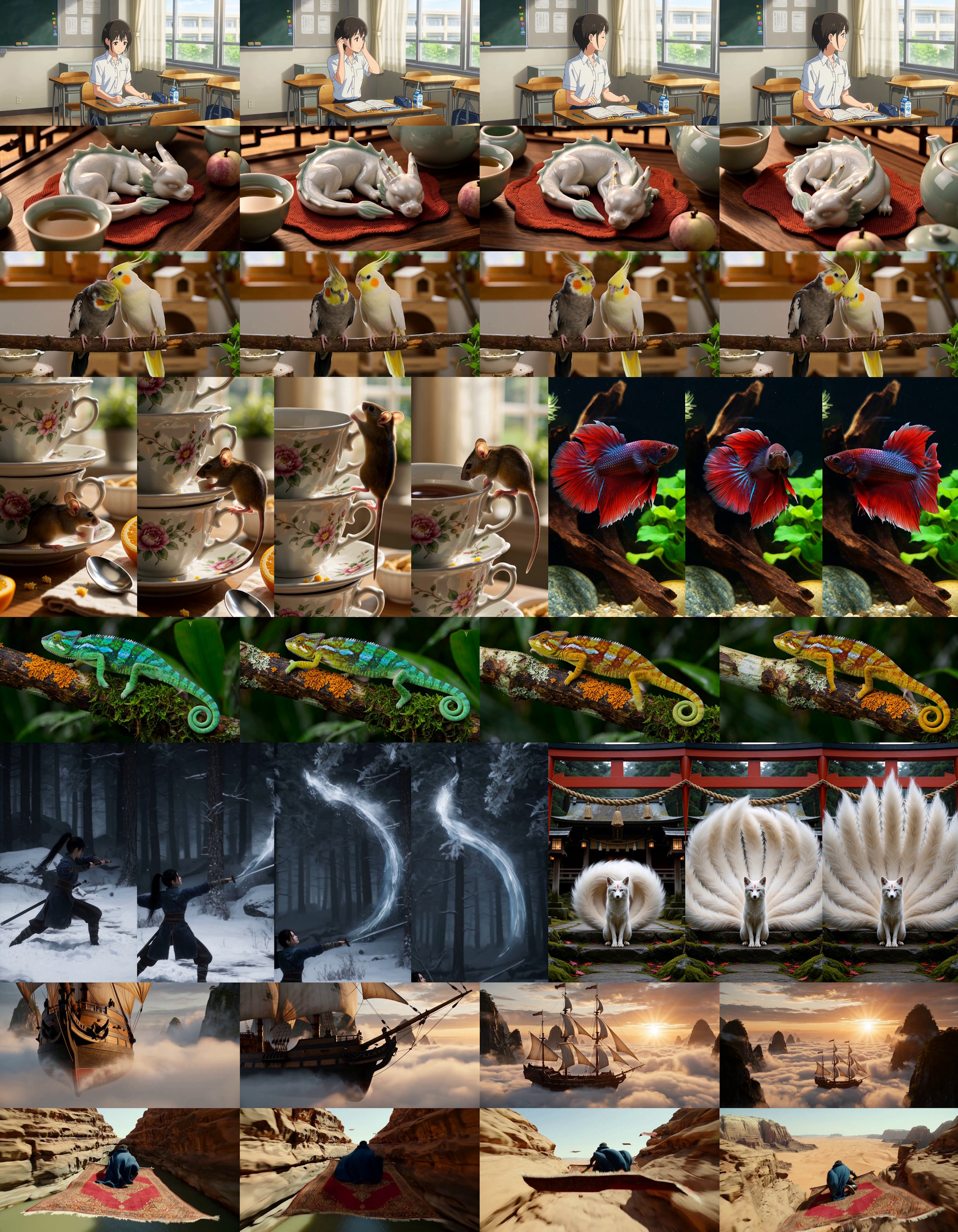}
\caption{\textbf{Few-step joint audio--video generation with MiniMax H3.}
The SMD-distilled student performs text-to-audio--video generation
at 8-NFE.
The gallery displays selected video-frame sequences covering character
animation, animals, and fantasy scenes.}
\label{fig:gallery_minimax_h3}
\end{figure}
\clearpage

\endgroup

\renewcommand{\contentsname}{Main Contents}
\tableofcontents
\label{sec:contents}
\newpage

\section{Introduction}
\label{sec:intro}

 \newcommand{\summarybox}[1]{%
   \begin{tcolorbox}[
     colback=gray!10,
     colframe=teal!60!black,
     arc=5pt,
     boxrule=0.8pt,
     boxsep=2.5pt,
     left=8pt, right=8pt,
     top=0.5pt, bottom=0.5pt,
     enhanced,
     drop shadow=gray!50
   ]
     #1
   \end{tcolorbox}%
 }



Diffusion and flow models~\cite{ddpm,scoresde,flowmatching,rectifiedflow} have become a foundation for image, video, and 3D generation~\cite{stablediffusion,qwenimage,wan,hunyuan3d,trellis2}.
Their exceptional generation quality, however, comes with a heavy inference burden, as the inherent denoising sampling typically requires tens to hundreds of sequential function evaluations. This high computational overhead thus creates a significant bottleneck for real-world deployment and interactive applications, making inference efficiency an urgent priority.
Few-step distillation~\cite{dmd,progressivedistill,consistencymodels,piflow,pdd} is the way to address this challenge. By compressing a pretrained teacher model, the denoising process could be significantly reduced to a few steps, enabling rapid generation without sacrificing, and sometimes even enhancing the original generation quality~\cite{dmd2}.
Consequently, few-step distillation now has become an important and indispensable stage to post-train a frontier diffusion foundation model.


\begingroup\looseness=-1
Among existing distillation paradigms, on-policy distillation, which trains a few-step student on states generated by the student itself, has emerged as the dominant approach for accelerating diffusion models.
Distribution Matching Distillation (DMD)~\cite{dmd} is no doubt standing as a pillar of this paradigm and has inspired a series of follow-up works~\cite{dmd2,senseflow,causvid,cdm}. By re-noising on-policy student generated samples, DMD approximates a reverse-KL gradient using the score difference between the teacher and the student, where the student score is dynamically tracked by an online fake model~\cite{dmd}. Despite its remarkable progress, DMD and its variants often suffer from low-diversity, unstable training dynamics and over-saturated output. To mitigate these issues, recent trajectory-based methods such as $\pi$-Flow~\cite{piflow} and PDD~\cite{pdd} simplify training by directly supervising student rollouts with teacher predictions within each sampling interval. Although this removes the need for online score estimation, the few-step generator must still represent the dense velocity field within each interval, relying on either a learned policy~\cite{piflow} or a parallel decoder~\cite{pdd}, which adds architectural components.

\endgroup


We show that simple on-policy velocity matching is already a strong baseline for few-step diffusion distillation. Our approach requires no auxiliary fake score model, architectural modification, or elaborate policy design. Instead, it simply matches the student velocity predictions to the teacher velocities queried at the same states, while keeping the student velocity constant within each sampling interval. This baseline produces strong 4-NFE results, with photorealistic detail and varied outputs.
\begin{takeaway}
\textbf{Finding 1.}
Direct velocity matching between student and teacher prediction is already a strong baseline for few-step diffusion distillation.
\end{takeaway}
Building on this finding, we further apply this simple baseline to a variety of diffusion teachers~\citep{flux1,qwenimage,wan}. However, when distilling video generation models, we observe that training dynamics can occasionally suffer from instability, as shown in Figure~\ref{fig:ablation_simple}. Through the curvature analysis on the velocity field, we observe actually within a coarse integration step, the dense teacher velocities can vary substantially~\citep{lee2023minimizing}, especially for modeling complex and high-dimensional modality like video, hence exposing a single student prediction to varying targets over an entire interval to calculate the imitation loss will make the optimization conflict and jitter. To address this issue, we introduce \textit{Pivot proxy}, a lightweight, training-only residual network that captures local velocity variations via a proxy path. By decoupling local velocity modeling from the overall student prediction, \textit{Pivot proxy} stabilizes training while incurring no inference overhead. To further correct the student errors, we exploit weaker predictions from the same teacher as informative negative references and introduce a repulsive regularization term, which steers the student away from these inferior predictions without compromising faithful imitation of the teacher target. We term the resulting framework \textit{Simplified Matching Distillation} (\shortours{}). True to its name, \shortours{} remains entirely regression-based. It requires no modification to the architecture or inference pipeline of the few-step generator, and involves neither score estimation nor adversarial objectives.
\begin{takeaway}
\textbf{Finding 2.}
Pivot proxy path and repulsive term introduced by SMD further improve the stability and quality of few-step diffusion distillation.
\end{takeaway}
Such simplicity makes SMD both practical and scalable. Trained purely on-policy without any target data, SMD applies to image generation and editing, 3D generation, video synthesis, and joint audio--video generation, scaling to models with up to 33B parameters. With only 4-NFE, SMD distilled FLUX.1-dev~\cite{flux1} and Qwen-Image~\cite{qwenimage} models outperform all compared few-step baselines on GenEval and DPG-Bench (Figure~\ref{fig:benchmark_overview}), with the FLUX.1-dev~\cite{flux1} student even surpassing its 50-NFE teacher on both. On Wan2.1-T2V~\cite{wan}, SMD remains competitive with strong 4-NFE baselines at both 1.3B and 14B scales. Beyond acceleration, the framework supports cross-architecture, causal video generation distillation, and multi-teacher distillation, where a single 4-NFE student outperforms its 28-NFE base model on all six reported metrics. These results position SMD as a powerful and versatile tool for post-training diffusion foundation models.
\begin{takeaway}
\textbf{Finding 3.}
SMD supports post-training diffusion models across modalities and scales.
\end{takeaway}

\section{Preliminaries}
\label{sec:preliminaries}

\paragraph{Diffusion and flow matching models.}
Modern diffusion and flow models can be formulated in terms of a learned velocity field that transports a simple noise distribution to the data distribution~\cite{flowmatching,rectifiedflow}.
Let $x_0\sim p_{\mathrm{data}}(x\mid c)$ be a data sample under condition $c$, and let $x_1\sim\mathcal N(0,I)$ be Gaussian noise.
A common linear probability path is
\begin{equation}
    x_t=(1-t)x_0+t x_1,
    \qquad t\in[0,1],
    \label{eq:flow_path}
\end{equation}
where $t=0$ is the data endpoint and $t=1$ is the noise endpoint.
The conditional velocity along this path is $\dot x_t=x_1-x_0$.
Flow matching trains a velocity model to predict this quantity from the interpolated state. We write its prediction as $v_T(x_t,t,c)$ when it serves as the teacher,
\begin{equation}
    \mathcal L_{\mathrm{FM}}
    =
    \mathbb E_{x_0,x_1,t}
    \left[
        \left\|
            v_T(x_t,t,c)-(x_1-x_0)
        \right\|_2^2
    \right].
    \label{eq:flow_matching}
\end{equation}
At the population optimum, the learned field approaches the marginal velocity
$v^*(x,t,c)=\mathbb E[x_1-x_0\mid x_t=x,c]$.
This velocity defines the generative ODE
\begin{equation}
    \frac{d x_t}{dt}=v_T(x_t,t,c).
    \label{eq:generative_ode}
\end{equation}
Generation starts from $x_1\sim\mathcal N(0,I)$ and integrates the ODE from $t=1$ to $t=0$.
Each model evaluation provides only the local velocity at the current state and time. A complete sample is obtained by integrating these predictions over the full time interval.
The state $x_t$ may be an image or a learned latent, and $c$ may contain text, images, or other conditioning signals~\cite{stablediffusion,dit}.
Some pretrained models natively predict noise or a clean sample rather than velocity.
We convert these outputs to velocity under the model's original parameterization.
Throughout this report, $v_T$ is frozen during distillation, $v_\theta$ denotes the student, and $\phi$ denotes the Pivot parameters.

\paragraph{Conditional guidance.}
Classifier-free guidance combines predictions with and without the condition~\cite{cfg}.
In velocity form, a guidance scale $w$ gives
\begin{equation}
    v_T^{(w)}(x_t,t,c)
    =
    v_T(x_t,t,\varnothing)
    +
    w\left[
        v_T(x_t,t,c)
        -
        v_T(x_t,t,\varnothing)
    \right].
    \label{eq:cfg_velocity}
\end{equation}
The conditional prediction is recovered at $w=1$, while larger values generally strengthen conditioning at a possible cost to diversity.
The weaker prediction need not come from an unconditional branch.
Recent guidance methods may instead use a weaker model, a perturbed forward pass, or a reduced set of conditions.
In each case, guidance modifies the local velocity used by the sampler without changing the underlying integration procedure.

\paragraph{Sampling schedules.}
Although the learned velocity field is continuous in time, inference evaluates it on a finite decreasing grid
\begin{equation}
    1=t_0>t_1>\cdots>t_N=0.
\end{equation}
The placement of these time points determines which regions of the transport receive more model evaluations.
Some pretrained models additionally warp the nominal time before passing it to the network.
One common rational shift is
\begin{equation}
    S_\gamma(t)
    =
    \frac{\gamma t}{1+(\gamma-1)t},
    \qquad \gamma>0.
    \label{eq:time_shift}
\end{equation}
We absorb this transformation into the sampling grid and use $t$ for the resulting model time below.
Given $\Delta_i=t_i-t_{i+1}>0$, a first-order update takes the form
\begin{equation}
    x_{i+1}
    =
    x_i-\Delta_i v_T(x_i,t_i,c).
    \label{eq:flow_update}
\end{equation}
Higher-order solvers evaluate the field at additional states within an interval to reduce discretization error~\cite{ddim,edm}.
Each network call is counted as one function evaluation, or NFE, and constitutes the main inference cost.
Standard samplers use many evaluations to follow the learned field closely.
Few-step sampling retains only a small number of coarse transitions, making the accuracy of each transition increasingly important.

\section{Method}
\label{sec:method}

SMD combines a simple on-policy
velocity-regression baseline with two training-only refinements.
For each coarse transition of the student rollout, the baseline queries a
frozen teacher at intermediate states along the student's current path and
matches a single coarse student velocity to the resulting local predictions.
Pivot augments this transition with an endpoint-preserving proxy path whose
local velocity can vary within the interval, while reference repulsion
complements the target teacher prediction with a weaker prediction from the
same teacher. Together, these components define SMD. Both refinements are
removed at inference, preserving the student architecture and sampler.

Throughout this section, $v_T$ denotes the frozen target teacher prediction,
including its native guidance rule, and $v_\theta$ denotes the student.
Conditioning is implicit unless stated otherwise.

\subsection{A Simple On-Policy Baseline}
\label{sec:method-baseline}

\begin{figure}[!t]
\centering
\begin{minipage}[t]{0.44\textwidth}
\vspace{0pt}
\centering
\includegraphics[width=\linewidth]{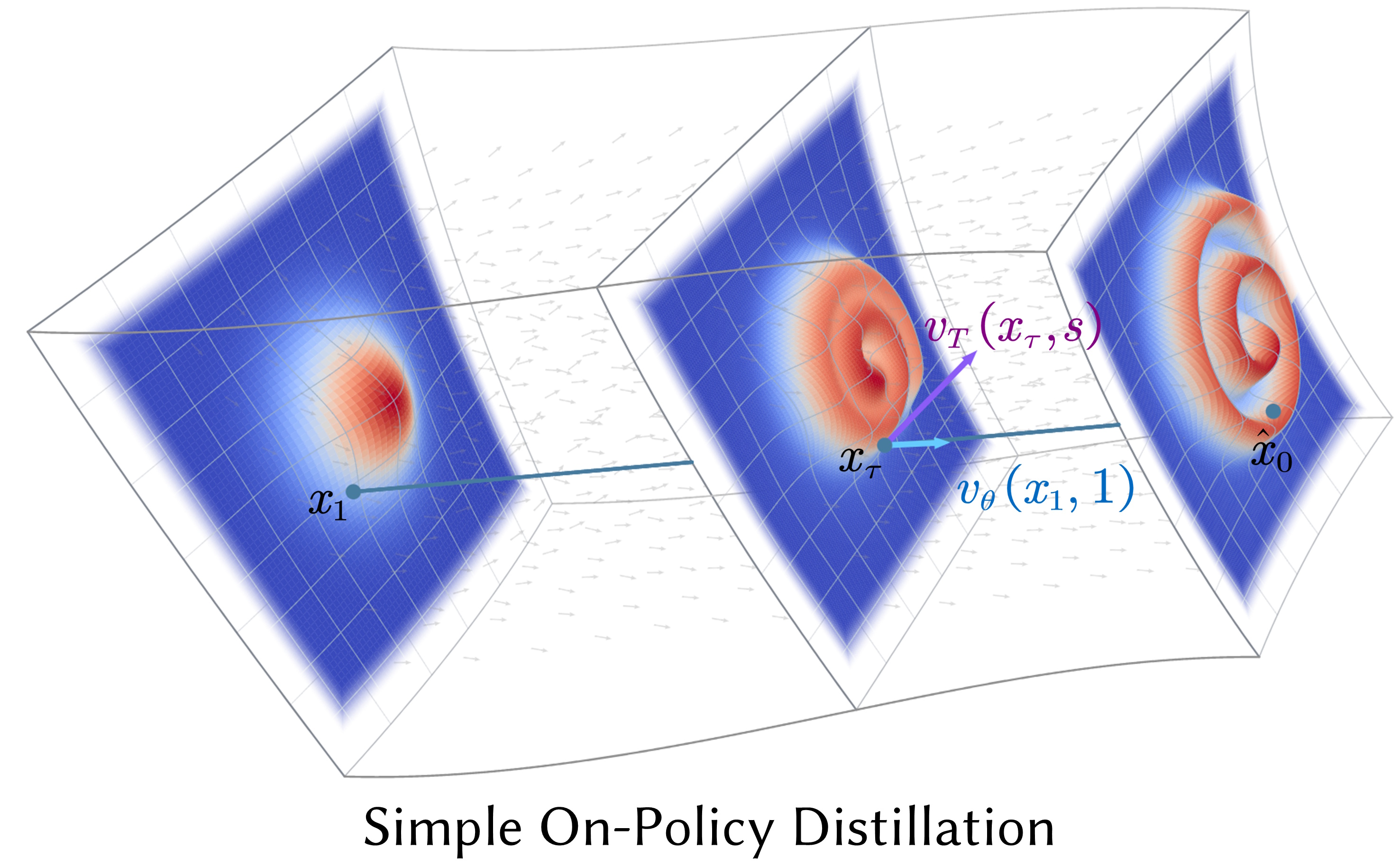}
\caption{\textbf{Simple on-policy distillation.}
In the one-step illustration, the student predicts one velocity $v_\theta(x_1,1)$
and matches it to teacher velocities $v_T(x_\tau,s)$ queried at intermediate
states along the transition to $\hat{x}_0$.
Algorithm~\ref{alg:minimal_training} gives the details.}
\label{fig:minimal_distillation}
\end{minipage}\hfill
\begin{minipage}[t]{0.51\textwidth}
\vspace{0pt}
\begin{algorithm}[H]
\small
\SetAlCapSkip{3pt}
\caption{Simple on-policy distillation}
\label{alg:minimal_training}
\KwIn{Student $v_\theta$, frozen teacher $v_T$. Grid $T=t_0>\cdots>t_N=0$.}
\KwOut{Rollout loss $\mathcal L$ for updating $\theta$.}
Sample $x\sim\mathcal N(0,I)$\;
$\mathcal L\gets 0$\;
\For{$i=0,\ldots,N-1$}{
    $x\gets\operatorname{stopgrad}(x)$\;
    $\Delta\gets t_i-t_{i+1}$\;
    $v_\theta\gets v_\theta(x,t_i)$\;
    Sample one or more query points $\tau\sim\mathcal U(0,1)$\;
    \ForEach{sampled query point $\tau$}{
        $s\gets t_i-\tau\Delta$\;
        $x_\tau\gets x-\tau\Delta v_\theta$\;
        $v_T\gets v_T(x_\tau,s)$\;
        $\mathcal L\gets\mathcal L+\|v_\theta-v_T\|_2$\;
    }
    $x\gets x-\Delta v_\theta$\;
}
\Return{the mean loss over all transitions and query points}\;
\end{algorithm}
\end{minipage}
\end{figure}

Let us set aside the complexities of existing distillation methods for a moment and start with a simple idea.
We match the student's velocity to the teacher's local predictions at states along the student's own coarse transition.
Despite the simplicity of both the state construction and the matching
objective, this formulation already enables effective few-step distillation.

Figure~\ref{fig:minimal_distillation} depicts the one-step special case from
$x_1$ to $\hat{x}_0$. We now formulate the same construction for an arbitrary
coarse transition, applied independently within every interval of a few-step
rollout.

Consider one transition from the current student rollout state $x_t$ to time
$t-\Delta$, where $\Delta>0$.
The student makes a single prediction at the transition start,
\begin{equation}
    v_\theta = v_\theta(x_t,t),
    \qquad
    x_{t-\Delta}=x_t-\Delta v_\theta .
    \label{eq:simple_coarse_update}
\end{equation}
For $\tau\sim\mathcal U(0,1)$, we construct an intermediate state along this
transition,
\begin{equation}
    x_\tau=x_t-\tau\Delta v_\theta,
    \qquad
    s=t-\tau\Delta .
    \label{eq:simple_intermediate_state}
\end{equation}
We then query the pretrained teacher velocity $v_T$ at $(x_\tau,s)$.
The same student prediction $v_\theta$ is used both to construct $x_\tau$ and
to match the teacher's local velocity. No additional student evaluation is
needed at the intermediate state. We optimize the $L_2$ distance,
\begin{equation}
    \mathcal L
    =
    \mathbb E_{\tau}
    \left[
        \left\|
            v_\theta-v_T(x_\tau,s)
        \right\|_2
    \right].
    \label{eq:simple_on_policy_objective}
\end{equation}
Throughout training, teacher outputs are treated as detached targets.

The equations above describe a single transition within a few-step student
rollout. During training, we sample different conditions and initial noises,
roll out the current student over the few-step grid, and apply the same
objective to each transition along the rollout.
Algorithm~\ref{alg:minimal_training} summarizes the rollout loss, with gradients stopped across transitions.

\begin{figure}[!t]
\centering
\includegraphics[width=1.0\textwidth]{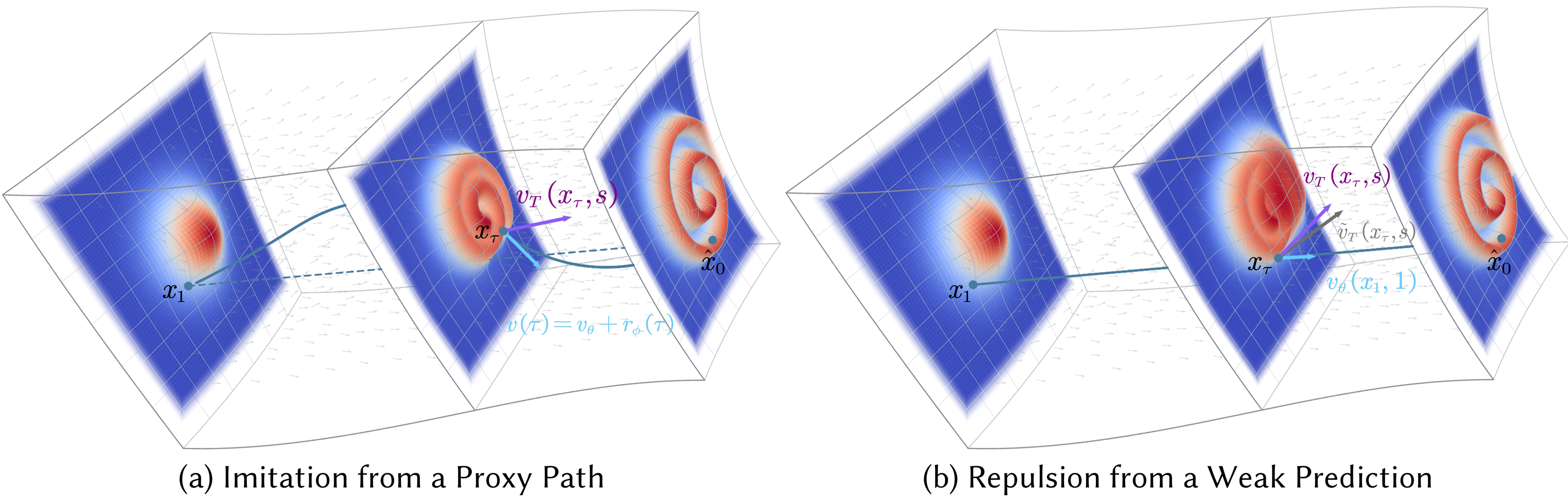}
\caption{\textbf{Refining the on-policy baseline.}
(a) A proxy path allows the local velocity to vary within a coarse transition while preserving its endpoints for a fixed $v_\theta$.
(b) Repulsion from a weaker prediction $\widetilde v_T$ complements imitation of the target teacher prediction $v_T$.
The two refinements are illustrated separately for clarity and are used only during training.}
\label{fig:proxy_repulsion}
\end{figure}

\begin{figure}[!t]
\centering
\includegraphics[width=\textwidth]{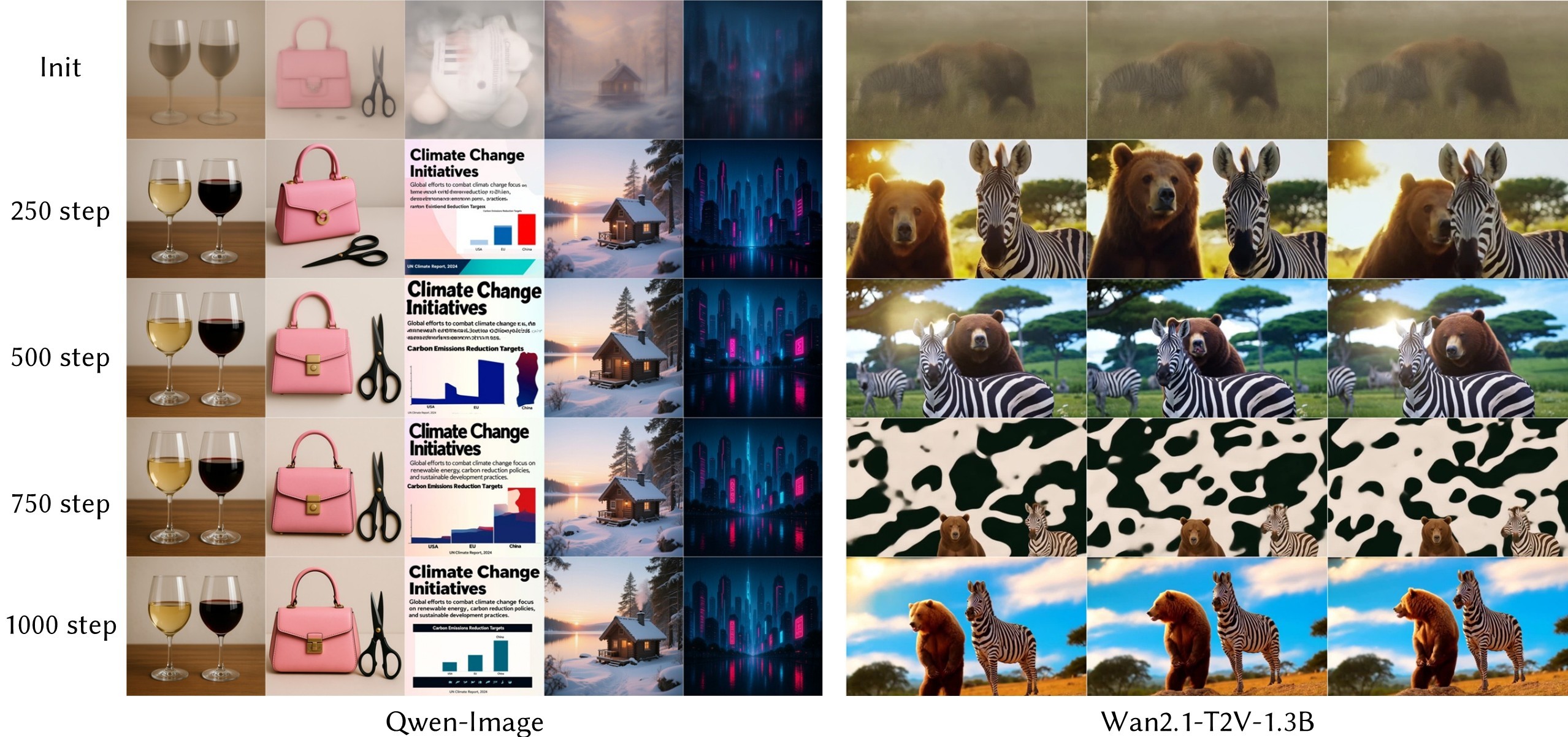}
\caption{
\textbf{Simple distillation already works.}
Fixed samples across 4-NFE training checkpoints show rapid improvement for Qwen-Image (left) but less consistent progress for Wan2.1-T2V-1.3B (right).
}
\label{fig:ablation_simple}
\end{figure}

Figure~\ref{fig:ablation_simple} shows fixed samples across training checkpoints
for two 4-NFE students. Qwen-Image shows rapid early improvements in visual
clarity and scene structure. Wan2.1-T2V-1.3B also improves substantially from
initialization, but its samples exhibit pronounced quality fluctuations, with
visible degradation followed by recovery across checkpoints. These observations
show that local velocity matching alone can yield substantial improvements in
few-step generation, although progress is not uniformly consistent across the
two models.

\begin{takeaway}
\textbf{Finding 4.}
Direct velocity matching rapidly improves Qwen-Image's 4-NFE samples and produces substantial gains in Wan2.1-T2V-1.3B, without explicit endpoint supervision.
\end{takeaway}

\begin{figure}[!ht]
\begin{minipage}[t]{0.47\textwidth}
\vspace{0pt}
\paragraph{Analysis.}
The results above show that local velocity matching alone can already yield
substantial improvements in few-step generation. With teacher initialization,
the student starts from a pretrained velocity predictor. The objective then
refines this prediction using teacher evaluations at intermediate states along
the student's own coarse transition. In this way, a prediction initially
trained to describe local dynamics is adapted to account for the teacher
velocities over a finite sampling interval.

Figure~\ref{fig:velocity_field_analysis} illustrates successive training updates in synthetic velocity fields.
At each update, we query the teacher at intermediate states on the current student transition and use its detached predictions to update the student velocity.
The updated transition then determines the query state for the next teacher evaluation.

When the teacher trajectory has low curvature and its velocity varies
moderately over the interval, accurate local teacher predictions provide largely
compatible supervision for refining the coarse transition endpoint. This allows
the student to improve its sampling behavior without an explicit endpoint
target, offering a plausible explanation for the rapid improvement of
Qwen-Image in Figure~\ref{fig:ablation_simple}. For trajectories with higher
curvature, however, the teacher velocity can change substantially in direction
within a single transition.
\end{minipage}\hfill
\begin{minipage}[t]{0.50\textwidth}
\vspace{0pt}
\centering
\includegraphics[width=\linewidth]{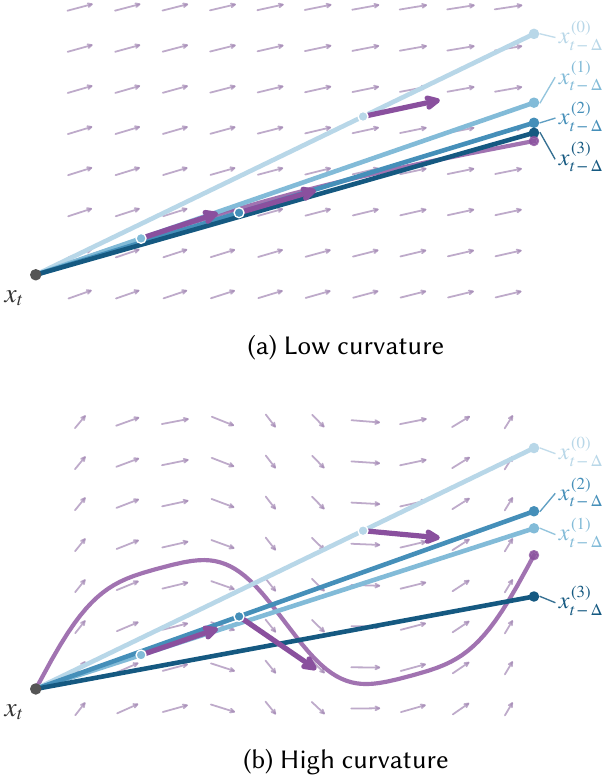}
\caption{\textbf{Curvature shapes on-policy updates.}
Three updates in low-curvature (top) and high-curvature (bottom) velocity fields.
Blue student steps darken across updates. Purple arrows and curves denote teacher queries and trajectories.}
\label{fig:velocity_field_analysis}
\end{minipage}

\par\medskip
The resulting regression targets may favor conflicting updates to the student,
making optimization less stable. This offers a possible explanation for the
degradation and recovery observed with Wan2.1-T2V-1.3B, consistent with prior
observations of large trajectory curvature in its high-noise
regime~\cite{tmd}. This analysis suggests room for improvement in
both how the student fits the teacher dynamics and how the teacher guides the
student toward high-quality generation.
\end{figure}

\subsection{Imitation from a Proxy Path}
\label{sec:method-pivot}

To accommodate variations in the teacher velocity within a coarse transition,
we introduce a proxy path that shares the endpoints of the student transition
while allowing its local velocity to vary, as illustrated in
\mbox{Figure~\ref{fig:proxy_repulsion}(a)}.
We add a time-dependent residual to the coarse student velocity,
\begin{equation}
    v(\tau)=v_\theta+r_\phi(\tau),
    \label{eq:pivot_local_velocity}
\end{equation}
subject to
\begin{equation}
    \int_0^1 r_\phi(\tau)\,d\tau=0.
    \label{eq:zero_integral}
\end{equation}
The corresponding proxy state is obtained by integrating this local velocity,
\begin{equation}
    x_\tau
    =
    x_t-\Delta\int_0^\tau v(r)\,dr,
    \qquad
    s=t-\tau\Delta.
    \label{eq:pivot_intermediate_state}
\end{equation}
\begin{samepage}
The zero-integral constraint ensures that the proxy path starts at $x_t$ and
ends at $x_t-\Delta v_\theta$. Thus, the residual can model local velocity
variations without introducing an additional displacement at the end of the
transition. We match local proxy velocities to teacher predictions at the same states,
\begin{equation}
    \mathcal L
    =
    \mathbb E_\tau
    \left[
        \left\|
            v(\tau)-v_T(x_\tau,s)
        \right\|_2
    \right].
    \label{eq:pivot_on_policy_loss}
\end{equation}
\end{samepage}
The student and the residual predictor are optimized jointly. The residual
adjusts both the intermediate query states and the local
velocities used for imitation, while the coarse student prediction continues
to determine the endpoint. At inference, the student directly takes the coarse
update in Eq.~\eqref{eq:simple_coarse_update}, without evaluating the residual
predictor.
Setting $r_\phi\equiv0$ recovers the simple baseline.

\begin{samepage}
\paragraph{Implementation.}
We represent the velocity residual with a finite basis expansion~\cite{ramsay2005functional},
choosing zero-integral functions to preserve the transition endpoints,
\begin{equation}
    r_\phi(\tau)
    =
    \sum_{m=1}^{P} a_m b_m(\tau),
    \qquad
    \int_0^1 b_m(\tau)\,d\tau=0,
    \label{eq:method_general_residual_basis}
\end{equation}
\end{samepage}
where each $a_m$ has the same shape as the student velocity.
Pivot predicts the coefficient vectors once per transition from features of the
student's $K$-th layer from the end, where $K=1$ denotes the final layer.
It processes these features with $M$ Transformer blocks and uses a
zero-initialized output projection to produce all $P$ coefficient vectors,
so training starts from the simple baseline. Under pretrained initialization,
the Pivot blocks are copied from the final $M$ student blocks.
Late-block sensitivity in Appendix~\ref{app:block_swap} supports this design.
Appendix~\ref{app:training-architecture} shows the architecture.

Pivot supports different bases. A polynomial path parameterization uses
\begin{equation}
    B_m(\tau)=\tau(1-\tau)(2\tau-1)^{m-1},
    \qquad
    b_m(\tau)=B'_m(\tau).
    \label{eq:method_polynomial_path_basis}
\end{equation}
Since $B_m(0)=B_m(1)=0$, its derivative satisfies the zero-integral constraint.
We explore this and other parameterizations
in our ablations, with full definitions in Appendix~\ref{app:residual_basis}.

\subsection{Repulsion from a Reference Prediction}
\label{sec:method-repulsion}

We further refine the distillation signal using an additional prediction from the
same teacher as a negative reference. We
complement imitation of the target prediction with repulsion from this reference,
as illustrated in \mbox{Figure~\ref{fig:proxy_repulsion}(b)}. We retain $v_T(x_\tau,s)$ as the
imitation target and denote the reference prediction at the same state and time
by $\widetilde v_T(x_\tau,s)$. Its construction depends on the teacher's
conditioning and inference procedure, as described below. We optimize
\begin{equation}
    \mathcal L
    =
    \mathbb E_\tau\left[
    \left\|v(\tau)-v_T(x_\tau,s)\right\|_2
    -
    \lambda\left\|v(\tau)-\widetilde v_T(x_\tau,s)\right\|_2
    \right],
    \qquad
    0\leq\lambda<1.
    \label{eq:repulsion_objective}
\end{equation}
The first term matches the target teacher prediction, while the second
encourages separation from the reference prediction. Setting $\lambda=0$
recovers imitation alone.

\paragraph{Constructing reference predictions.}
The repulsive objective does not require a particular construction of
$\widetilde v_T$. For teachers using classifier-free guidance, an unconditional
or conditional prediction can serve as the negative reference, corresponding to reference guidance scales $w=0$ and $w=1$, respectively, while $v_T$ remains the guided
prediction used for imitation. For embedded-guidance models, we use a lower guidance input instead.
For models with guidance already distilled into
their weights, we can instead obtain $\widetilde v_T$ through an additional
forward pass with a layer skipped. These choices use the same pretrained model
and adapt the negative reference to its inference procedure.
\paragraph{Training and inference.}
We train the student and Pivot jointly on on-policy rollouts using
Eq.~\eqref{eq:repulsion_objective}. The proxy path is
used only to construct supervision at the sampled query points. The next rollout
state follows the student's coarse update and is detached before the next
transition. We average losses over transitions, query points, and batch samples.
Optimizer updates occur outside the rollout.

At inference, only the student follows the coarse updates on the same time grid.
Neither teacher predictions nor Pivot are evaluated, so the student architecture
and number of function evaluations are unchanged.
Algorithms~\ref{alg:pivot_training} and~\ref{alg:pivot_inference} summarize the
training loss and student sampler, respectively.

\begin{takeaway}
\textbf{Finding 5.}
The student can learn the coarse update without fitting every local velocity with a single prediction.
Pivot models within-step velocity variations through a zero-integral residual, preserving the student's transition endpoints.
\end{takeaway}

\begin{figure}[!t]
\begin{algorithm}[H]
\small
\caption{On-policy distillation with a proxy path}
\label{alg:pivot_training}
\KwIn{Student parameters $\theta$, Pivot parameters $\phi$, frozen teacher prediction rules $v_T$ and $\widetilde v_T$.}
\KwIn{Time grid $T=t_0>t_1>\cdots>t_N=0$, zero-integral bases $\{b_m\}_{m=1}^{P}$, repulsion weight $\lambda$.}
\KwOut{Rollout loss $\mathcal L$ for updating $\theta$ and $\phi$.}
Sample $x\sim\mathcal N(0,I)$\;
$\mathcal L\gets 0$\;
\For{$i=0,\ldots,N-1$}{
    $x\gets\operatorname{stopgrad}(x)$\;
    $\Delta\gets t_i-t_{i+1}$\;
    $v_\theta\gets v_\theta(x,t_i)$\tcp*{Retain features from the $K$-th layer from the end}
    Predict $\{a_m\}_{m=1}^{P}$ with Pivot from the retained features\;
    Sample one or more query points $\tau\sim\mathcal U(0,1)$\;
    \ForEach{sampled query point $\tau$}{
        $s\gets t_i-\tau\Delta$\;
        $r_\phi(\tau)\gets\sum_{m=1}^{P}a_m b_m(\tau)$\;
        $v(\tau)\gets v_\theta+r_\phi(\tau)$\;
        $x_\tau\gets x-\Delta\left[\tau v_\theta+\sum_{m=1}^{P}a_m\int_0^\tau b_m(r)\,dr\right]$\;
        $v_T\gets v_T(x_\tau,s)$\;
        $\widetilde v_T\gets\widetilde v_T(x_\tau,s)$\;
        $\mathcal L\gets\mathcal L+\|v(\tau)-v_T\|_2-\lambda\|v(\tau)-\widetilde v_T\|_2$\;
    }
    $x\gets x-\Delta v_\theta$\tcp*{Continue the student rollout}
}
\Return{the mean loss over all transitions and query points}\;
\end{algorithm}

\begin{algorithm}[H]
\small
\caption{Few-step student inference}
\label{alg:pivot_inference}
\KwIn{Trained student $v_\theta$, time grid $T=t_0>\cdots>t_N=0$.}
\KwOut{Generated sample $\hat x_0$.}
Sample $x\sim\mathcal N(0,I)$\;
\For{$i=0,\ldots,N-1$}{
    $\Delta\gets t_i-t_{i+1}$\;
    $x\gets x-\Delta v_\theta(x,t_i)$\;
}
\Return{$\hat x_0=x$}\;
\end{algorithm}
\end{figure}

\begingroup
\setlength{\textfloatsep}{10pt plus 2pt minus 2pt}
\setlength{\floatsep}{8pt plus 2pt minus 2pt}
\setlength{\intextsep}{8pt plus 2pt minus 2pt}
\makeatletter
\setlength{\@fptop}{0pt}
\makeatother
\section{Few-step Distillation}
\label{sec:fewstep}

We evaluate SMD under a few-step sampling budget across image generation and editing, 3D shape and texture generation, video, and joint audio--video generation.
Figures~\ref{fig:gallery_qwen_image}--\ref{fig:gallery_minimax_h3} give an overview of the distilled students' outputs.
For tasks with quantitative benchmarks, we compare with existing few-step methods and include multi-step teachers as references where available.
We evaluate 3D generation and reference-conditioned audio--video generation qualitatively.
Appendix~\ref{app:fewstep-details} provides training and evaluation details.

\subsection{Image Generation}
\label{sec:fewstep-image}

We first examine prompt alignment and variation across repeated generations, two aspects of image generation that a few-step student should retain.
We distill FLUX.1-dev~\cite{flux1} (12B) and Qwen-Image~\cite{qwenimage} (20B) into 4-NFE text-to-image models using prompt-only, on-policy training.
For FLUX.1-dev, we compare with Hyper-SD~\cite{hypersd}, SenseFlow~\cite{senseflow}, and $\pi$-Flow~\cite{piflow}.
For Qwen-Image, we compare with Lightning~\cite{qwenimagelightning}, TwinFlow~\cite{twinflow}, $\pi$-Flow, and the Euler variant of PDD~\cite{pdd}.
We evaluate publicly released baseline checkpoints under matched inference and evaluation settings. For PDD, we report the results from the original paper, as no public checkpoint is available.
All few-step models use 4-NFE.
GenEval~\cite{geneval} measures compositional alignment and DPG-Bench~\cite{dpgbench} dense-prompt alignment. OneIG-Bench-EN~\cite{oneig} evaluates alignment, text, diversity, style, and reasoning.

\begin{figure}[!t]
\centering
\includegraphics[width=0.9\textwidth]{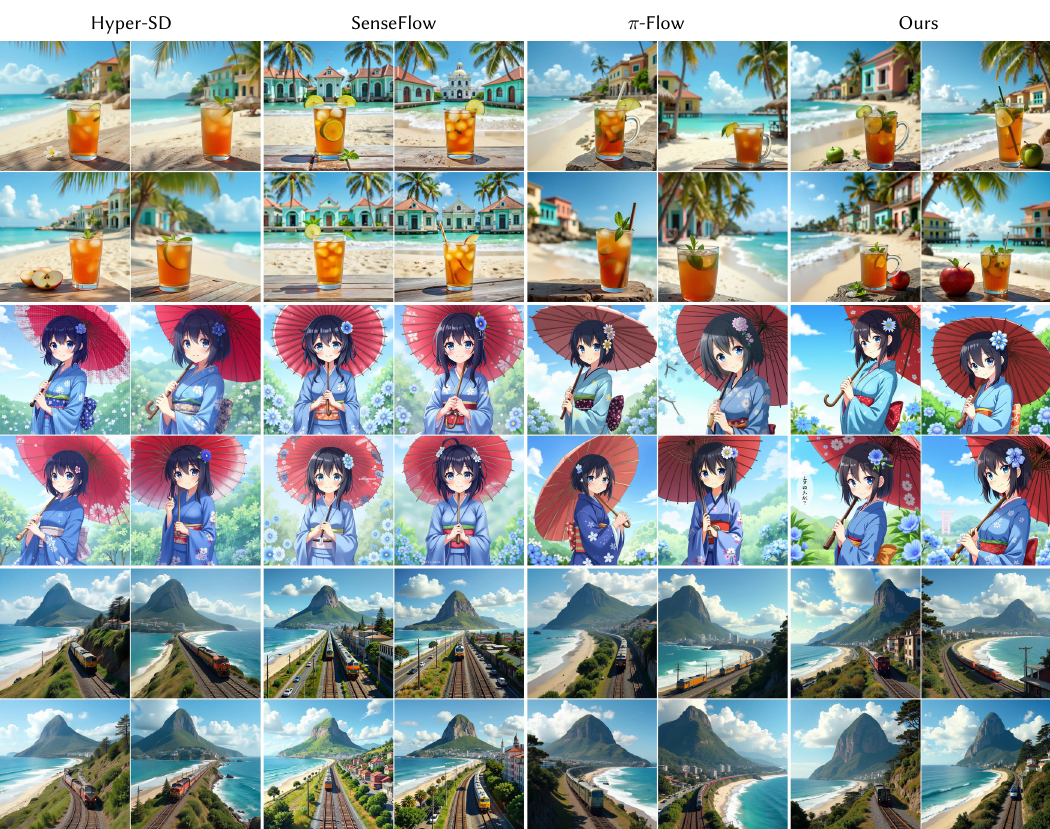}
\caption{\textbf{Image quality and variation with FLUX.1-dev.}
Each two-column group shows four 4-NFE samples per prompt for one method, allowing comparison of composition and detail.}
\label{fig:image_gen_flux1dev}
\end{figure}

\begin{table*}[!tbp]
\caption{Text-to-image generation with FLUX.1-dev. NFE denotes the number of network evaluations at inference. Best and second-best results among the few-step methods are indicated by boldface and underlining, respectively. The multi-step Teacher is excluded from this ranking.}
\vspace{-0.1in}
\centering
\small
\fitcomparisontable{0.95}{
\begin{tabular}{lccccccccc}
\toprule
\multirow{2}{*}{Method}
& \multirow{2}{*}{NFE}
& \multirow{2}{*}{GenEval$\uparrow$}
& \multirow{2}{*}{DPG-Bench$\uparrow$}
& \multicolumn{6}{c}{OneIG-Bench-EN$\uparrow$} \\ \cmidrule(lr){5-10}
& & & & Overall & Align. & Text & Diversity & Style & Reasoning \\ \midrule
\rowcolor{gray!10}
Teacher & 50 & 0.664 & 84.2 & 0.430 & 0.790 & 0.556 & 0.239 & 0.308 & 0.258 \\ \midrule
Hyper-SD & 4 & 0.667 & 83.0 & 0.399 & 0.786 & 0.423 & 0.185 & \clb{0.361} & 0.239 \\
SenseFlow & 4 & 0.643 & 82.6 & 0.380 & 0.774 & 0.388 & 0.156 & 0.348 & 0.234 \\
$\pi$-Flow & 4 & \clb{0.675} & \clb{84.3} & \clb{0.419} & \clb{0.799} & \cb{0.460} & \cb{0.229} & 0.360 & \clb{0.251} \\
Ours & 4 & \cb{0.715} & \cb{84.8} & \cb{0.420} & \cb{0.810} & \clb{0.455} & \clb{0.209} & \cb{0.370} & \cb{0.256} \\ \bottomrule
\end{tabular}
}
\label{tab:image_generation}
\end{table*}

Tables~\ref{tab:image_generation} and~\ref{tab:qwen_image_generation} report the quantitative results.
At 4-NFE, SMD obtains the highest GenEval and DPG-Bench scores among the compared few-step methods on both backbones.
For FLUX.1-dev, GenEval increases from 0.675 with $\pi$-Flow to 0.715 with SMD, and DPG-Bench increases from 84.3 to 84.8. Both exceed the corresponding 50-NFE teacher scores.
For Qwen-Image, SMD reaches 0.873 on GenEval and 88.7 on DPG-Bench.
SMD and the teacher both have a reported OneIG Overall score of 0.540.
The complete OneIG breakdown also shows stronger Qwen-Image text rendering, with a Text score of 0.944 compared with 0.889 for the teacher and 0.921 for Lightning.

\begin{figure}[!t]
\centering
\includegraphics[width=0.9\textwidth]{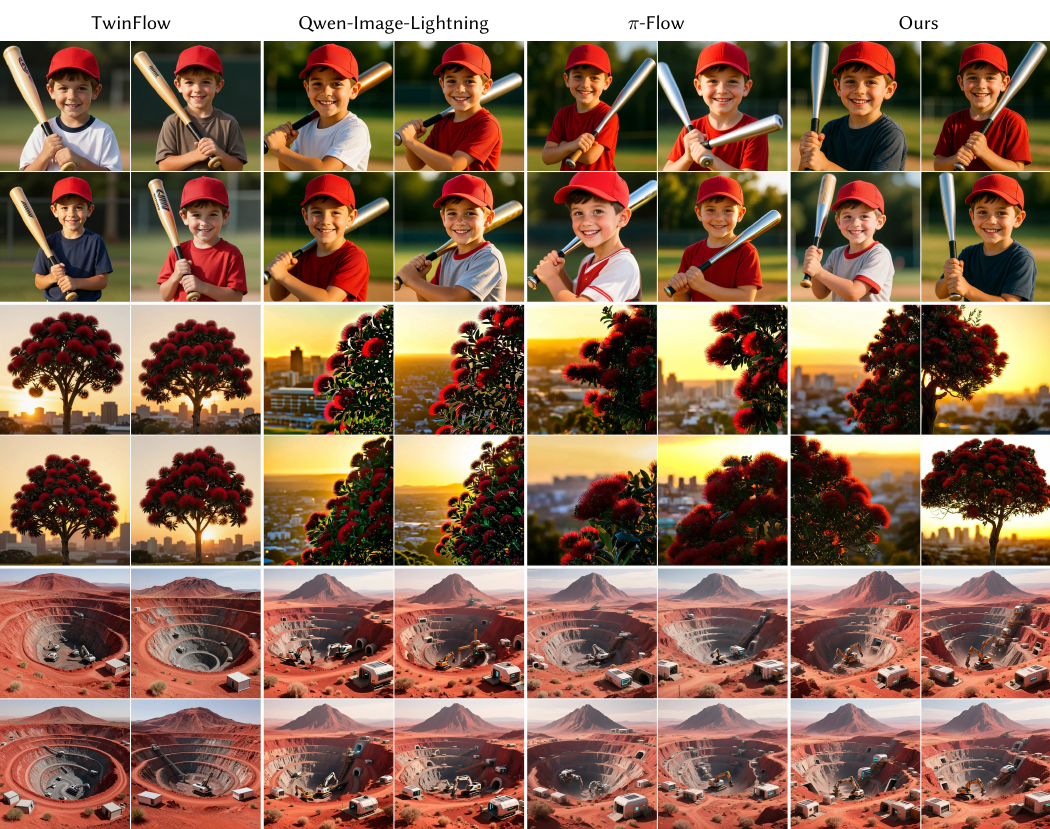}
\caption{\textbf{Image quality and variation with Qwen-Image.}
Each two-column group shows four 4-NFE samples per prompt for one method, allowing comparison of pose, viewpoint, and composition.}
\label{fig:image_gen_qwen_image}
\end{figure}

\begin{table*}[!tbp]
\caption{Text-to-image generation with Qwen-Image. NFE counts network evaluations. $50\times2$ includes both CFG branches. Best and second-best few-step results are bold and underlined. $\dagger$ denotes Euler-variant results from PDD~\cite{pdd}, Tables 3 and 6--8. No public checkpoint was available.}
\vspace{-0.1in}
\centering
\small
\scalebox{0.92}{
\begin{tabular}{lccccccccc}
\toprule
\multirow{2}{*}{Method}
& \multirow{2}{*}{NFE}
& \multirow{2}{*}{GenEval$\uparrow$}
& \multirow{2}{*}{DPG-Bench$\uparrow$}
& \multicolumn{6}{c}{OneIG-Bench-EN$\uparrow$} \\ \cmidrule(lr){5-10}
& & & & Overall & Align. & Text & Diversity & Style & Reasoning \\ \midrule
\rowcolor{gray!10}
Teacher & $50\times2$ & 0.874 & 88.3 & 0.540 & 0.880 & 0.889 & 0.195 & 0.428 & 0.307 \\ \midrule
Lightning & 4 & 0.852 & 88.3 & 0.527 & \cb{0.886} & \clb{0.921} & 0.108 & 0.411 & \cb{0.310} \\
TwinFlow & 4 & 0.818 & 86.0 & 0.502 & 0.860 & 0.872 & 0.152 & 0.361 & 0.263 \\
$\pi$-Flow & 4 & 0.850 & 88.2 & 0.532 & 0.878 & 0.876 & \clb{0.180} & \cb{0.430} & 0.296 \\
PDD$^{\dagger}$ & 4 & \clb{0.860} & \clb{88.5} & \clb{0.535} & 0.879 & 0.897 & \cb{0.192} & 0.417 & 0.292 \\
Ours & 4 & \cb{0.873} & \cb{88.7} & \cb{0.540} & \clb{0.883} & \cb{0.944} & 0.144 & \clb{0.429} & \clb{0.300} \\ \bottomrule
\end{tabular}
}
\label{tab:qwen_image_generation}
\end{table*}

Beyond prompt alignment, SMD retains meaningful variation across repeated generations.
Its OneIG Diversity score exceeds the DMD-based SenseFlow baseline on FLUX.1-dev and Lightning on Qwen-Image.
Figures~\ref{fig:image_gen_flux1dev} and~\ref{fig:image_gen_qwen_image} further show variations in viewpoint, pose, and composition across four samples from each method and prompt.

SMD achieves these results with a compact training and inference design.
Our Qwen-Image student is trained with a global batch size of 16, compared with the reported batch size of 2048 for PDD~\cite{pdd}.
At inference, the student directly predicts each coarse update without $\pi$-Flow's substep integration.
Unlike PDD, which learns parallel output heads that can be fused for inference, SMD retains the original student output head and uses Pivot only during training.

\subsection{Image Editing}
\label{sec:fewstep-editing}

We next extend SMD to instruction-based image editing, where the student must apply the requested modification while preserving unrelated content from the source image.
We distill Qwen-Image-Edit-2511~\cite{qwenimage} to 4-NFE using the source image and editing instruction as conditions for the on-policy rollout, without requiring target edited images.
We compare with the publicly released Lightning checkpoint~\cite{qwenimageeditlightning} under matched settings and include the multi-step teacher as a reference.
We evaluate operation-level editing performance on ImgEdit-Bench~\cite{imgedit}, and semantic consistency, perceptual quality, and overall quality on GEdit-Bench-EN~\cite{step1xedit}.

\begin{figure}[!t]
\centering
\includegraphics[width=0.9\textwidth]{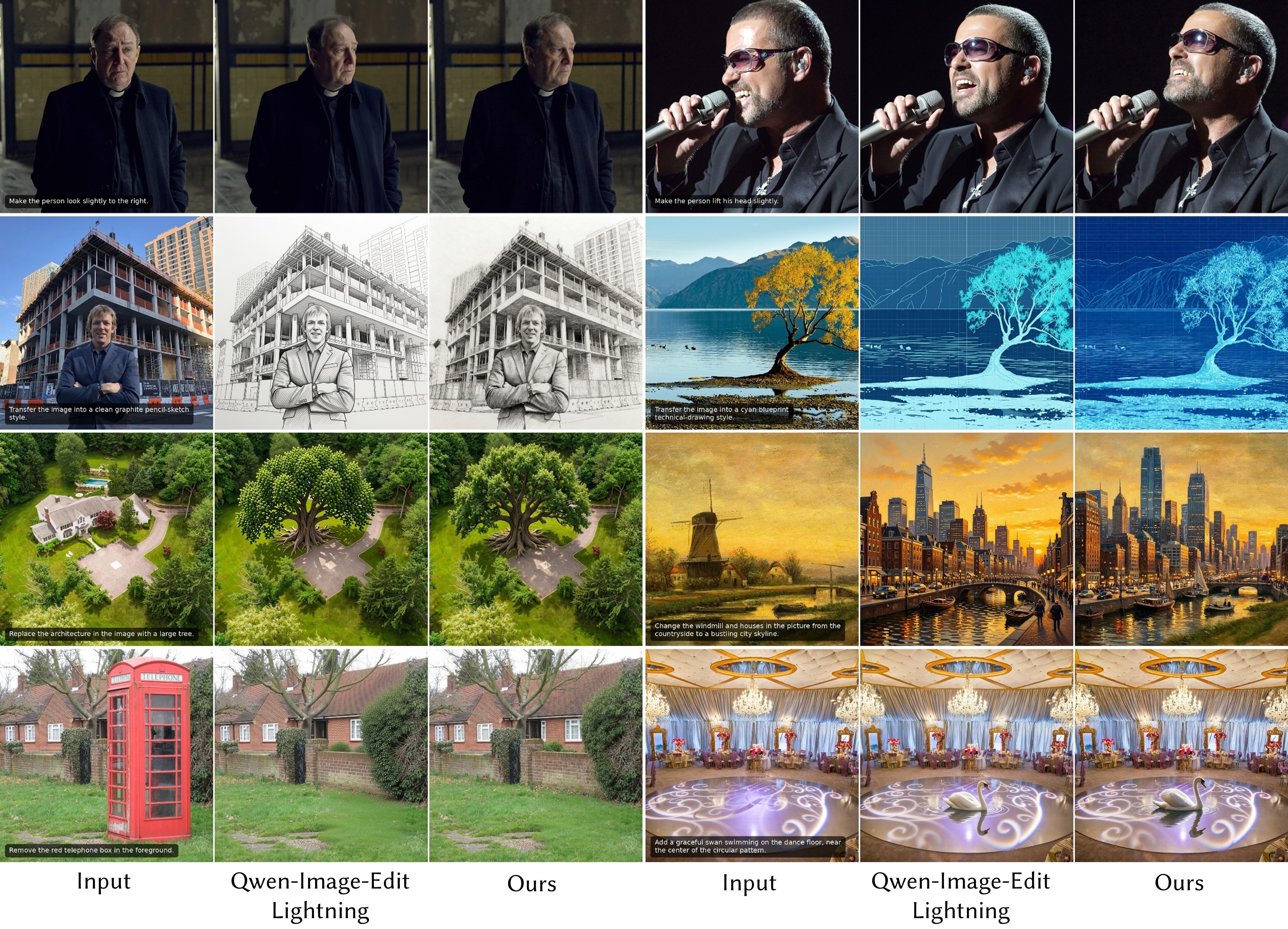}
\caption{\textbf{Instruction-guided image editing.}
Each triplet shows the input, Lightning, and ours with Qwen-Image-Edit-2511 at 4-NFE under the same editing instruction.}
\label{fig:image_edit_qwen_image_edit}
\end{figure}

\begin{figure}[!t]
\centering
\includegraphics[width=0.9\textwidth]{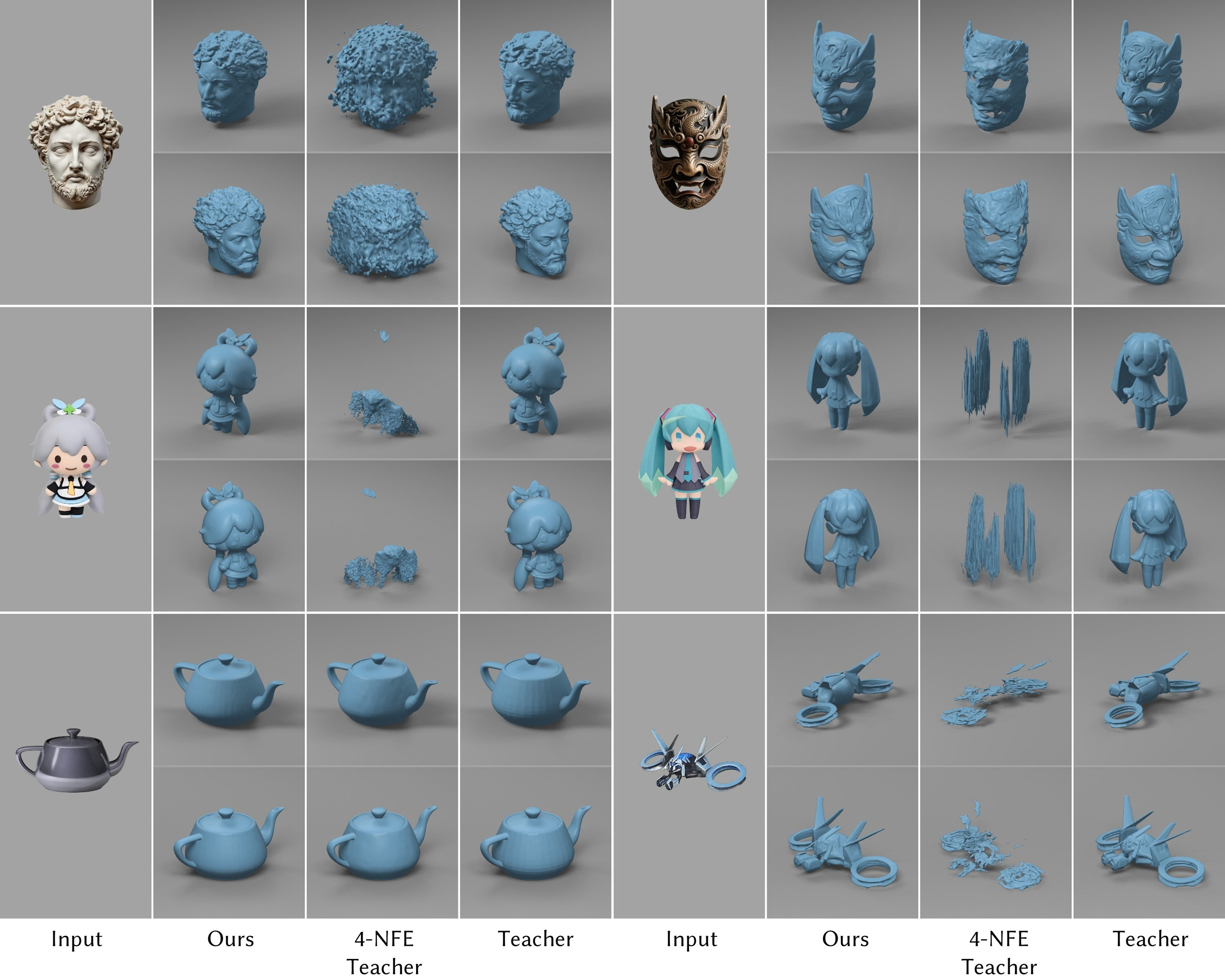}
\caption{\textbf{Few-step image-to-shape generation.}
Hunyuan3D-2.1 outputs are compared under the same input images.
Each example shows the input image and two rendered views from ours, the 4-NFE teacher, and the multi-step teacher.}
\label{fig:hunyuan3d_shape}
\end{figure}

\newcommand{\rothead}[1]{\scalebox{0.8}{\rotatebox[origin=c]{70}{\strut #1}}}

\begin{table*}[!tbp]
\caption{Instruction-based editing with Qwen-Image-Edit-2511. $G_{\mathrm{SC}}$, $G_{\mathrm{PQ}}$, and $G_{\mathrm{O}}$ denote GPT-4.1-based semantic consistency, perceptual quality, and overall quality on GEdit-Bench-EN. Boldface marks the better 4-NFE student result. The multi-step Teacher is reported as a reference.}
\vspace{-0.1in}
\centering
\small
\fitcomparisontable{0.90}{
\begin{tabular}{lcccccccccccccc}
\toprule
\multirow{3}{*}{Method}
& \multirow{3}{*}{NFE}
& \multicolumn{10}{c}{ImgEdit-Bench$\uparrow$}
& \multicolumn{3}{c}{GEdit-Bench-EN$\uparrow$} \\ \cmidrule(lr){3-12} \cmidrule(lr){13-15}
& & \rothead{Overall} & \rothead{Adjust} & \rothead{Style} & \rothead{Background} & \rothead{Extract} & \rothead{Remove} & \rothead{Replace} & \rothead{Add} & \rothead{Hybrid} & \rothead{Action}
& $G_{\mathrm{SC}}$ & $G_{\mathrm{PQ}}$ & $G_{\mathrm{O}}$ \\ \midrule
\rowcolor{gray!10}
Teacher & $40\times2$ & 4.51 & 4.57 & 4.89 & 4.36 & 4.13 & 4.46 & 4.70 & 4.54 & 4.16 & 4.81 & 8.30 & 8.20 & 7.88 \\ \midrule
Lightning & 4 & 4.48 & 4.40 & \cb{4.76} & \cb{4.39} & \cb{4.21} & 4.54 & 4.72 & \cb{4.67} & 3.87 & \cb{4.80} & 8.95 & 7.74 & 8.14 \\
Ours & 4 & \cb{4.51} & \cb{4.58} & 4.74 & 4.37 & 3.98 & \cb{4.74} & \cb{4.76} & \cb{4.67} & \cb{3.95} & \cb{4.80} & \cb{8.96} & \cb{7.79} & \cb{8.19} \\ \bottomrule
\end{tabular}
}
\label{tab:image_editing}
\end{table*}

Table~\ref{tab:image_editing} shows that SMD raises the ImgEdit-Bench overall score from Lightning's 4.48 to 4.51, the score also reported for the multi-step teacher.
It also improves the three GEdit-Bench metrics, including overall quality from 8.14 to 8.19.
The largest gains over Lightning are in removal and adjustment, while Lightning remains stronger on extraction.

Figure~\ref{fig:image_edit_qwen_image_edit} provides qualitative comparisons across editing instructions.
In the building-to-tree example, Lightning produces a stylized canopy with regular, repeated foliage patterns, while our result has more irregular foliage and blends more naturally with the surrounding vegetation.

\subsection{3D Shape and Texture}
\label{sec:fewstep-3d}

We next examine whether the same distillation formulation applies to different 3D latent representations.
Hunyuan3D-2.1~\cite{hunyuan3d21} uses a fixed-size set of shape latents.
\mbox{TRELLIS.2}~\cite{trellis2} uses spatially structured sparse latents for geometry and PBR appearance.
We assess the generated assets qualitatively through rendered views, surface visualizations, and detail crops.

For Hunyuan3D-2.1, we distill the shape-generation model to 4-NFE and compare it with the teacher sampled directly at 4-NFE and with standard multi-step teacher inference.
As shown in Figure~\ref{fig:hunyuan3d_shape}, the student preserves recognizable object structure and surface details, whereas directly running the teacher in four steps often produces fragmented or incomplete geometry.

TRELLIS.2 follows a multi-stage generation pipeline.
Its high-resolution cascade first generates a sparse structure and coarse shape, upsamples their spatial support for high-resolution shape generation, and finally synthesizes shape-conditioned textures.
We distill the high-resolution shape and texture models independently to 4-NFE, while leaving the preceding sparse-structure and coarse-shape stages unchanged.
Thus, the reported 4-NFE applies separately to each of the two distilled stages rather than to the complete asset-generation pipeline.

\begin{figure}[!t]
\centering
\includegraphics[width=0.9\textwidth]{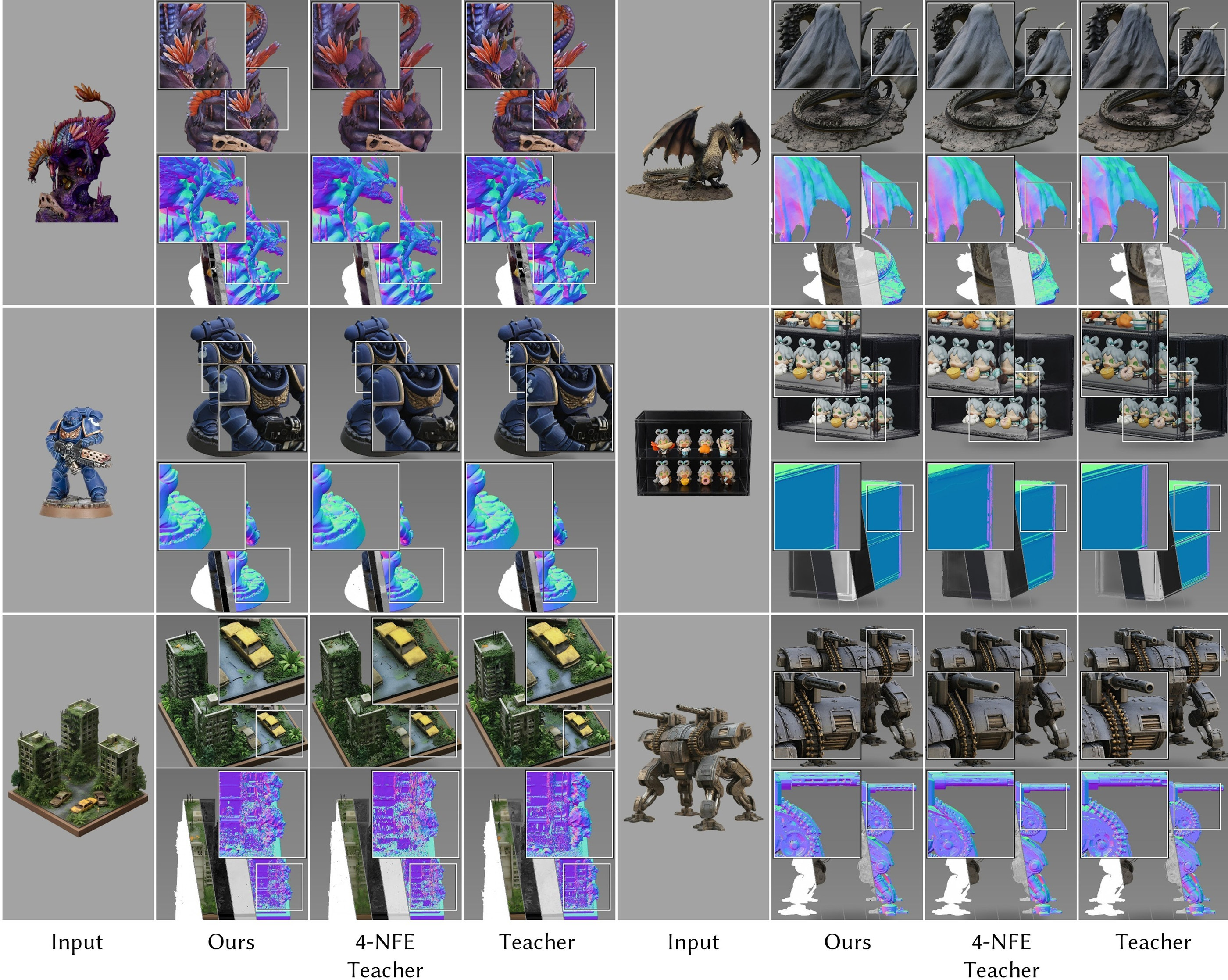}
\caption{\textbf{Few-step shape and texture generation.}
TRELLIS.2 renders, detail crops, and surface visualizations compare SMD with direct 4-NFE teacher sampling and standard multi-step teacher inference.
Only the final high-resolution shape and texture stages are distilled, each to 4-NFE, while the preceding stages remain unchanged.}
\label{fig:trellis2}
\end{figure}

Figure~\ref{fig:trellis2} compares the generated geometry and texture through rendered views and detail crops.
Together with Hunyuan3D-2.1, these results show that SMD applies to both fixed-size shape latents and spatially structured sparse representations.

\subsection{Video Generation}
\label{sec:fewstep-video}

We further evaluate the method on text-to-video generation, where distillation must preserve temporal structure as well as frame quality.
We distill Wan2.1-T2V~\cite{wan} at both 1.3B and 14B scales to 4-NFE.
We compare with our reproduced DMD2~\cite{dmd2} baseline and publicly released rCM~\cite{rcm} and AnyFlow~\cite{anyflow} checkpoints under matched settings, and quote PDD results from the original paper~\cite{pdd}.
Evaluation uses VBench~\cite{vbench} on the Self-Forcing prompt set~\cite{selfforcing}.

\begin{figure}[!t]
\centering
\includegraphics[width=0.9\textwidth]{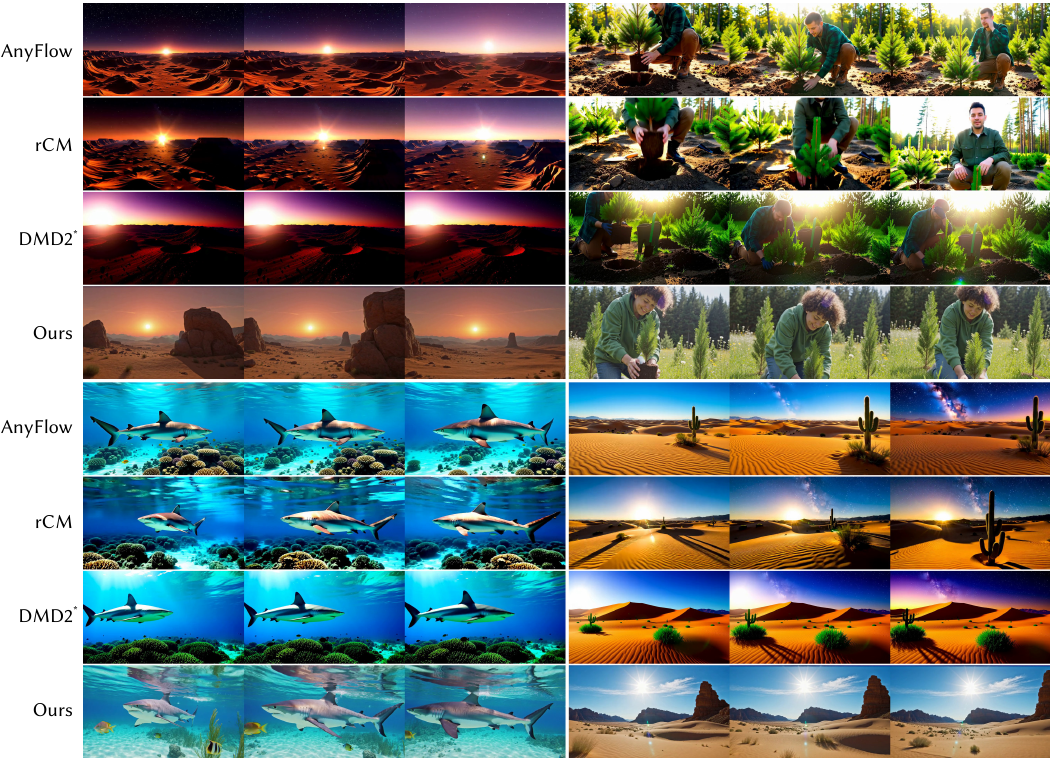}
\caption{\textbf{Few-step video generation.}
Wan2.1-T2V-14B methods are compared under matched prompts, with three chronological frames per video.}
\label{fig:video_gen_wan_14b}
\end{figure}

\begin{table*}[!t]
\caption{VBench evaluation of Wan2.1-T2V at 4-NFE.
DD, MS, and C denote Dynamic Degree, Motion Smoothness, and Color.
Boldface and underlining indicate the best and second-best few-step results within each backbone.
$^{*}$Our reproduced DMD2 results.
$^{\dagger}$Teacher and PDD results are quoted from PDD~\cite{pdd}.
The teacher uses UniPC. PDD uses Euler for 1.3B and Midpoint for 14B.}
\label{tab:vbench}
\centering
\small
\setlength{\tabcolsep}{4.5pt}

\begin{tabular}{@{}lc|ccccc|ccccc@{}}
\toprule
& & \multicolumn{5}{c|}{\textbf{Wan2.1-T2V-1.3B}}
  & \multicolumn{5}{c}{\textbf{Wan2.1-T2V-14B}} \\
\cmidrule(lr){3-7}\cmidrule(lr){8-12}
Method & NFE
& Quality & Semantic & DD & MS & C
& Quality & Semantic & DD & MS & C \\
\midrule
\rowcolor{gray!10}
Teacher$^{\dagger}$ & $50{\times}2$
& 84.90 & 79.22 & -- & -- & --
& 84.56 & 81.24 & -- & -- & -- \\ \midrule

DMD2$^{*}$ & 4
& 84.68 & 79.39 & 45.00 & \underline{98.81} & 84.63
& 84.33 & 80.26 & 49.72 & \textbf{98.76} & \textbf{87.81} \\

rCM & 4
& 84.61 & \underline{79.56} & \textbf{81.94} & 97.64 & 85.39
& 85.19 & \underline{81.21} & \underline{76.39} & 97.81 & 82.91 \\

AnyFlow & 4
& 85.40 & \textbf{81.47} & 68.33 & 98.29 & \underline{87.67}
& \textbf{85.78} & \textbf{82.10} & 73.61 & 98.10 & 85.79 \\

PDD$^{\dagger}$ & 4
& \textbf{85.99} & 78.22 & 72.50 & \textbf{99.02} & 85.41
& 85.69 & 80.71 & 74.72 & \underline{98.52} & \underline{86.59} \\

Ours & 4
& \underline{85.64} & 78.97 & \underline{79.44} & 98.66 & \textbf{89.19}
& \underline{85.72} & 80.52 & \textbf{83.89} & \underline{98.52} & 86.18 \\
\bottomrule
\end{tabular}
\end{table*}

Table~\ref{tab:vbench} reports the Quality and Semantic aggregates together with selected motion and color-adherence dimensions.
Across both Wan model scales, SMD remains competitive with strong 4-NFE baselines while retaining the original student inference architecture.
Relative to the reported multi-step teacher results, SMD improves the Quality aggregate at both scales while remaining close in Semantic.
Appendix~\ref{app:vbench-breakdown} reports all 16 VBench dimensions, comprising seven Quality and nine Semantic scores.

Figure~\ref{fig:video_gen_wan_14b} provides qualitative comparisons for the 14B models.
In the illustrated underwater and desert scenes, our outputs have less pronounced color saturation and contrast than the baselines, with softer lighting transitions.
The frames also show changes in subject position and lighting.

\begin{figure}[!t]
\centering
\includegraphics[width=0.99\textwidth]{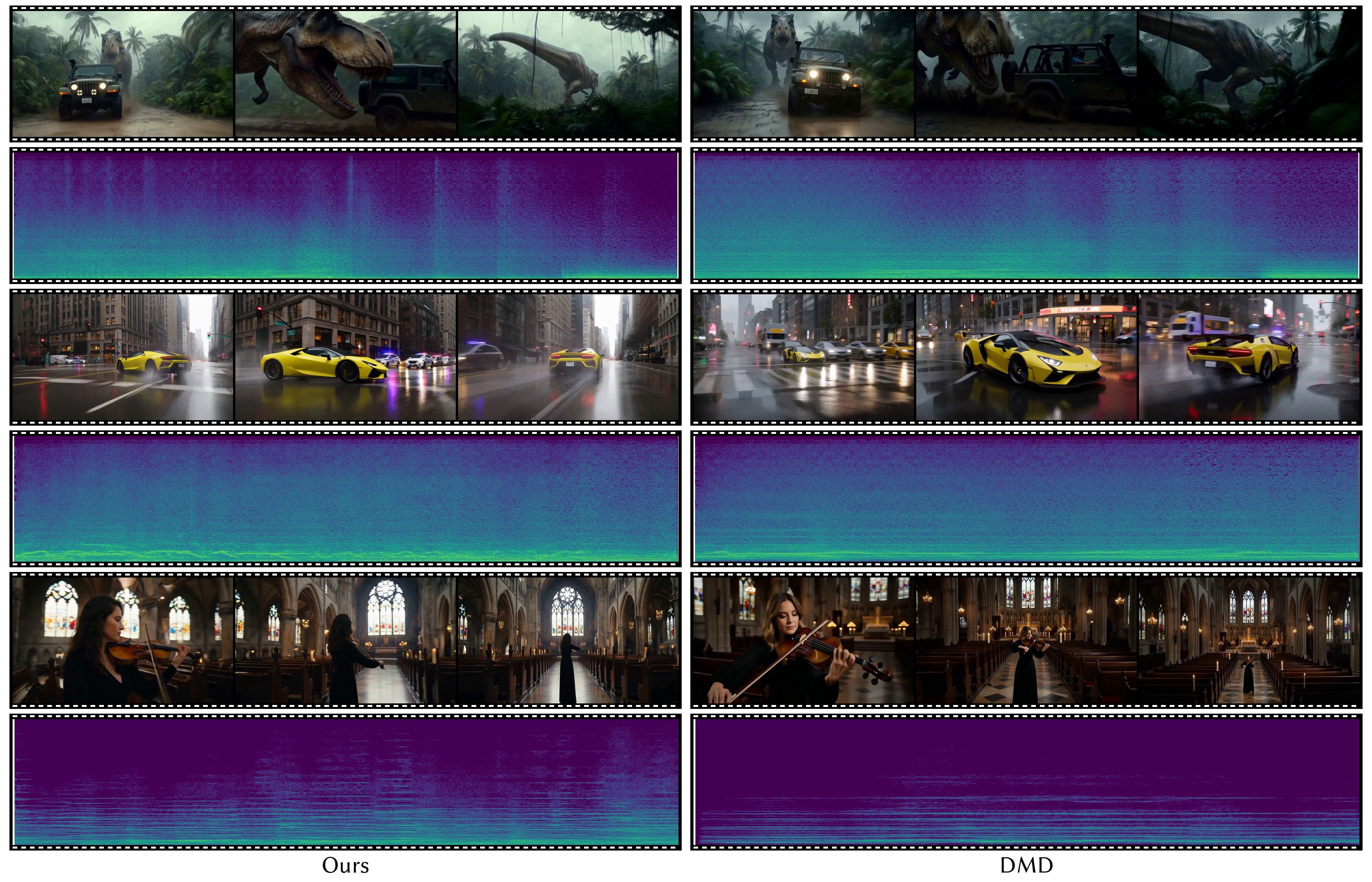}
\caption{\textbf{Joint audio--video generation at 4-NFE.}
SMD and LightX2V-DMD v1.1 are compared under matched settings through video frames and full-clip audio spectrograms.}
\label{fig:h3_t2va_4nfe_audio}
\end{figure}

\subsection{Joint Audio--Video Generation}
\label{sec:fewstep-h3}

We further extend SMD to MiniMax H3~\cite{minimaxh3}, a 33B Transformer that jointly generates video and audio within a shared forward pass.
Unlike image or video generation alone, few-step distillation in this setting must preserve visual dynamics and sound generation simultaneously.
We train text-conditioned students at 4-NFE and 8-NFE by applying the same on-policy objective to both modalities.
We compare with LightX2V-DMD v1.1~\cite{h3turbo} under matched inference budgets on the same 100 evaluation prompts.
We evaluate four video and two audio dimensions (Appendix~\ref{app:h3-evaluation}).

\begin{table}[!tbp]
\caption{Joint audio--video generation with H3 at 4-NFE and 8-NFE on a common evaluation set.
All metrics are higher-is-better. Boldface indicates the larger mean within each inference budget.}
\label{tab:h3_generation}
\centering
\small
\setlength{\tabcolsep}{4pt}
\begin{tabular}{@{}lccccccc@{}}
\toprule
\multirow{2}{*}{Method} & \multirow{2}{*}{NFE} & \multicolumn{4}{c}{Video} & \multicolumn{2}{c}{Audio} \\
\cmidrule(lr){3-6}\cmidrule(l){7-8}
& & Prompt & Structural & Temporal & Naturalness & Clarity & Quality \\
\midrule
LightX2V-DMD v1.1 & 4 & \textbf{3.797} & \textbf{3.653} & 3.644 & 3.556 & 1.829 & 3.133 \\
Ours & 4 & 3.609 & 3.638 & \textbf{3.738} & \textbf{3.638} & \textbf{3.486} & \textbf{3.514} \\
\midrule
LightX2V-DMD v1.1 & 8 & \textbf{3.922} & 3.625 & 3.597 & 3.528 & 3.209 & 3.440 \\
Ours & 8 & 3.863 & \textbf{3.863} & \textbf{3.834} & \textbf{3.625} & \textbf{3.619} & \textbf{3.500} \\
\bottomrule
\end{tabular}
\end{table}

As shown in Table~\ref{tab:h3_generation}, SMD performs favorably across both inference budgets.
LightX2V-DMD achieves higher prompt alignment, whereas SMD consistently improves temporal coherence, visual naturalness, audio clarity, and audio quality.
Structural coherence is comparable at 4-NFE and improves substantially with SMD at 8-NFE.
The audio difference is particularly pronounced at 4-NFE. The DMD baseline often produces more diffuse and muddled spectra, while SMD retains cleaner temporal and frequency structure, consistent with its substantially higher audio clarity and quality scores.
Figures~\ref{fig:h3_t2va_4nfe_audio} and~\ref{fig:h3_t2va_8nfe} provide qualitative comparisons at the two inference budgets.

\begin{figure}[!t]
\centering
\includegraphics[width=\textwidth]{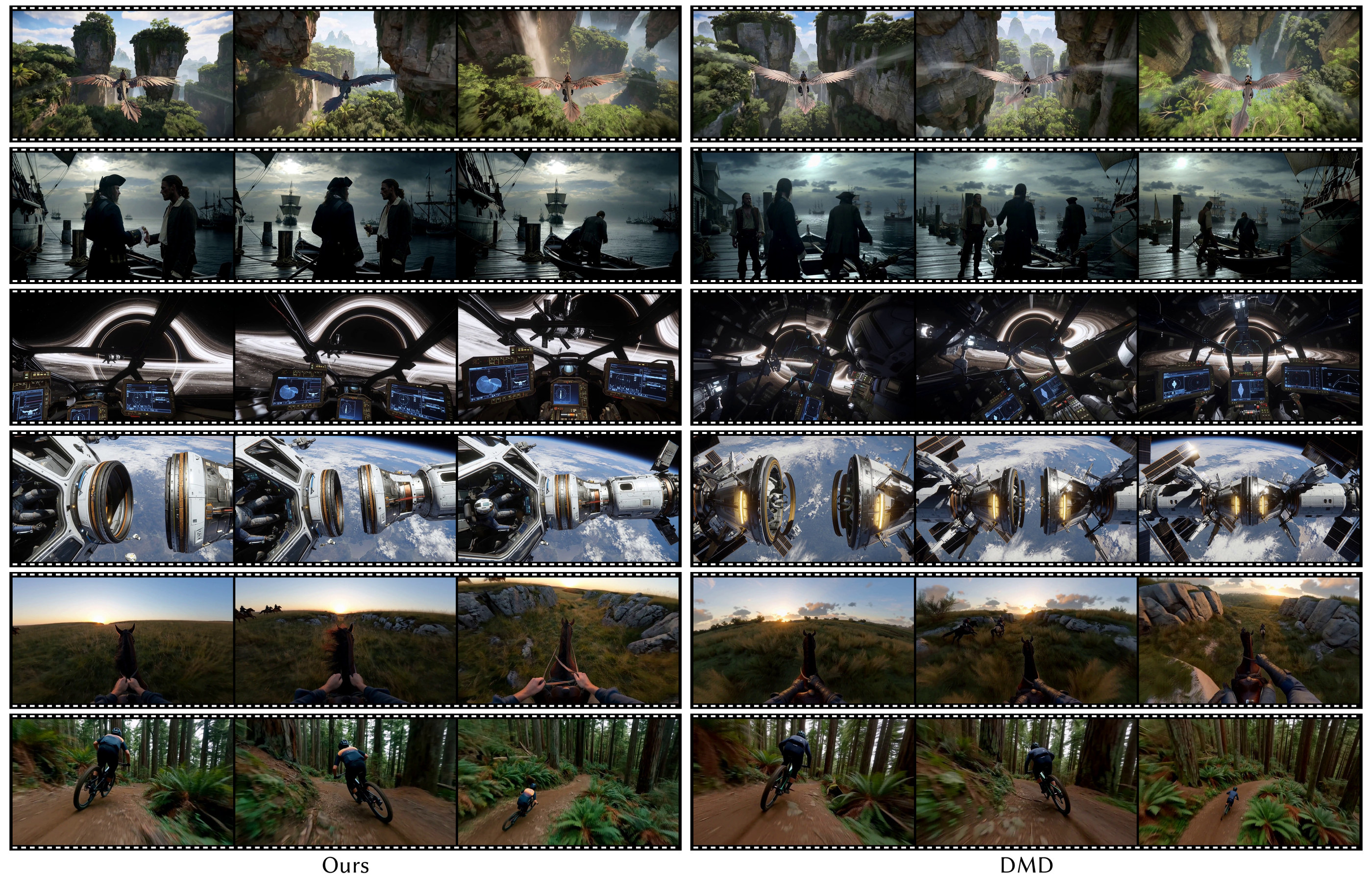}
\caption{\textbf{Joint audio--video generation at 8-NFE.}
Qualitative comparison between SMD and LightX2V-DMD v1.1 under matched prompts, seeds, and timestamps.}
\label{fig:h3_t2va_8nfe}
\end{figure}

\paragraph{Reference-conditioned generation.}
Beyond text-conditioned generation, SMD can also be applied to reference-conditioned H3.
We train with instructions and reference audio--video clips, without requiring ground-truth target outputs.
As shown in Figure~\ref{fig:h3_ref2va}, the resulting 4-NFE student retains a range of reference-conditioned capabilities, including appearance editing, composition, and reference-guided motion and appearance transfer.
We assess this setting qualitatively.

\begin{figure}[!b]
\centering
\includegraphics[width=\textwidth]{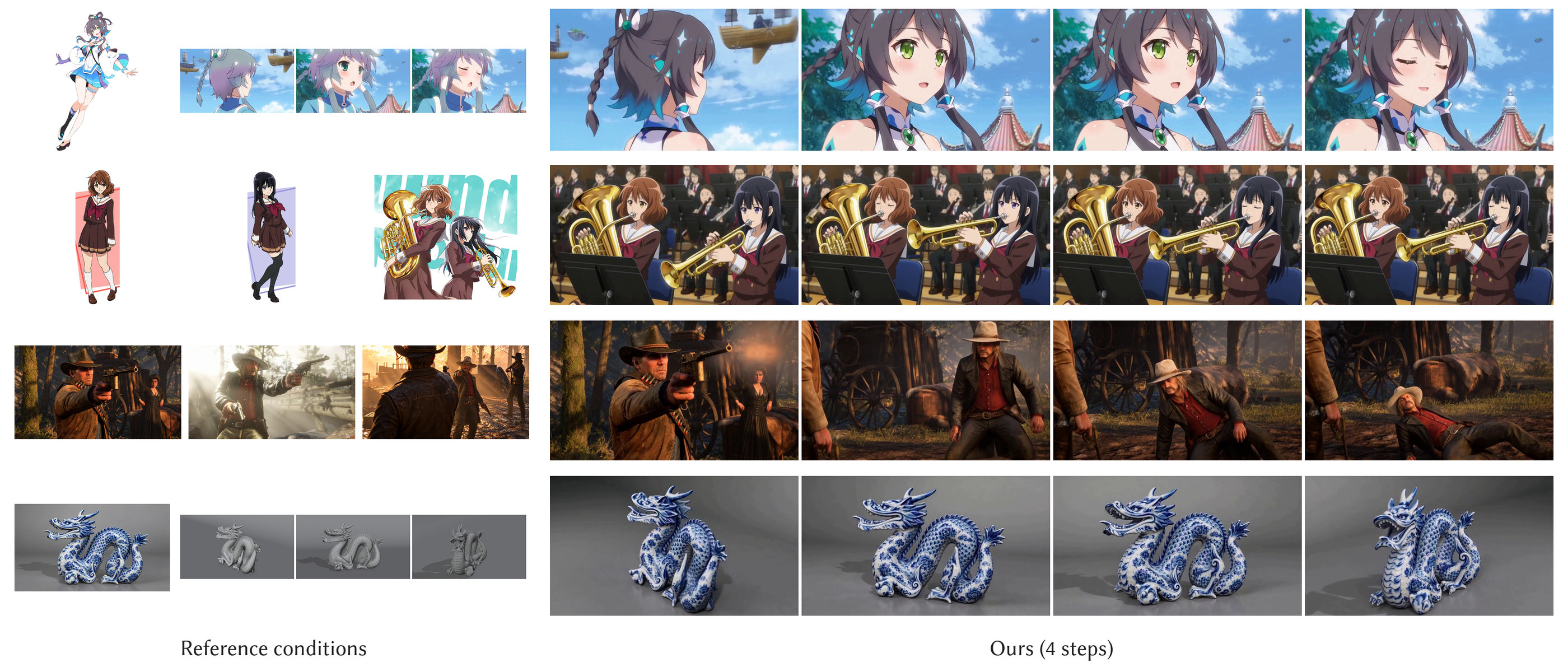}
\caption{\textbf{Reference-conditioned generation at 4-NFE.}
Qualitative results from a single SMD checkpoint across different forms of reference-conditioned generation.}
\label{fig:h3_ref2va}
\end{figure}

\begin{takeaway}
\textbf{Finding 6.}
At 4-NFE, SMD leads the compared image-generation baselines on GenEval and DPG-Bench.
The same objective also improves temporal coherence and audio quality over the DMD baseline in the 33B joint audio--video model.
\end{takeaway}

\endgroup

\section{Beyond Self-Distillation}
\label{sec:beyond}

Velocity matching does not require identical teacher and student architectures. With compatible prediction representations, the supervision need not come from the student's own pretrained model.
We study cross-model and multi-teacher distillation, then causal video compression.

\subsection{Cross-Model Distillation}
\label{sec:beyond-cross-model}

We ask whether a 4-NFE student can benefit from a larger teacher while retaining its own architecture.
For a FLUX.2-klein-4B student, we compare FLUX.2-klein-4B and FLUX.2-klein-9B teachers to test cross-capacity transfer.
For Lens-RL-3.8B~\cite{lens}, we compare self-distillation with supervision from the same 9B FLUX.2 teacher, testing cross-architecture transfer.

\par\medskip\noindent
\begin{minipage}{\textwidth}
\captionsetup{type=table}
\caption{Cross-model distillation at 4-NFE. Bold marks the better teacher for each student and metric.}
\vspace{-0.1in}
\centering
\small
\fitcomparisontable{0.88}{
\begin{tabular}{lllccccccc}
\toprule
Teacher & Student & NFE & GenEval$\uparrow$ & DPG-Bench$\uparrow$ & \multicolumn{5}{c}{OneIG-Bench-EN$\uparrow$} \\ \cmidrule(lr){6-10}
& & & & & Align. & Text & Diversity & Style & Reasoning \\ \midrule
FLUX.2-klein-4B & \multirow{2}{*}{FLUX.2-klein-4B} & 4 & 0.789 & 83.34 & 0.826 & 0.663 & 0.215 & 0.419 & \cb{0.249} \\
FLUX.2-klein-9B & & 4 & \cb{0.791} & \cb{84.26} & \cb{0.843} & \cb{0.744} & \cb{0.224} & \cb{0.441} & 0.246 \\ \midrule
Lens-RL-3.8B & \multirow{2}{*}{Lens-RL-3.8B} & 4 & \cb{0.829} & \cb{87.30} & \cb{0.866} & \cb{0.954} & 0.184 & 0.310 & \cb{0.277} \\
FLUX.2-klein-9B & & 4 & 0.770 & 85.13 & 0.834 & 0.888 & \cb{0.234} & \cb{0.414} & 0.247 \\ \bottomrule
\end{tabular}
}
\label{tab:teacher_student_swap}
\end{minipage}
\par\medskip

\begin{figure}[!t]
\centering
\includegraphics[width=0.96\textwidth]{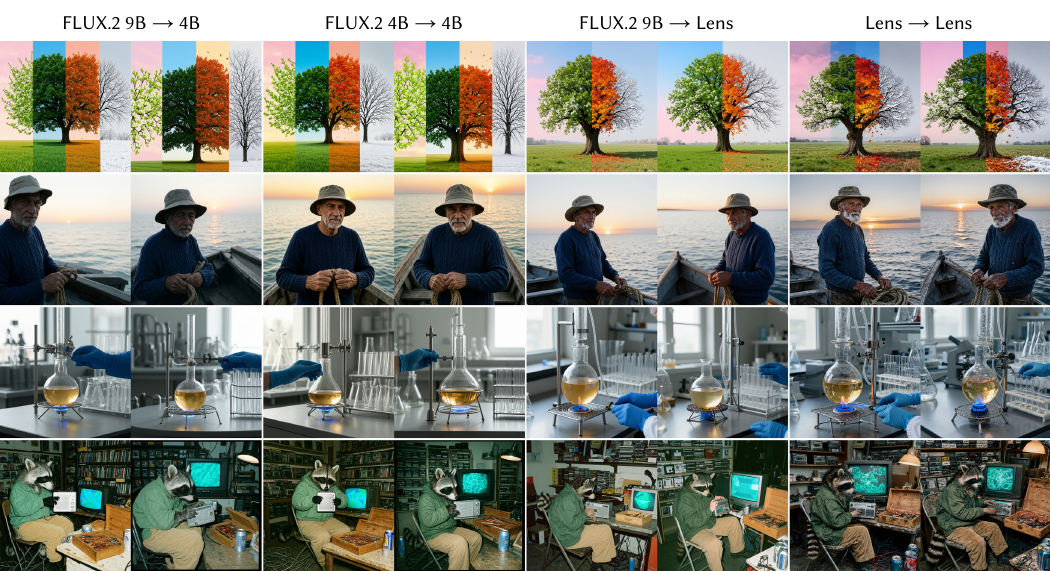}
\caption{\textbf{Changing the teacher changes the student outputs.}
From left to right, each pair of columns shows FLUX.2-klein-9B $\rightarrow$ 4B, FLUX.2-klein-4B $\rightarrow$ 4B, FLUX.2-klein-9B $\rightarrow$ Lens-RL-3.8B, and Lens-RL-3.8B self-distillation.
Each pair contains two 4-NFE generations for the same prompt.}
\label{fig:flux_lens}
\end{figure}

Table~\ref{tab:teacher_student_swap} shows that the 9B teacher improves the FLUX.2-klein-4B student on six of seven reported metrics, with only a small decrease in reasoning.
For Lens, it improves diversity and style but reduces the other reported scores.
These results show that cross-model velocity supervision is feasible, with gains that depend on the teacher-student pairing.
Figure~\ref{fig:flux_lens} compares outputs from these teacher-student pairings.

\subsection{Multiple Experts as Teachers}
\label{sec:beyond-experts}

We next ask whether one student can combine complementary supervision from several sources.
We distill four reward-specialized LoRA teachers for CLIPScore, GenEval, OCR, and PickScore into a single 4-NFE SD3.5-Medium student.
Self-distillation uses the base teacher, each specialist student uses one fixed reward teacher, and routed multi-teacher distillation selects the teacher associated with each prompt category.
All students share the same initialization and use imitation without repulsion.

\begin{figure}[!t]
\centering
\includegraphics[width=0.99\textwidth]{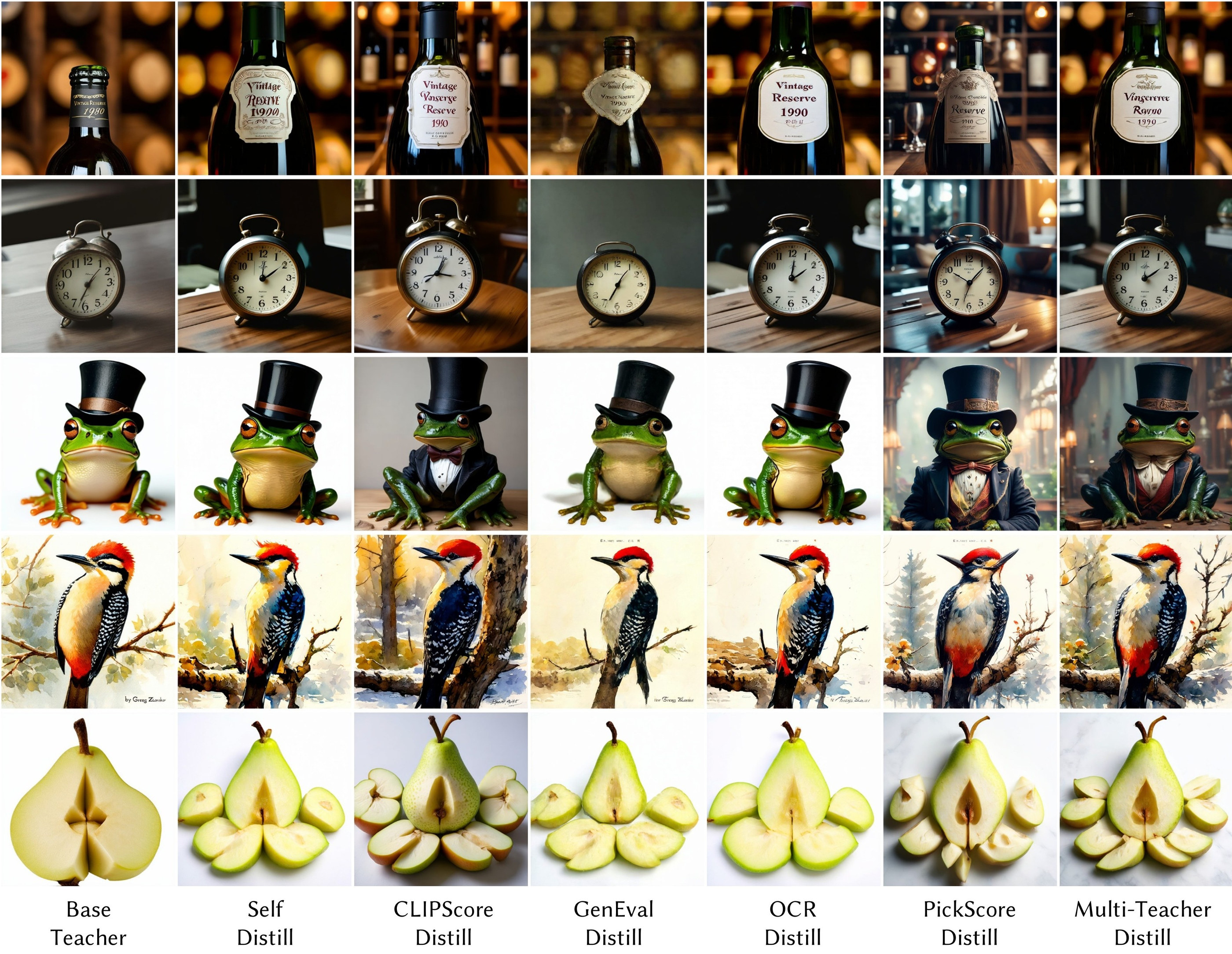}
\caption{\textbf{Multiple experts supervise one student.}
SD3.5-Medium columns show the multi-step base teacher and 4-NFE students trained by self-distillation, four individual reward specialists, or routed multi-teacher supervision.
Each row uses a shared prompt. Routing is training-only.}
\label{fig:sd_multi}
\end{figure}

Table~\ref{tab:teacher_signal} shows that routed multi-teacher distillation improves all six reported metrics over self-distillation, including HPSv2 and ImageReward, which are not direct specialization objectives, and exceeds the original 28-NFE base teacher on every reported metric.
Individual specialists remain strongest on several corresponding criteria, whereas the routed student gives the best OCR score and broadly strong results elsewhere.
Figure~\ref{fig:sd_multi} compares generations from the base teacher and the distilled students.
At inference, only the student is evaluated.

\begin{table}[!b]
\caption{Reward-specialized distillation on SD3.5-Medium. Teacher denotes the original multi-step model. All students use 4-NFE. Bold and underlined scores rank first and second among students.}
\centering
\small
\setlength{\tabcolsep}{6pt}
\renewcommand{\arraystretch}{1.08}
\begin{tabular}{lccccccc}
\toprule
Setting & NFE & GenEval$\uparrow$ & OCR$\uparrow$ & PickScore$\uparrow$ & HPSv2$\uparrow$ & CLIPScore$\uparrow$ & ImageReward$\uparrow$ \\ \midrule
\rowcolor{gray!10}
Teacher & 28 & 0.6279 & 0.5813 & 21.76 & 0.2802 & 0.2837 & 0.8512 \\ \midrule
Self-distill & 4 & 0.6153 & 0.3019 & 21.60 & 0.2848 & 0.2741 & 0.8163 \\
CLIPScore specialist & 4 & 0.7082 & 0.4202 & 22.22 & \clb{0.3055} & \cb{0.3011} & \cb{1.223} \\
GenEval specialist & 4 & \cb{0.7385} & 0.3153 & 21.53 & 0.2658 & 0.2799 & 0.9196 \\
OCR specialist & 4 & 0.6410 & \clb{0.5944} & 21.71 & 0.2841 & 0.2803 & 0.9002 \\
PickScore specialist & 4 & 0.6551 & 0.3738 & \cb{22.81} & \cb{0.3107} & 0.2706 & 1.120 \\
Routed multi-teacher & 4 & \clb{0.7289} & \cb{0.5948} & \clb{22.56} & 0.2995 & \clb{0.2848} & \clb{1.141} \\ \bottomrule
\end{tabular}
\label{tab:teacher_signal}
\end{table}

\subsection{Few-Step Causal Video Generation}
\label{sec:beyond-causal-video}

\begin{figure}[p]
\centering
\includegraphics[width=\textwidth]{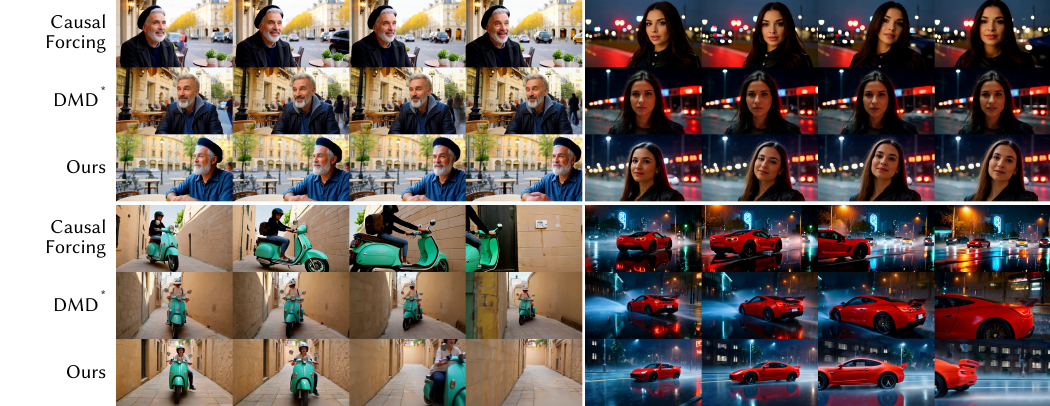}
\caption{\textbf{Replacing the DMD compression stage with on-policy regression.}
Each case shows four chronological frames at 4-NFE for Causal Forcing, DMD$^{*}$, and ours.}
\label{fig:causal_video_main}
\par\vspace{8pt}
\includegraphics[width=\textwidth]{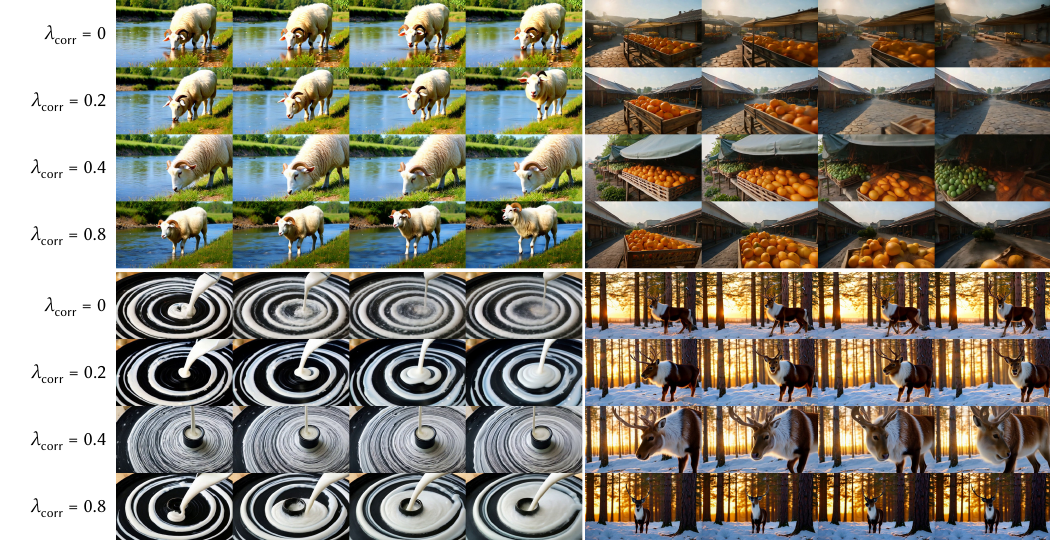}
\caption{\textbf{Correction strength changes the global video solution.}
Four chronological frames are shown for each prompt and correction weight $\lambda_{\mathrm{corr}}$. $\lambda_{\mathrm{corr}}=0$ is pure imitation.
The effect on composition, appearance, and temporal evolution is prompt-dependent and non-monotonic.}
\label{fig:causal_video_correction}
\end{figure}

Fast causal video generation is important when frames must be produced incrementally, including interactive generation and learned world models.
Recent forcing pipelines typically convert a bidirectional video diffusion model into a multi-step causal generator and then use DMD-style distribution matching for few-step compression~\cite{causvid,selfforcing,causalforcing}.
We isolate this final stage and ask whether on-policy regression can replace DMD.
We apply the on-policy regression framework to compress an existing causal Wan2.1-T2V-1.3B generator into a 4-NFE student.

\paragraph{Two teachers with different roles.}
At each query point on the student's rollout, a frozen multi-step causal teacher supplies the local velocity $v_{\mathrm{ar}}$ that the student directly matches.
Because this teacher follows the same temporal factorization as the student, its prediction is compatible with causal inference.
A frozen bidirectional Wan2.1-T2V-14B teacher observes the complete intermediate video state and supplies a correction from its conditional and unconditional predictions.
\begin{equation}
\Delta v_{\mathrm{bi}}
=\operatorname{sg}\!\left(v_{\mathrm{bi}}^{\mathrm{uncond}}-v_{\mathrm{bi}}^{\mathrm{cond}}\right),
\qquad
\widetilde{x}_q=\operatorname{sg}\!\left(x_q+\Delta v_{\mathrm{bi}}\right),
\qquad
\mathcal{L}_{\mathrm{corr}}=\frac{1}{2}\left\|x_q-\widetilde{x}_q\right\|_2^2.
\label{eq:causal_video_correction}
\end{equation}
The pseudo-target contains the detached student state, so the MSE injects the prescribed correction direction rather than fitting a separately generated video.
The causal teacher provides the imitation target and the bidirectional teacher supplies global correction. Neither is used at inference.

\begin{wraptable}{R}{0.49\textwidth}
\centering
\captionsetup{font=small}
\caption{\textbf{Causal video generation.}
The top panel compares with Causal Forcing~\cite{causalforcing}.
The bottom panel varies the correction weight. Bold marks the best in each panel.}
\label{tab:causal_video_distillation}
\small
\setlength{\tabcolsep}{3pt}
\renewcommand{\arraystretch}{1.05}
\textbf{Main comparison}\par\vspace{3pt}
\begin{tabular*}{\linewidth}{@{\extracolsep{\fill}}lccc@{}}
\toprule
Method & Quality & Semantic & Total \\
\midrule
Causal Forcing & 84.59 & \textbf{81.84} & \textbf{84.04} \\
DMD$^{*}$ & 84.15 & 80.82 & 83.48 \\
Ours & \textbf{84.97} & 79.86 & 83.95 \\
\bottomrule
\end{tabular*}
\par\vspace{7pt}
\textbf{Correction sweep}\par\vspace{3pt}
\begin{tabular*}{\linewidth}{@{\extracolsep{\fill}}lccc@{}}
\toprule
$\lambda_{\mathrm{corr}}$ & Quality & Semantic & Total \\
\midrule
$0$ & \textbf{85.36} & 79.64 & \textbf{84.22} \\
$0.2$ & 84.97 & \textbf{79.86} & 83.95 \\
$0.4$ & 84.68 & 79.41 & 83.63 \\
$0.8$ & 84.88 & 79.53 & 83.81 \\
\bottomrule
\end{tabular*}
\end{wraptable}

\paragraph{Comparison with distribution matching.}
We compare against the released Causal Forcing result and DMD$^{*}$, our parameter-efficient control with rank-256 LoRA for both the causal generator and learned fake-score model. $^{*}$ denotes a baseline trained by us.
Our student uses correction weight $\lambda_{\mathrm{corr}}=0.2$ without reference repulsion.
In the top panel of Table~\ref{tab:causal_video_distillation}, it improves Quality and Total over DMD$^{*}$ and comes within $0.09$ of Causal Forcing in Total, without training a fake-score model.
Figure~\ref{fig:causal_video_main} shows the corresponding video comparisons.

\paragraph{Role of bidirectional correction.}
The bottom panel of Table~\ref{tab:causal_video_distillation} isolates the correction signal by fixing the causal-teacher guidance and disabling reference repulsion while varying only $\lambda_{\mathrm{corr}}$.
A moderate weight of $0.2$ gives the best Semantic score in the sweep, but the highest Total is obtained without correction and larger weights do not improve monotonically.
Figure~\ref{fig:causal_video_correction} shows prompt-dependent changes in global appearance as the correction weight varies.

\par\medskip
\reportfinishwrap 

\begin{takeaway}
\textbf{Finding 7.}
On-policy regression supports both cross-model transfer and the consolidation of specialized teachers.
With routed supervision, a single 4-NFE student surpasses the multi-step base teacher on all six reported metrics.
\end{takeaway}

\begingroup
\setlength{\textfloatsep}{10pt plus 2pt minus 2pt}
\setlength{\floatsep}{8pt plus 2pt minus 2pt}
\setlength{\intextsep}{8pt plus 2pt minus 2pt}
\section{Ablation Studies}
\label{sec:ablation}

We ablate SMD on Qwen-Image and Wan2.1-T2V-1.3B, focusing on training prompt coverage, Pivot design, and teacher repulsion.
Training and evaluation details are provided in Appendix~\ref{app:ablation-details}.

\subsection{Prompt Coverage}
\label{sec:ablation-prompt-diversity}

We vary the training prompt category and count when distilling Qwen-Image with SMD.
Each student uses between one and 30K training prompts from Text, Portrait, Objects, or Nature.
A Mixed setting combines all four 30K sets.
Table~\ref{tab:ablation_prompt} reports the complete comparison.

\paragraph{Prompt category and count.}
With 30K training prompts, Text-focused training gives the highest Overall among the four categories, with the largest cross-category differences appearing in text generation.
Figure~\ref{fig:ablation_category} compares outputs across training prompt categories.
Increasing the training set from one to 300 prompts substantially improves the Text scores of Portrait, Objects, and Nature students.
One prompt gives the Text-trained student an Overall of 0.538, versus 0.541 at 30K.

\begin{table}[!htb]
\centering
\small
\setlength{\tabcolsep}{3.5pt}
\renewcommand{\arraystretch}{1.08}
\caption{Qwen-Image training prompt categories and counts. Bold marks each category's best result.}
\label{tab:ablation_prompt}
\begin{tabular*}{\linewidth}{@{\extracolsep{\fill}}llrrrrrr@{}}
\toprule
Category & Prompts & Overall & Align. & Text & Diversity & Style & Reasoning \\
\midrule
Text & 1 & 0.538 & 0.838 & 0.939 & \textbf{0.265} & 0.370 & 0.278 \\
Text & 300 & 0.541 & 0.878 & \textbf{0.966} & 0.134 & 0.420 & 0.305 \\
Text & 3K & 0.540 & \textbf{0.881} & 0.964 & 0.129 & 0.418 & \textbf{0.307} \\
Text & 30K & \textbf{0.541} & 0.876 & 0.961 & 0.150 & \textbf{0.421} & 0.299 \\
\midrule
Portrait & 1 & 0.484 & 0.847 & 0.610 & \textbf{0.310} & 0.411 & 0.239 \\
Portrait & 300 & \textbf{0.540} & 0.886 & \textbf{0.926} & 0.155 & \textbf{0.442} & 0.290 \\
Portrait & 3K & 0.525 & 0.883 & 0.881 & 0.142 & 0.428 & 0.292 \\
Portrait & 30K & 0.537 & \textbf{0.889} & 0.903 & 0.158 & 0.439 & \textbf{0.295} \\
\midrule
Objects & 1 & 0.493 & 0.854 & 0.636 & \textbf{0.303} & 0.431 & 0.242 \\
Objects & 300 & \textbf{0.534} & 0.887 & \textbf{0.912} & 0.134 & \textbf{0.437} & \textbf{0.301} \\
Objects & 3K & 0.533 & \textbf{0.889} & 0.903 & 0.145 & 0.434 & 0.295 \\
Objects & 30K & 0.532 & 0.889 & 0.907 & 0.137 & 0.429 & 0.297 \\
\midrule
Nature & 1 & 0.418 & 0.849 & 0.407 & \textbf{0.253} & 0.344 & 0.237 \\
Nature & 300 & 0.513 & 0.883 & 0.804 & 0.160 & 0.426 & 0.291 \\
Nature & 3K & \textbf{0.519} & \textbf{0.886} & \textbf{0.852} & 0.137 & 0.424 & \textbf{0.298} \\
Nature & 30K & 0.517 & 0.884 & 0.830 & 0.159 & \textbf{0.429} & 0.283 \\
\midrule
Mixed & 120K & 0.538 & 0.885 & 0.938 & 0.138 & 0.432 & 0.300 \\
\bottomrule
\end{tabular*}
\end{table}

\begin{figure}[!htb]
\centering
\includegraphics[width=\textwidth]{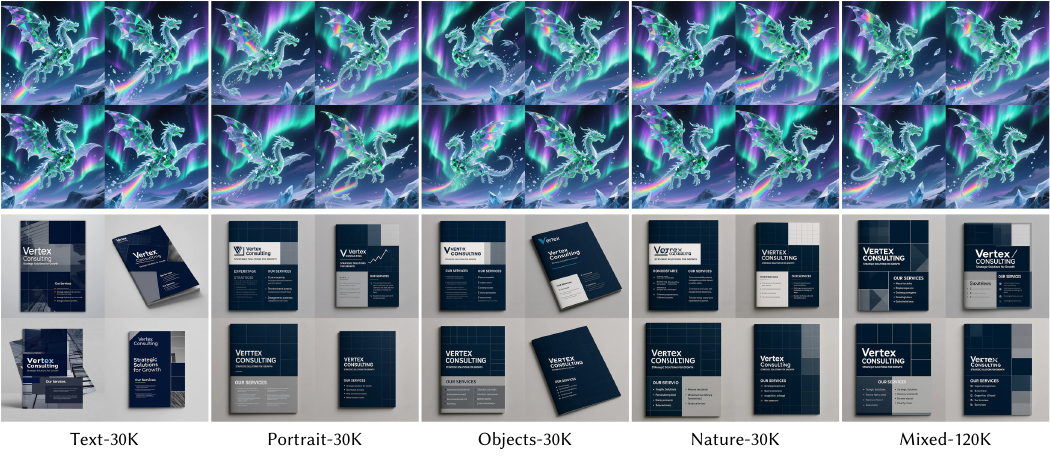}
\caption{\textbf{Category-specific training changes student outputs.}
Qwen-Image students use four 30K-prompt categories or their 120K mixture. Each cell shows four samples for the same test prompt.}
\label{fig:ablation_category}
\end{figure}

\begin{figure}[!htb]
\centering
\includegraphics[width=\textwidth]{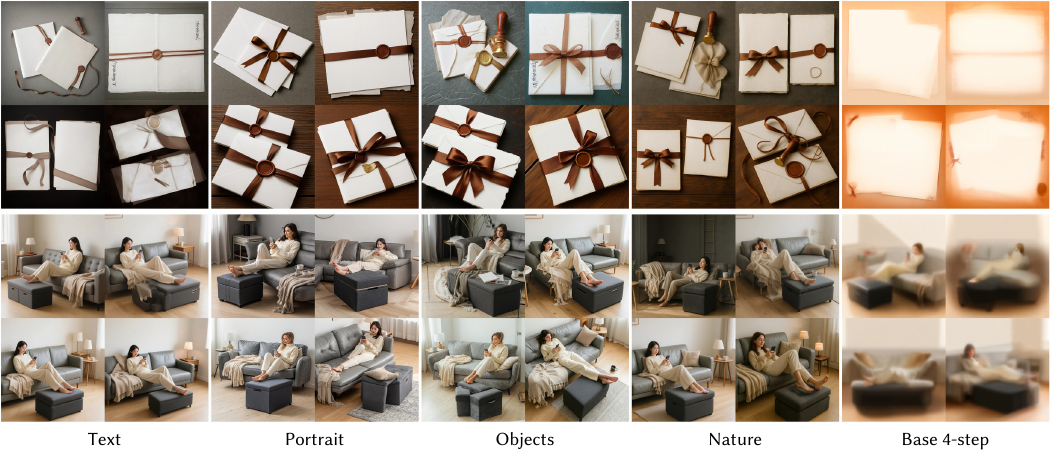}
\caption{\textbf{Generalization beyond a single training prompt.}
Each Qwen-Image student sees only one training prompt but responds to unseen prompts.
Base 4-step shows direct 4-NFE sampling of the pretrained model with its native settings.}
\label{fig:ablation_single_prompt}
\end{figure}

\paragraph{Generation beyond the training prompt.}
The students in Figure~\ref{fig:ablation_single_prompt} still produce content relevant to unseen object and portrait prompts, despite having been distilled on only one prompt.
Single-prompt Wan2.1-T2V-1.3B students also respond to unseen prompts but can retain training-specific visual characteristics. The desert-bloom student's dot-like patterns persist in Figure~\ref{fig:ablation_wan_prompt}.

\begin{figure}[!htb]
\centering
\includegraphics[width=\textwidth]{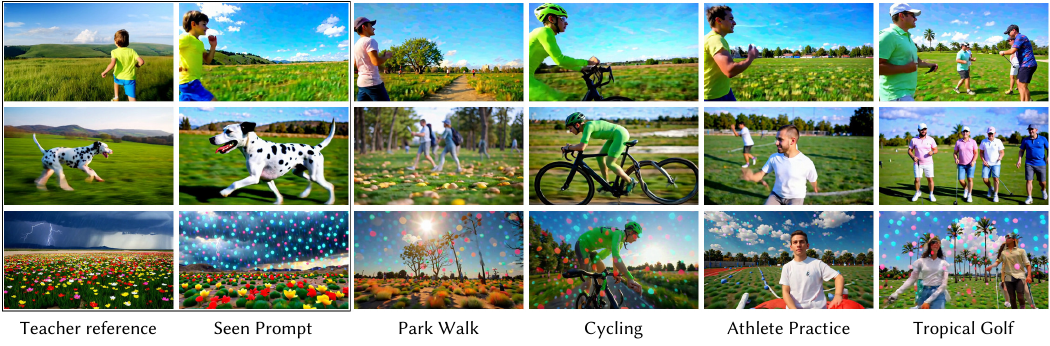}
\caption{\textbf{Visual characteristics can transfer across video prompts.}
Rows show single-prompt SMD students on Wan2.1-T2V-1.3B.
The teacher column provides a visual comparison for each training prompt. Other columns show student outputs for seen and unseen prompts.}
\label{fig:ablation_wan_prompt}
\end{figure}

\begin{takeaway}
\textbf{Finding 8.}
Surprisingly, SMD can work even with a single training prompt.
Both image and video students still respond to unseen prompts in our qualitative comparisons.
\end{takeaway}

\subsection{Pivot Design}
\label{sec:ablation-pivot}

We compare SMD with and without the Pivot residual, then vary Pivot's initialization, the student layer supplying its features, its block count, and its residual basis.
The notation follows Section~\ref{sec:method-pivot}.

\paragraph{Pivot and initialization.}
Table~\ref{tab:ablation_architecture} shows that adding the Pivot residual improves Qwen-Image Overall and Wan2.1-T2V-1.3B Total with either pretrained or random initialization at $K=4$, $M=3$.
The gain therefore does not depend on copying pretrained blocks.

\begin{table}[!htb]
\centering
\small
\setlength{\tabcolsep}{3.5pt}
\renewcommand{\arraystretch}{1.0}
\caption{Pivot design ablations in SMD. Bold marks the best result within each group.}
\label{tab:ablation_architecture}
\label{tab:ablation_pivot_qwen}
\label{tab:ablation_pivot_wan}
\label{tab:ablation_hook}
\textbf{Qwen-Image}
\par\vspace{3pt}
\begin{tabular*}{\linewidth}{@{\extracolsep{\fill}}lrrrrrr@{}}
\toprule
Setting & Overall & Align. & Text & Diversity & Style & Reasoning \\
\midrule
\multicolumn{7}{l}{\textit{Pivot and initialization, $K=4$, $M=3$}} \\
w/o Pivot residual & 0.535 & 0.886 & 0.937 & 0.127 & 0.424 & 0.300 \\
Pretrained & 0.538 & 0.885 & 0.938 & \textbf{0.138} & 0.432 & 0.300 \\
Random & \textbf{0.541} & \textbf{0.887} & \textbf{0.944} & 0.135 & \textbf{0.438} & \textbf{0.303} \\
\midrule
\multicolumn{7}{l}{\textit{Varying $K$, random initialization, $M=3$}} \\
$K=1$ & 0.540 & \textbf{0.889} & 0.943 & 0.130 & 0.433 & \textbf{0.305} \\
$K=4$ & \textbf{0.541} & 0.887 & \textbf{0.944} & \textbf{0.135} & \textbf{0.438} & 0.303 \\
\midrule
\multicolumn{7}{l}{\textit{Varying $K$, random initialization, $M=7$}} \\
$K=1$ & 0.534 & 0.886 & 0.945 & 0.108 & 0.430 & 0.302 \\
$K=8$ & \textbf{0.547} & \textbf{0.891} & \textbf{0.967} & \textbf{0.132} & \textbf{0.434} & \textbf{0.310} \\
\bottomrule
\end{tabular*}
\par\vspace{8pt}
\textbf{Wan2.1-T2V-1.3B}
\par\vspace{3pt}
\begin{minipage}[t]{0.48\linewidth}
\centering
\textit{Pivot and initialization}\par\vspace{3pt}
\begin{tabular*}{\linewidth}{@{\extracolsep{\fill}}lrrr@{}}
\toprule
Setting & Quality & Semantic & Total \\
\midrule
w/o Pivot residual & 84.83 & 76.16 & 83.09 \\
Pretrained & \textbf{85.61} & 78.29 & \textbf{84.14} \\
Random & 85.30 & \textbf{79.28} & 84.10 \\
\bottomrule
\end{tabular*}
\end{minipage}\hfill
\begin{minipage}[t]{0.48\linewidth}
\centering
\textit{Varying $K$, random, $M=3$}\par\vspace{3pt}
\begin{tabular*}{\linewidth}{@{\extracolsep{\fill}}lrrr@{}}
\toprule
$K$ & Quality & Semantic & Total \\
\midrule
1 & 84.80 & 77.34 & 83.31 \\
4 & 85.30 & \textbf{79.28} & \textbf{84.10} \\
8 & \textbf{85.83} & 76.53 & 83.97 \\
\bottomrule
\end{tabular*}
\end{minipage}
\end{table}

\begin{figure}[!htb]
\centering
\includegraphics[width=\textwidth]{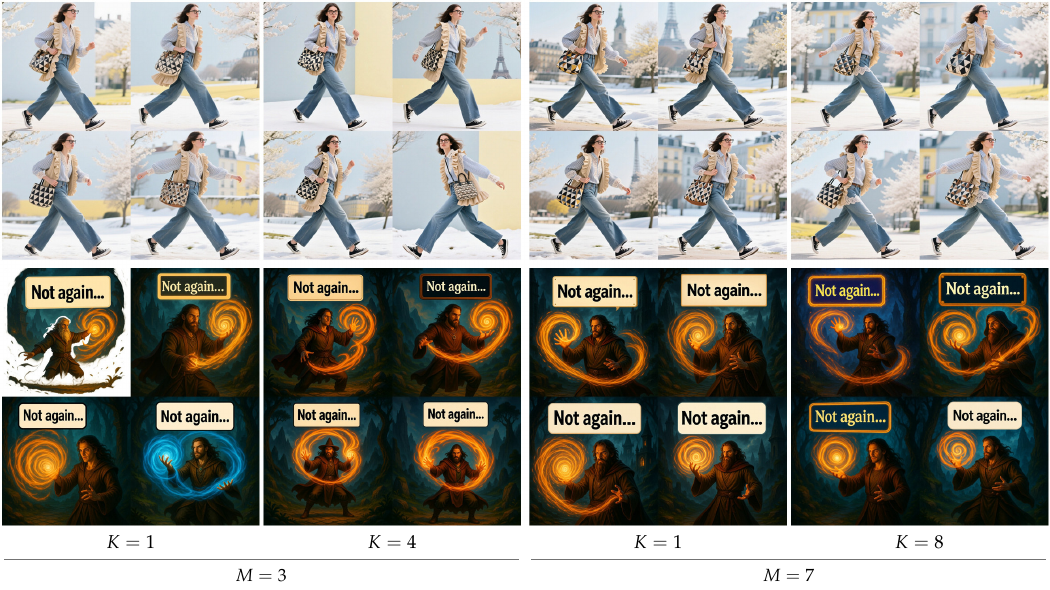}
\caption{\textbf{Changing the student layer used by Pivot.}
Qwen-Image students use $K=1,4$ at $M=3$ and $K=1,8$ at $M=7$.
Each cell contains four samples of the same test prompt.}
\label{fig:ablation_hook}
\end{figure}

\begin{figure}[!htb]
\centering
\includegraphics[width=\textwidth]{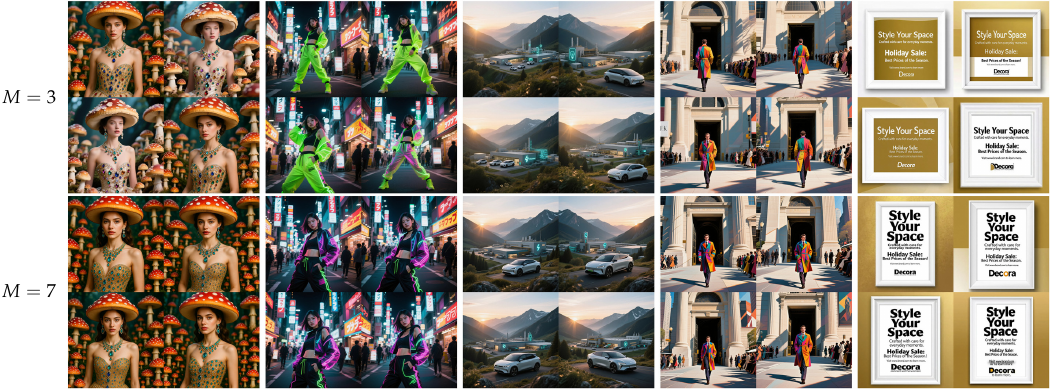}
\caption{\textbf{Varying the number of Pivot blocks.}
Qwen-Image students use $M=3,7$ at $K=1$.
Within each column, the two settings use the same prompt and four seeds.}
\label{fig:ablation_capacity}
\end{figure}

\paragraph{Varying $K$ and $M$.}
For Qwen-Image at $M=7$, using $K=8$ rather than $K=1$ improves all five OneIG dimensions.
For Wan2.1-T2V-1.3B at $M=3$, $K=4$ gives the highest Total (Table~\ref{tab:ablation_architecture}).
For Qwen-Image at $K=1$, increasing Pivot from three to seven blocks instead lowers Overall from 0.540 to 0.534.
In some qualitative examples, larger $K$ also produces more similar compositions across seeds.
Figures~\ref{fig:ablation_hook}--\ref{fig:ablation_wan_hook} illustrate these comparisons.

\begin{figure}[!htb]
\centering
\includegraphics[width=\textwidth]{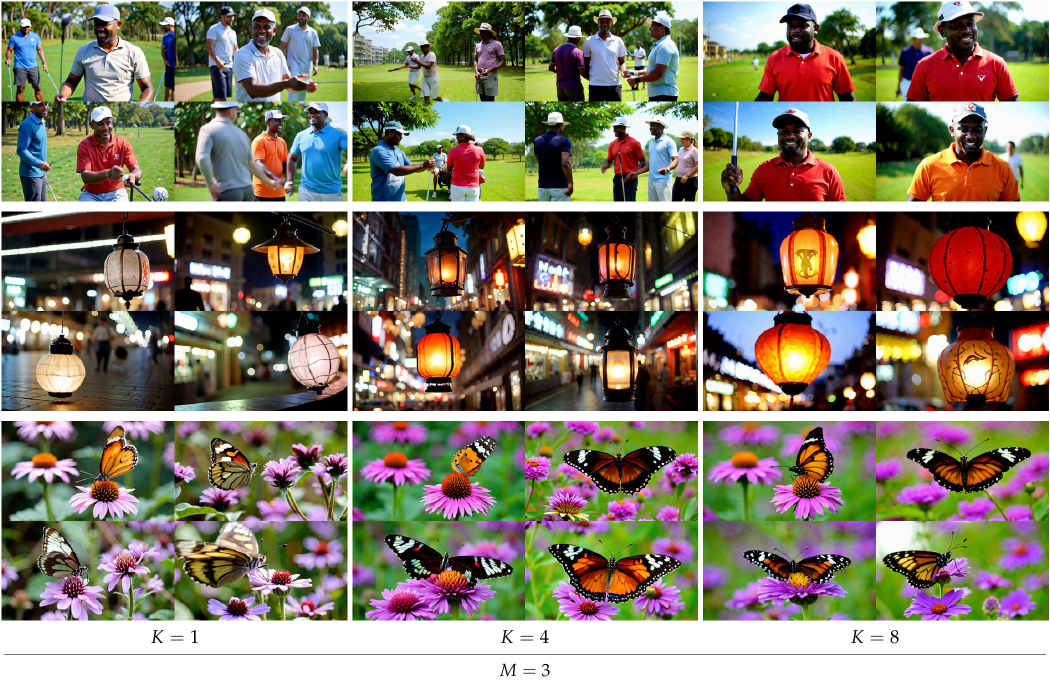}
\caption{\textbf{Changing the student layer used by Pivot.}
Wan2.1-T2V-1.3B students use $K=1,4,8$ with three Pivot blocks. Each cell contains four samples.}
\label{fig:ablation_wan_hook}
\end{figure}

\paragraph{Residual basis.}
With the other Pivot settings fixed, Piecewise ($P=32$) gives the best Qwen-Image Overall, whereas Polynomial ($P=4$) gives the best Wan2.1-T2V-1.3B Total (Table~\ref{tab:ablation_basis}).
Increasing Piecewise from 4 to 32 elements improves Qwen-Image but not Wan2.1-T2V-1.3B.

\begin{table}[!htb]
\centering
\small
\setlength{\tabcolsep}{3.5pt}
\renewcommand{\arraystretch}{1.08}
\setlength{\tabcolsep}{2.5pt}
\caption{Residual basis family and size. The remaining Pivot settings are fixed within each model.}
\label{tab:ablation_basis}
\begin{tabular*}{\linewidth}{@{\extracolsep{\fill}}lcrrrrrrrrr@{}}
\toprule
& & \multicolumn{6}{c}{\textbf{Qwen-Image}} & \multicolumn{3}{c}{\textbf{Wan2.1-T2V-1.3B}} \\
\cmidrule(lr){3-8}\cmidrule(lr){9-11}
Basis & $P$ & Overall & Align. & Text & Diversity & Style & Reasoning & Quality & Semantic & Total \\
\midrule
Polynomial & 4 & 0.538 & 0.885 & 0.938 & 0.138 & \textbf{0.432} & 0.300 & \textbf{85.61} & 78.29 & \textbf{84.14} \\
Spectral & 4 & 0.538 & 0.887 & 0.943 & 0.129 & 0.427 & \textbf{0.307} & 84.91 & 77.76 & 83.48 \\
Collocation & 4 & 0.533 & 0.881 & 0.928 & 0.143 & 0.422 & 0.291 & 84.79 & \textbf{79.73} & 83.78 \\
Piecewise & 4 & 0.540 & 0.884 & 0.944 & \textbf{0.145} & 0.429 & 0.300 & 85.20 & 78.56 & 83.87 \\
Piecewise & 32 & \textbf{0.543} & \textbf{0.889} & \textbf{0.962} & 0.132 & 0.426 & 0.306 & 84.59 & 78.31 & 83.33 \\
\bottomrule
\end{tabular*}
\end{table}

\begin{takeaway}
\textbf{Finding 9.}
Different residual bases give relatively similar aggregate scores.
Larger $K$ can improve prompt alignment, but some qualitative examples show less varied compositions.
\end{takeaway}

\subsection{Teacher Repulsion}
\label{sec:ablation-objective}

We vary the reference CFG scale $w$ and repulsion weight $\lambda$ while keeping the target teacher prediction and Pivot fixed.
The comparison includes unconditional ($w=0$) and conditional ($w=1$) references, as well as SMD without repulsion ($\lambda=0$).

\begin{figure}[!htb]
\centering
\includegraphics[width=\textwidth]{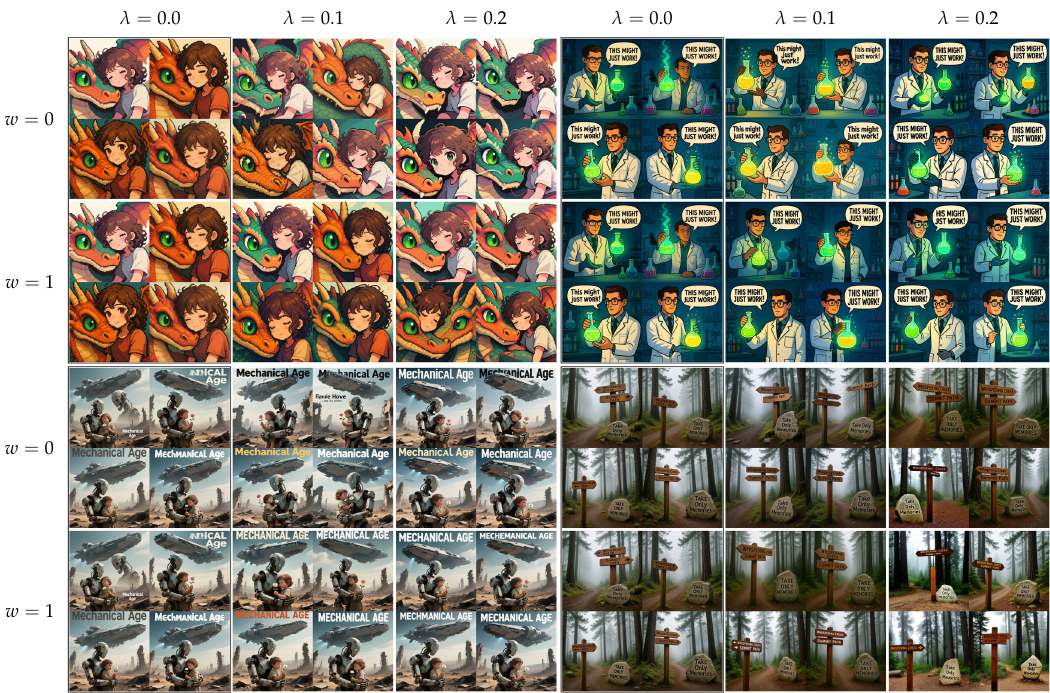}
\caption{\textbf{Reference and weight ablations in SMD.}
Qwen-Image rows vary $w$ and columns vary $\lambda$.
The boxed $\lambda=0$ cells repeat the same samples without repulsion.}
\label{fig:ablation_repulsion}
\end{figure}

\begin{table}[!htb]
\centering
\small
\setlength{\tabcolsep}{4pt}
\renewcommand{\arraystretch}{1.10}
\caption{Reference CFG scale and repulsion weight in SMD.}
\label{tab:ablation_repulsion_all}
\label{tab:ablation_repulsion_qwen}
\label{tab:ablation_repulsion_wan}
\textbf{Qwen-Image}\par\vspace{3pt}
\begin{tabular*}{\linewidth}{@{\extracolsep{\fill}}lccccc@{}}
\toprule
& w/o repulsion & \multicolumn{2}{c}{$w=0$} & \multicolumn{2}{c}{$w=1$} \\
\cmidrule(lr){2-2}\cmidrule(lr){3-4}\cmidrule(lr){5-6}
Metric & $\lambda=0$ & $\lambda=0.1$ & $\lambda=0.2$ & $\lambda=0.1$ & $\lambda=0.2$ \\
\midrule
Overall & 0.539 & 0.539 & \textbf{0.542} & 0.528 & 0.538 \\
Align. & 0.883 & \textbf{0.888} & 0.886 & 0.883 & 0.885 \\
Text & 0.943 & \textbf{0.960} & 0.958 & 0.899 & 0.938 \\
Diversity & \textbf{0.142} & 0.123 & 0.139 & 0.138 & 0.138 \\
Style & \textbf{0.434} & 0.429 & 0.421 & 0.425 & 0.432 \\
Reasoning & 0.295 & 0.298 & \textbf{0.304} & 0.294 & 0.300 \\
\bottomrule
\end{tabular*}
\par\vspace{8pt}
\textbf{Wan2.1-T2V-1.3B}\par\vspace{3pt}
\begin{tabular*}{\linewidth}{@{\extracolsep{\fill}}lccccccc@{}}
\toprule
& w/o repulsion & \multicolumn{3}{c}{$w=0$} & \multicolumn{3}{c}{$w=1$} \\
\cmidrule(lr){2-2}\cmidrule(lr){3-5}\cmidrule(lr){6-8}
Metric & $\lambda=0$ & $\lambda=0.1$ & $\lambda=0.2$ & $\lambda=0.4$ & $\lambda=0.1$ & $\lambda=0.2$ & $\lambda=0.4$ \\
\midrule
Quality & 84.97 & 85.61 & \textbf{85.83} & 85.24 & 84.05 & 84.66 & 84.92 \\
Semantic & 77.62 & \textbf{78.29} & 77.59 & 77.59 & 77.07 & 75.59 & 78.21 \\
Total & 83.50 & 84.14 & \textbf{84.18} & 83.71 & 82.65 & 82.85 & 83.57 \\
\bottomrule
\end{tabular*}
\end{table}

An unconditional reference with $\lambda=0.2$ improves over SMD without repulsion and gives the best Qwen-Image Overall and Wan2.1-T2V-1.3B Total in these sweeps (Table~\ref{tab:ablation_repulsion_all}).
The conditional reference does not give the same benefit, and increasing the unconditional weight to 0.4 reduces Wan2.1-T2V-1.3B Total.
Figures~\ref{fig:ablation_repulsion} and~\ref{fig:ablation_wan_repulsion} show the corresponding appearance changes.

\begin{figure}[!htb]
\centering
\includegraphics[width=\textwidth]{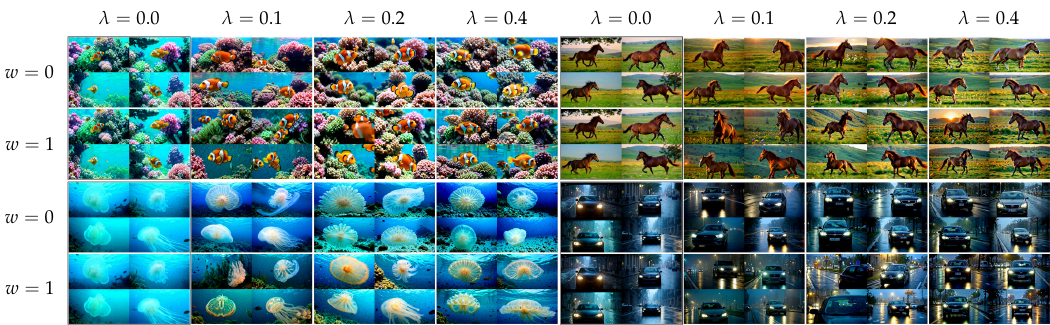}
\caption{\textbf{Teacher-repulsion ablation on Wan2.1-T2V-1.3B.}
Wan2.1-T2V-1.3B uses the same layout, with three nonzero weights per reference.}
\label{fig:ablation_wan_repulsion}
\end{figure}

\begin{takeaway}
\textbf{Finding 10.}
The teacher's unconditional prediction provides an effective weak reference for repulsion on both tested backbones.
Moderate repulsion improves the results, but increasing its weight further can weaken the benefit.
\end{takeaway}
\endgroup

\section{Related Work}
\label{sec:related}

Few-step generation from diffusion and flow-based models has attracted broad research interest. Existing approaches largely fall into trajectory-level and distribution-level paradigms. Beyond these two main lines, we survey methods that bridge or extend both paradigms, along with other complementary acceleration techniques. Finally, we discuss the emerging connections between few-step distillation and reinforcement learning for diffusion models.

\paragraph{Trajectory-Level Distillation.}
Trajectory-level methods train a student to reproduce the teacher's sampling trajectory in fewer steps. The earliest approach, progressive distillation~\cite{progressivedistill}, repeatedly halves the step count by training students to match teacher predictions over merged intervals. Guidance distillation~\cite{guidancedistillation} later folded classifier-free guidance into the student, removing the need for dual forward passes at inference. Both can degrade at very few steps, motivating subsequent objectives.

Consistency models~\cite{consistencymodels} introduced a self-consistency constraint requiring that any point on the same probability-flow ODE trajectory map to the same clean output, allowing both distillation from a pretrained teacher and standalone training. Follow-up work improved stability via better schedules and loss weighting~\cite{improvedconsistency}, enabled efficient fine-tuning from pretrained diffusion models~\cite{ect}, moved the formulation to latent spaces~\cite{lcm, ilcm, lcmlora}, generalized it to multi-step generation~\cite{multistepcm, pcm, tcm, hypersd, dcm, t2vturbo}, and developed continuous-time variants that remove discretization dependence~\cite{scm, sanasprint, rcm}. Target-driven timestep selection~\cite{tdd} further improves consistency distillation by optimizing which timestep pairs to train on. A concurrent report~\cite{damot2i} offers practical guidance on consistency distillation.

Flow-map methods establish the mathematical foundations connecting flow maps to stochastic interpolants~\cite{flowmapmatching, scot}, learning solution operators that transport states between arbitrary time pairs rather than mapping only to the clean endpoint. Consistency trajectory models~\cite{ctm} first parameterized the full ODE solution. MeanFlow~\cite{meanflow} derived a self-consistency identity for the time-averaged velocity that enables one-step generation. Extensions cover objective decompositions~\cite{splitmeanflow}, JVP-free formulations~\cite{modularmeanflow, decoupledmeanflow}, continuous-time unification with flow matching~\cite{alignyourflow, alphaflow, soflow}, data-free training~\cite{freeflow, boot}, and recursive formulations that support any-step generation from a single model~\cite{rcgm, facm}. Another line of trajectory-level distillation follows rectified flow~\cite{rectifiedflow}, where methods such as InstaFlow~\cite{instaflow} and PeRFlow~\cite{perflow} straighten ODE trajectories for fewer-step integration. OPSD-V~\citep{liu2026opsd} uses real long videos as privileged temporal context to construct cleaner teacher caches, providing dense denoising-level supervision along the student's on-policy rollout for post-training few-step autoregressive video generators.

$\pi$-Flow and PDD are particularly close to our regression-based formulation~\cite{piflow,pdd}.
$\pi$-Flow predicts a network-free policy that supplies time-varying velocities for integration over substeps within one network evaluation.
PDD predicts multiple denoising updates in parallel, and its output heads can be fused for inference.
SMD instead places within-interval residual capacity in a training-only branch whose basis functions integrate to zero.
The deployed student keeps its original head and requires neither Pivot nor substep integration.

\paragraph{Distribution-Level Distillation.}
Distribution-level methods directly compare the student's generated distribution with the teacher or data distribution.
Their training objectives can draw on discriminator feedback, score estimates, or sample statistics.

Adversarial distillation draws on the generative adversarial network~\cite{gan, stylegan, stylegan2, stylegan3} paradigm, using learned discriminators to provide distribution-level feedback. ADD~\cite{add} combined score distillation with adversarial losses for few-step generation, LADD~\cite{ladd} moved discrimination to latent space using generative features from pretrained diffusion models, and progressive adversarial schemes~\cite{sdxllightning} interpolate between trajectory and adversarial objectives across distillation stages. Further developments include GAN-diffusion hybrids~\cite{ufogen, flashdiff}, post-training adversarial fine-tuning for video~\cite{apt, osv, phasevideo}, adversarial distillation for 3D generation~\cite{cad}, intermediate-state adversarial matching~\cite{advdm}, and controllability-preserving distillation~\cite{jdm, unidad}.

Score-based distribution matching emerged as an alternative that avoids explicit discriminator training. Score distillation sampling~\cite{dreamfusion} first used a pretrained diffusion model's score to optimize parametric representations for text-to-3D generation. Variational score distillation~\cite{vsd} refined this by introducing a learned approximation to the variational distribution's score, reducing over-smoothing. Diff-Instruct~\cite{diffinstruct} generalized this mathematical framework to training arbitrary generators from pretrained diffusion models, deriving the integral KL objective and training an auxiliary score network on student samples. Distribution Matching Distillation (DMD)~\cite{dmd} scaled this approach to text-to-image synthesis, and DMD2~\cite{dmd2} improved optimization stability and multi-step extensions. SwiftBrush~\cite{swiftbrush, swiftbrushv2} demonstrated that pure variational score distillation, without adversarial training, suffices for one-step synthesis that surpasses the teacher.

Extensions include decoupling CFG internalization from distribution matching~\cite{ddmd}, separating signals across noise levels~\cite{flashdmd, phaseddmd}, hierarchical structure-to-detail decomposition~\cite{hierarchdmd}, removing the auxiliary network via flow-map surrogates~\cite{dmdwofake}, scaling to flow architectures~\cite{senseflow, cdm}, self-adversarial flow formulations~\cite{twinflow}, video adaptation through transition and score-gradient matching~\cite{tmd, sgmd}, adaptive matching~\cite{amd}, diversity-preserving objectives~\cite{dpdmd}, and teacher-feature drifting~\cite{tfd}.

Alternatives to KL include Fisher divergence via score identity distillation~\cite{sid, fewstepsid}, implicit matching~\cite{sim}, $f$-divergence weighting~\cite{fdmd}, proper scoring rules~\cite{varflow}, EM distillation~\cite{emd}, and moment matching approaches~\cite{imm, mmd}. A recent line of work applies DMD-style streaming distillation to convert bidirectional video diffusion models into fast autoregressive generators~\cite{causvid, causalforcing, causalppforcing, selfforcing, rewardforcing}.

Score-regularized continuous-time consistency~\cite{rcm} combines consistency training with a lightweight DMD-style correction. Trajectory distribution matching~\cite{tdm} evaluates divergences along intermediate states. Video-specific designs balance coverage and sharpness~\cite{modeseekingmeetsmean, anyflow, magicdistill, dfd}. Scale-wise distillation~\cite{swd} decomposes generation by spatial frequency.

\paragraph{Other Distillation and Acceleration Techniques.}
Other approaches connect few-step generation with step-size conditioning, adaptive trajectories, and self-training.
Shortcut models~\cite{shortcuts} learn step-size-conditioned predictions that interpolate between local velocity estimation and global one-step generation.
Instance-aware adaptive trajectories~\cite{rayflow} learn per-sample flow paths, while ArcFlow~\cite{arcflow} treats policy-based generation as an imitation learning problem.
Other work studies training one-step diffusion models without teacher distillation~\cite{wodmd} and accelerating diffusion models with a single sample~\cite{glance}.

Complementary acceleration techniques further reduce inference cost. Sampling methods and schedules improve efficiency through fast scheduling~\cite{fastdpm}, analytic variance estimation~\cite{analyticdpm}, pseudo-numerical methods~\cite{pnmd}, exponential integrators~\cite{dpmsolver}, unified predictor-corrector frameworks~\cite{unipc}, and optimized timestep placement~\cite{alignyoursteps, jumpyoursteps}. Caching exploits temporal redundancy. DeepCache~\cite{deepcache} skips redundant U-Net computations, T-GATE~\cite{tgate} decomposes and caches attention outputs at scheduled timesteps, and token-level caching~\cite{toca}, spectral-evolution policies~\cite{seacache}, and sensitivity-based strategies~\cite{sencache} target transformer architectures. For video, pruning with adaptive routing~\cite{pare} and spectral progressive generation~\cite{spectral} exploit inter-frame redundancy. FreeU~\cite{freeu} re-weights skip connections without retraining, while FreeInit~\cite{freeinit} refines initial noise via low-frequency iteration. Architectural acceleration through sparse attention~\cite{vsa} and sliding-tile attention~\cite{fast} exploits spatiotemporal locality.

\paragraph{Connections to Reinforcement Learning.}
Preference alignment and reinforcement learning have become important components in the training of modern large generative models. Recently, many works have applied RL methods such as GRPO~\cite{grpo} and DPO~\cite{dpo} to diffusion models through the reverse process~\cite{flowgrpo, dgpo, flowopd, diffusionopd} and the forward process~\cite{awm, diffusionnft, ram}. Human preference models~\cite{imagereward, visionreward, videoreward, pickscore, hps} provide the reward signals in these settings.

A growing body of work combines few-step distillation with diffusion RL. Unified frameworks augment distribution matching with non-differentiable rewards~\cite{tdmr1, dmdr, advdmd, reinforcefew}, and the distribution matching objective itself can be reconceptualized as a reward signal for distilled generators~\cite{Rdm}.

\section{Conclusion}
\label{sec:conclusion}
Strong few-step diffusion distillation need not rely on auxiliary score models, adversarial training, or specialized inference architectures.
SMD builds on simple on-policy velocity regression with a training-only Pivot proxy for local velocity variation and repulsion from weaker teacher predictions.
Across diverse generative tasks and models with up to 33B parameters, SMD delivers strong few-step performance while leaving student inference unchanged.
At 4-NFE, it outperforms strong few-step baselines and surpasses multi-step teachers on several benchmarks.
Our experiments further extend SMD to cross-model, multi-teacher, and causal-video distillation, demonstrating the applicability of on-policy regression beyond self-distillation.


\begingroup
\setlength{\columnsep}{18pt}
\renewcommand{\bibfont}{\small\raggedright}
\setlength{\bibsep}{1.5pt plus 0.5pt}
\renewcommand{\bibsection}{}
\let\reportbibliography\thebibliography
\renewcommand{\thebibliography}[1]{%
  \reportbibliography{#1}%
  \clubpenalty=10000\widowpenalty=10000\relax
}
\begin{multicols}{2}[\section*{\refname}]
\bibliography{report/references}
\bibliographystyle{unsrtnat}
\end{multicols}
\endgroup

\clearpage
\appendix
\reportappendixtrue
\appendixtableofcontents
\label{sec:appendix-contents}
\newpage

\section{Supplementary Analysis and Mathematical Details}
\label{sec:appendix}

\subsection{Residual Parameterizations}
\label{app:residual_basis}

The four parameterizations compared in Table~\ref{tab:ablation_basis} give Pivot different ways to bend a training trajectory while preserving the coarse transition used at inference.

\begin{figure}[!t]
    \centering
    \includegraphics[width=\textwidth]{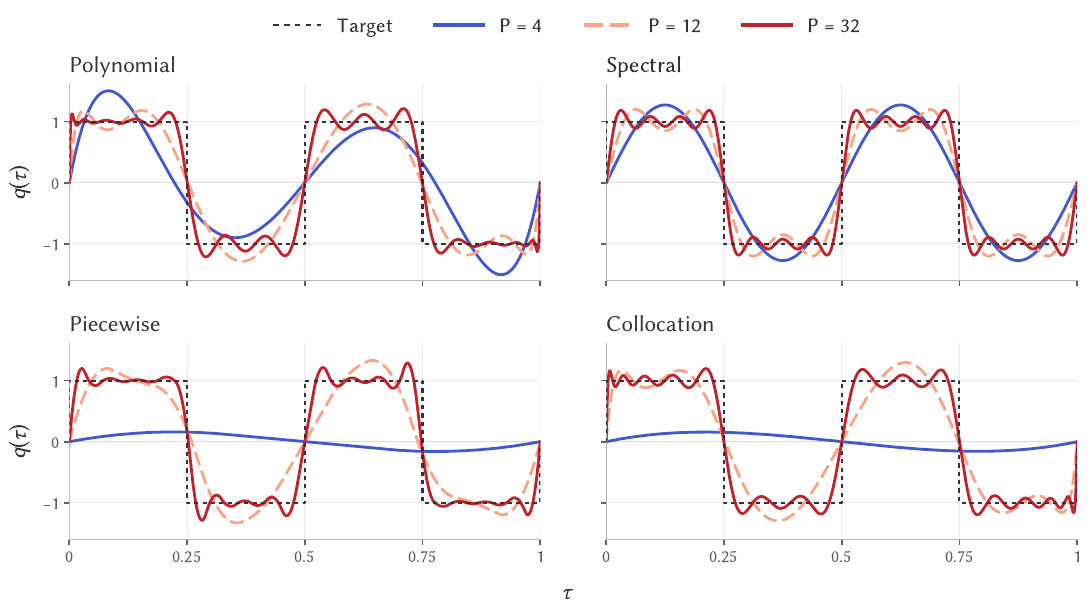}
    \caption{\textbf{Endpoint-preserving path approximation.}
    Least-squares fits to the square-wave target in Eq.~\eqref{eq:residual_fitting_target}, using $P\in\{4,12,32\}$ coefficients.
    All fitted paths preserve both endpoints.}
    \label{fig:residual_basis_fitting}
\end{figure}

\paragraph{From a velocity residual to a path correction.}
Recall that $\tau\in[0,1]$ measures progress through a segment of duration $\Delta$.
Pivot predicts $P$ coefficient vectors $a_m$, each with the same shape as the student velocity, and holds them fixed within the segment.
The basis functions control how these coefficients contribute along the interval.
\begin{equation}
    r_\phi(\tau)=\sum_{m=1}^{P}a_m b_m(\tau),
    \qquad
    q_\phi(\tau)=\sum_{m=1}^{P}a_m I_m(\tau),
    \qquad
    I_m(\tau)=\int_0^\tau b_m(u)\,du.
    \label{eq:appendix_residual_path}
\end{equation}
The residual $r_\phi$ modifies velocity, and its integral $q_\phi$ corrects the straight-line path.
\begin{equation}
    v(\tau)=v_\theta+r_\phi(\tau),
    \qquad
    x_\tau=x_t-\Delta\bigl[\tau v_\theta+q_\phi(\tau)\bigr].
    \label{eq:appendix_corrected_state}
\end{equation}
Thus the actual state displacement relative to the straight line is $-\Delta q_\phi(\tau)$.
Requiring every $b_m$ to have zero integral gives
\begin{equation}
    \int_0^1 b_m(\tau)\,d\tau=0
    \quad\Longrightarrow\quad
    q_\phi(0)=q_\phi(1)=0
    \quad\Longrightarrow\quad
    x_{\tau=1}=x_t-\Delta v_\theta.
    \label{eq:appendix_endpoint_constraint}
\end{equation}
The residual can therefore change intermediate states without moving either endpoint for a fixed $v_\theta$.
The constraint concerns the integral, so the residual need not vanish at the endpoints.
Figure~\ref{fig:residual_basis_fitting} compares how the four constructions fit the same challenging path correction as the coefficient budget increases.
The controlled fitting protocol is described below.

\paragraph{Polynomial basis.}
We can preserve endpoints using functions that vanish at both ends.
\begin{equation}
    B_m(\tau)=\tau(1-\tau)(2\tau-1)^{m-1},
    \qquad
    b_m(\tau)=\frac{dB_m(\tau)}{d\tau},
    \quad 1\le m\le P.
    \label{eq:polynomial_path_basis}
\end{equation}
The factor $\tau(1-\tau)$ anchors the path correction to zero at both endpoints. Successive powers of $2\tau-1$ add polynomial variation within the interval.
Since $B_m(0)=0$, the integrated basis is simply $I_m=B_m$ and $q_\phi=\sum_m a_mB_m$.
Differentiating this path gives $r_\phi$, with zero integral because $B_m(1)-B_m(0)=0$.
For example, the first mode has $B_1(\tau)=\tau(1-\tau)$ and $b_1(\tau)=1-2\tau$. Its velocity contribution changes sign halfway through the segment, producing a single path excursion that returns to zero.
With $P$ coefficients, this construction spans $P$ independent polynomial modes.

\paragraph{Spectral basis.}
Instead of constructing the path first, we can directly choose velocities whose positive and negative contributions cancel.
\begin{equation}
    b_m(\tau)=\cos(m\pi\tau),
    \qquad
    I_m(\tau)=\frac{\sin(m\pi\tau)}{m\pi},
    \quad 1\le m\le P.
    \label{eq:spectral_basis}
\end{equation}
Each $b_m(\tau)$ integrates to zero on $[0,1]$.
Low indices describe broad changes across the segment, while higher indices allow more rapid oscillations.
Integration reduces the amplitude of mode $m$ by $1/(m\pi)$, so equal velocity coefficients do not produce equal path-displacement amplitudes.
The constant mode $m=0$ is excluded because it would change the coarse endpoint.
The remaining $P$ cosine modes are mutually orthogonal and independent.

\paragraph{Centered nodal interpolation.}
The remaining two designs predict values $a_m$ at $P\ge2$ nodes $\tau_m$.
Let $\ell_m(\tau)$ be interpolation functions satisfying $\ell_m(\tau_k)=\delta_{mk}$.
The coefficients are the nodal values of an \emph{uncentered} velocity interpolant $\widetilde r_\phi(\tau)=\sum_m a_m\ell_m(\tau)$.
To remove its interval mean, define
\begin{equation}
    w_m=\int_0^1\ell_m(\tau)\,d\tau,
    \qquad
    b_m(\tau)=\ell_m(\tau)-w_m.
    \label{eq:centered_nodal_basis}
\end{equation}
The resulting residual is $r_\phi(\tau)=\widetilde r_\phi(\tau)-\mu_\phi$, where $\mu_\phi=\sum_m w_ma_m$ is the mean of $\widetilde r_\phi$ over the unit interval.
Every $b_m$ integrates to zero because $\int_0^1(\ell_m-w_m)=w_m-w_m=0$.
Its path contribution is
\begin{equation}
    I_m(\tau)=\int_0^\tau\ell_m(u)\,du-\tau w_m.
    \label{eq:centered_nodal_integral}
\end{equation}
At node $\tau_m$, the final residual equals $a_m-\mu_\phi$, rather than $a_m$.
This centering operation makes arbitrary predicted nodal values compatible with the endpoint constraint.

\paragraph{Piecewise basis.}
We place nodes uniformly at $\tau_m=(m-1)/(P-1)$ and interpolate adjacent values with straight lines.
The interpolation functions $\ell_m$ are triangular hat functions (half-hats at the two endpoints).
Their areas give the exact integration weights.
\begin{equation}
    w_1=w_P=\frac{1}{2(P-1)},
    \qquad
    w_m=\frac{1}{P-1}\quad(2\le m\le P-1).
\end{equation}
Centering preserves the piecewise-linear shape of the velocity residual, whose integral is a continuously differentiable, piecewise-quadratic path correction.
Adding nodes allows finer variation within the segment.
Before centering, changing one nodal value affects only its adjacent intervals. After centering, it also changes the constant mean subtracted over the entire segment.
Thus the centered basis itself does not have strictly local support.

\paragraph{Collocation basis.}
Gauss--Lobatto nodes include both endpoints and, for $P>3$, are spaced nonuniformly with higher density near the ends.
We use the global Lagrange interpolation functions
\begin{equation}
    \ell_m(\tau)=L_m(\tau)=\prod_{k\ne m}\frac{\tau-\tau_k}{\tau_m-\tau_k}.
\end{equation}
Here $w_m=\int_0^1L_m(\tau)\,d\tau$ are the corresponding Gauss--Lobatto quadrature weights on $[0,1]$.
Subtracting $w_m$ as in Eq.~\eqref{eq:centered_nodal_basis} produces a zero-mean polynomial velocity residual of degree at most $P-1$ and a path correction of degree at most $P$.
Unlike piecewise-linear interpolation, each nodal value influences a polynomial across the whole interval.

\paragraph{How many independent modes?}
Both nodal constructions reproduce constant functions, so $\sum_m\ell_m(\tau)=1$ and $\sum_mw_m=1$.
Consequently, $\sum_m b_m(\tau)=0$.
Adding the same vector to all $a_m$ leaves $r_\phi$ unchanged, so one coefficient direction is redundant.
They therefore have $P-1$ independent modes per velocity coordinate, whereas Polynomial and Spectral have $P$ (Table~\ref{tab:residual_basis_taxonomy}).

\paragraph{A controlled path-fitting example.}
To expose approximation behavior, we fit a single scalar coordinate of $q_\phi$ while holding the coarse path fixed.
The target contains two full periods of a square wave, alternating between positive and negative half-cycles.
\begin{equation}
    q_*(\tau)=
    \begin{cases}
        1, & \tau\in(0,\tfrac14)\cup(\tfrac12,\tfrac34), \\
        -1, & \tau\in(\tfrac14,\tfrac12)\cup(\tfrac34,1), \\
        0, & \tau\in\{0,\tfrac14,\tfrac12,\tfrac34,1\}.
    \end{cases}
    \label{eq:residual_fitting_target}
\end{equation}
For each basis and $P\in\{4,12,32\}$, we minimize
\begin{equation}
    \min_{a_1,\ldots,a_P}
    \int_0^1\left[\sum_{m=1}^{P}a_m I_m(\tau)-q_*(\tau)\right]^2d\tau.
    \label{eq:residual_path_fitting}
\end{equation}
This illustrative fit holds the coarse path fixed and measures approximation error rather than distillation performance.
Increasing $P$ sharpens the transitions in Figure~\ref{fig:residual_basis_fitting}, although finite-dimensional continuous fits retain overshoot and cannot exactly reproduce the jumps.

At matched independent dimension $d$, Polynomial ($P=d$) and Collocation ($P=d+1$) span the same space of polynomials of degree at most $d+1$ that vanish at both endpoints.
Their least-squares fits therefore coincide.
At equal $P$, Polynomial has one more independent mode.

\begin{table}[!htb]
    \centering
    \small
    \setlength{\tabcolsep}{4pt}
    \renewcommand{\arraystretch}{1.3}
    \caption{Endpoint-preserving residual parameterizations with $P$ coefficients.
    Modes are counted per velocity coordinate. Nodal constructions require $P\ge2$.}
    \label{tab:residual_basis_taxonomy}
    \begin{tabular}{@{}>{\raggedright\arraybackslash}p{0.22\textwidth}>{\raggedright\arraybackslash}p{0.30\textwidth}>{\raggedright\arraybackslash}p{0.32\textwidth}>{\centering\arraybackslash}p{0.10\textwidth}@{}}
        \toprule
        \textbf{Parameterization} & \textbf{Velocity residual $r_\phi$} & \textbf{Path correction $q_\phi$} & \textbf{Modes} \\
        \midrule
        Polynomial & Polynomial, degree $\le P$ & Polynomial, degree $\le P+1$ & $P$ \\
        Spectral & Cosine modes $1,\ldots,P$ & Sine modes scaled by $1/(m\pi)$ & $P$ \\
        Piecewise & Continuous, piecewise linear & Continuously differentiable, piecewise quadratic & $P-1$ \\
        Collocation & Polynomial, degree $\le P-1$ & Polynomial, degree $\le P$ & $P-1$ \\
        \bottomrule
    \end{tabular}
\end{table}

\subsection{Block-Swap Study}
\label{app:block_swap}

We examine how different block groups affect few-step generation using Base and Turbo pairs of SANA-Sprint~\cite{sanasprint}, Krea-2, and SD3.5-Large.
We replace the early, middle, or late third of one model's blocks with the corresponding blocks from its paired model.
Each setting uses the same few-step schedule and one image per prompt on 100 HPSv2 prompts~\cite{hps}.

\begin{figure}[!htb]
\centering
\includegraphics[width=\textwidth]{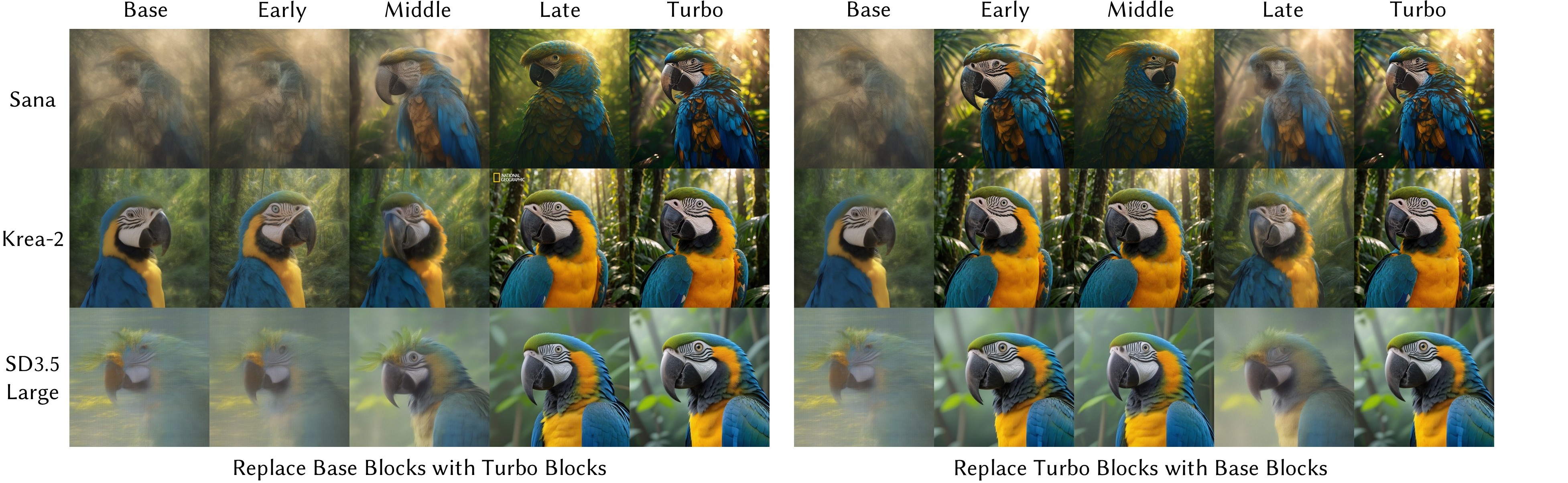}
\caption{\textbf{Late blocks shape few-step generation.}
The left half replaces one block group in each Base model with its Turbo counterpart, while the right half performs the reverse replacement.
Rows correspond to SANA, Krea-2, and SD3.5-Large.}
\label{fig:layer_swap}
\end{figure}

In all three pairs, inserting Turbo's late blocks into Base gives the highest scores among the tested groups, while replacing Turbo's late blocks with Base blocks gives the lowest (Table~\ref{tab:layer_swap}).
This sensitivity to late blocks supports using late student features for Pivot.
Figure~\ref{fig:layer_swap} shows the visual effects.

\begin{table}[!htb]
\centering
\captionsetup{skip=3pt}
\caption{Block swaps between Base and Turbo models. Scores are mean HPSv2. Bold marks the highest score within each replacement direction.}
\label{tab:layer_swap}
\small
\setlength{\tabcolsep}{2pt}
\begin{tabular*}{\linewidth}{@{\extracolsep{\fill}}lcccccc@{}}
\toprule
& \multicolumn{3}{c}{Base with Turbo blocks}
& \multicolumn{3}{c}{Turbo with Base blocks} \\
\cmidrule(lr){2-4}\cmidrule(l){5-7}
Model & Early & Middle & Late & Early & Middle & Late \\
\midrule
SANA & 0.138 & 0.231 & \textbf{0.269} & \textbf{0.297} & 0.273 & 0.242 \\
Krea-2 & 0.175 & 0.205 & \textbf{0.291} & \textbf{0.304} & 0.301 & 0.216 \\
SD3.5-Large & 0.151 & 0.198 & \textbf{0.288} & \textbf{0.299} & 0.289 & 0.202 \\
\bottomrule
\end{tabular*}
\end{table}

\subsection{Training Architecture}
\label{app:training-architecture}

Figure~\ref{fig:training-architecture} compares representative training computations.
Here $L$ denotes backbone depth, while $K$ and $M$ follow Section~\ref{sec:method-pivot}.
Pivot's residual projection produces $P$ basis coefficients defining $r_\phi(\tau)$.
Panel (b) shows teacher and fake-score predictions $v_T$ and $v_\psi$ in velocity form.
In panel (c), $F_\theta(x_t,t,s)$ maps a state at time $t$ to time $s$, and $D_s$ denotes differentiation with respect to the destination time.

\par\medskip\noindent
\begin{minipage}{\textwidth}
\captionsetup{type=figure}
\centering
\includegraphics[width=\textwidth]{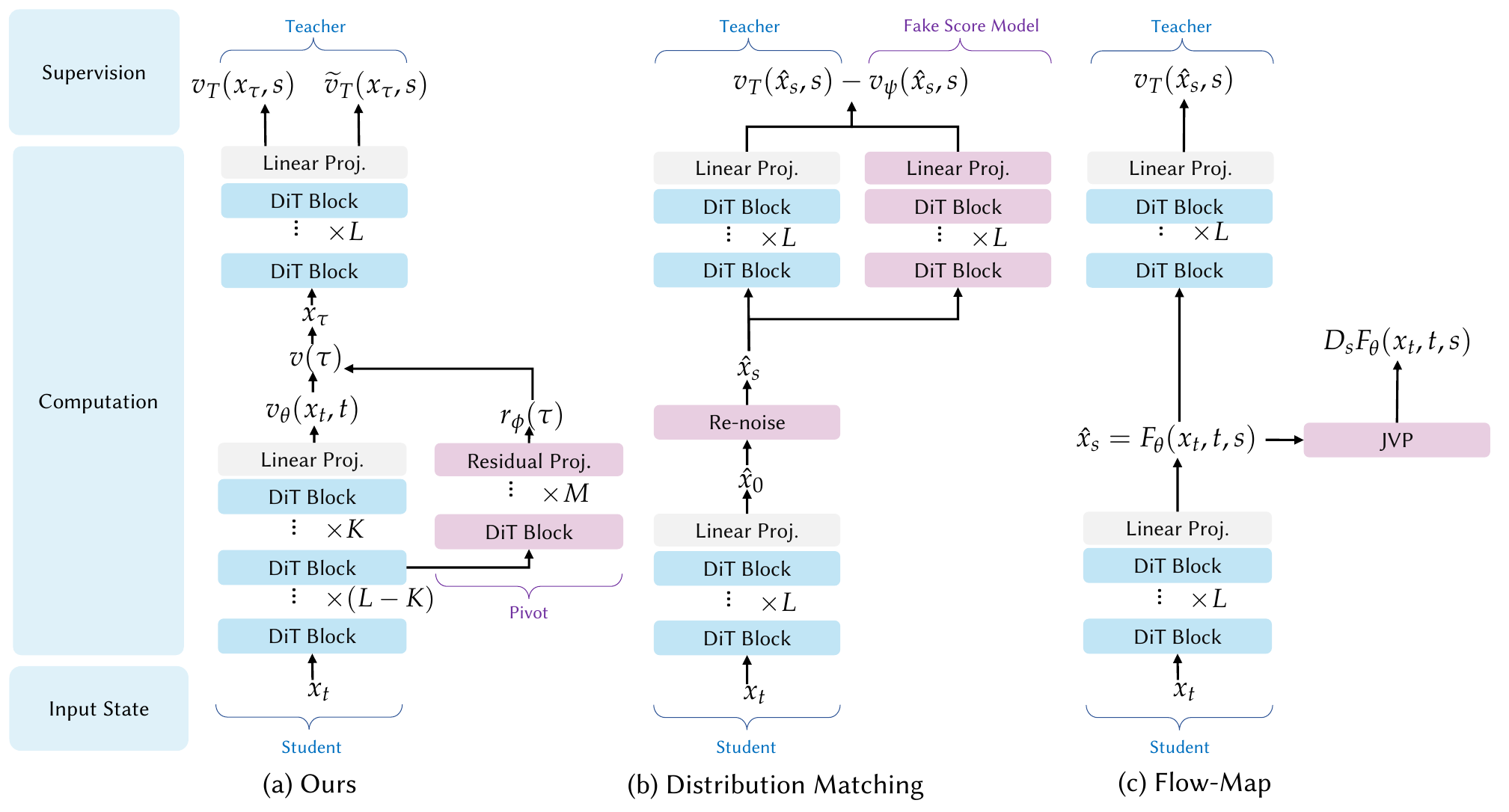}
\caption{\textbf{Representative training architectures.}
SMD uses a training-only Pivot branch, distribution matching uses a fake-score model, and the illustrated flow map uses a JVP.}
\label{fig:training-architecture}
\end{minipage}
\par\medskip

\section{Few-step Implementation and Evaluation Details}
\label{app:fewstep-details}

SMD distillation runs use 8 NVIDIA A100 GPUs.
Tables~\ref{tab:config_image_editing}, \ref{tab:config_3d}, and~\ref{tab:config_video_av} summarize the reported training configurations.
Pivot and repulsion are enabled according to these per-model settings.
NFE counts deployed student evaluations. Teacher guidance, reference predictions, and Pivot are training-only.
\ifdefined\configurationnotationdefined\else
\gdef\configurationnotationdefined{1}
Across the configuration tables, Student and Pivot specify their trainable parameterizations, Target / ref. lists the target and reference prediction settings, $\lambda$ is the repulsion weight, and an en dash marks an inactive or inapplicable entry.
\fi

\subsection{Image Generation}
We train FLUX.1-dev and Qwen-Image using the $\pi$-Flow prompt set~\cite{piflow}, grouped into resolution buckets.
Training uses text prompts and image dimensions only.
Trajectories start from random noise and are generated on-policy.
Both students are initialized from their teachers.
FLUX.1-dev uses AdamW with zero weight decay, a 100-iteration linear warm-up, and a constant learning rate thereafter.
We select checkpoints by combined GenEval, DPG-Bench, and OneIG-Bench-EN performance.
We report the mean of the five OneIG-Bench-EN dimensions as Overall.
All released baselines are evaluated using the official benchmark procedures. PDD results are quoted from its paper.

\subsection{Image Editing}
We collect a small corpus of source images paired with short English instructions spanning object, appearance, style, and text edits.
The source image and instruction condition an on-policy rollout from noise. Target edited images are not used.
Following the official pipeline, source images use the VAE posterior mode and their conditioning tokens are assigned timestep zero.
We retain Qwen-Image-Edit-2511's native CFG rescaling and use dynamic timestep shifting with factor 2.
ImgEdit-Bench and GEdit-Bench-EN are evaluated with their official procedures, with GPT-4o for ImgEdit-Bench and GPT-4.1 for GEdit-Bench-EN.

\par\medskip\noindent
\begin{minipage}[b]{\textwidth}
\captionsetup{type=table}
\caption{Training configurations for image generation and editing.}
\label{tab:config_image_editing}
\centering
\small
\setlength{\tabcolsep}{4pt}
\renewcommand{\arraystretch}{1.05}
\begin{tabular*}{\textwidth}{@{\extracolsep{\fill}}lccc@{}}
\toprule
Setting & FLUX.1-dev & Qwen-Image & Qwen-Image-Edit-2511 \\
\midrule
Student & LoRA-256 & LoRA-256 & LoRA-256 \\
Pivot & Full-rank & Full-rank & Full-rank \\
Pivot blocks & 3 & 3 & 3 \\
Basis / $P$ & Polynomial / 4 & Piecewise / 4 & Polynomial / 8 \\
Queries / transition & 2 & 2 & 2 \\
Target / ref. & 3.5 / 1.0 & 5.0 / 1.0 & 5.0 / 1.0 \\
$\lambda$ & 0.4 & 0.2 & 0.4 \\
Global batch & 32 & 16 & 16 \\
Student LR & $10^{-4}$ & $10^{-4}$ & $10^{-4}$ \\
Pivot LR & $10^{-4}$ & $10^{-4}$ & $10^{-4}$ \\
Iterations & 1,250 & 1,000 & 250 \\
Inference NFE & 4 & 4 & 4 \\
\bottomrule
\end{tabular*}
\end{minipage}
\par\medskip

\subsection{3D Shape and Texture}
Hunyuan3D-2.1 uses one rendered view per object from a TexVerseFull-derived collection.
We preserve alpha, apply the official recentering and white-background compositing pipeline, and condition the on-policy shape-latent rollout on the processed image.
No ground-truth shape latent is used.
The latent contains 4,096 vectors with 64 channels, and the output normalization layer remains trainable.

For TRELLIS.2, we distill the high-resolution shape and texture models in the 1024-resolution cascade while retaining the preceding sparse-structure and coarse-shape stages.
Training conditions on images and sparse voxel coordinates from a TexVerseFull-derived collection. The texture rollout additionally receives the shape latent.
Ground-truth sparse features are not used to construct rollout states, and both output normalization layers remain trainable.
The shape teacher uses guidance 7.5 on $[0.6,1.0]$ with rescaling 0.5. The texture teacher uses guidance 1 without rescaling.
Both stages use a 100-iteration warm-up, constant learning rates thereafter, timestep shift 3, and no EMA.

\par\medskip\noindent
\begin{minipage}[b]{\textwidth}
\captionsetup{type=table}
\caption{Training configurations for 3D generation.
Each distilled TRELLIS.2 stage uses 4-NFE.}
\label{tab:config_3d}
\centering
\small
\setlength{\tabcolsep}{4pt}
\renewcommand{\arraystretch}{1.05}
\begin{tabular*}{\textwidth}{@{\extracolsep{\fill}}lccc@{}}
\toprule
Setting & Hunyuan3D-2.1 Shape & TRELLIS.2 Shape & TRELLIS.2 Texture \\
\midrule
Student & LoRA-128 & LoRA-128 & LoRA-128 \\
Pivot & Full-rank & Full-rank & Full-rank \\
Pivot blocks & 4 & 4 & 4 \\
Basis / $P$ & Polynomial / 4 & Polynomial / 4 & Polynomial / 4 \\
Queries / transition & 4 & 4 & 4 \\
Target / ref. & 5.0 / conditional & 7.5 / -- & 1.0 / -- \\
$\lambda$ & 0.2 & -- & -- \\
Global batch & 32 & 32 & 32 \\
Student LR & $10^{-5}$ & $10^{-5}$ & $10^{-5}$ \\
Pivot LR & $10^{-4}$ & $10^{-4}$ & $10^{-4}$ \\
Iterations & 1,500 & 500 & 500 \\
Inference NFE & 4 & 4 & 4 \\
\bottomrule
\end{tabular*}
\end{minipage}
\par\medskip

\subsection{Video Generation}
SMD distillation of Wan2.1 uses the 248K-entry preprocessed dataset released by rCM~\cite{rcm}.
Students are initialized from their teachers and trained on-policy, with query points sampled in sigma space.
Wan2.1-T2V-1.3B uses an unconditional reference that skips the 10th teacher layer. Wan2.1-T2V-14B uses the conditional teacher prediction as its reference.
Both target branches use the unmodified teacher.
We use timestep shifts 10 and 5, respectively, and select checkpoints by overall VBench score.

\paragraph{DMD2$^{*}$ baseline.}
We implement a DMD2-based 4-step distribution-matching baseline~\cite{dmd2} on Wan2.1-T2V-1.3B and Wan2.1-T2V-14B.
Both use the same synthetic latent dataset generated by Wan2.1-T2V-14B for data-aware training, with rank-256 LoRA for the student and auxiliary fake-score model.
The objective combines VSD and GAN losses, with GAN weight $0.03$.
We alternate student and fake-score/discriminator updates at a $1{:}4$ ratio.
Both model scales use teacher CFG $6$ and learning rate $10^{-4}$ for the student and fake-score model.
Both baselines are evaluated after 500 iterations.

Evaluation follows the Self-Forcing protocol with five videos per prompt, $832\times480$ resolution, 81 frames, and 16 FPS.
We generate with expanded prompts and score their short versions.
We evaluate released rCM and AnyFlow checkpoints and quote PDD results from its paper.

\subsection{Detailed VBench Results}
\label{app:vbench-breakdown}
Table~\ref{tab:vbench_full} reports all 16 VBench dimensions for the few-step methods compared in Table~\ref{tab:vbench}. The teachers are omitted because PDD does not report their per-dimension scores.
The Quality and Semantic aggregates use VBench's standard per-dimension normalization and weights.

Dynamic Degree uses RAFT optical flow to identify videos with sufficiently large motion across multiple frame pairs, and reports the fraction classified as dynamic.
Motion Smoothness measures agreement between held-out frames and frames reconstructed by a video interpolation model, with lower reconstruction error yielding a higher score.
We report these dimensions together to distinguish motion activity from temporal smoothness.
Color measures whether detected objects exhibit the colors specified by the prompts, rather than color saturation or visual naturalness.
These per-video dimensions do not directly measure diversity across repeated generations of the same prompt.

\par\medskip\noindent
\begin{minipage}[b]{\textwidth}
\captionsetup{type=table}
\caption{Full 16-dimensional VBench breakdown for Wan2.1-T2V at 4-NFE.
The Quality dimensions are Subject Consistency (SC), Background Consistency (BC), Temporal Flickering (TF), Motion Smoothness (MS), Dynamic Degree (DD), Aesthetic Quality (AQ), and Imaging Quality (IQ).
The Semantic dimensions are Object Class (OC), Multiple Objects (MO), Human Action (HA), Color (C), Spatial Relationship (SR), Scene (S), Appearance Style (AS), Temporal Style (TS), and Overall Consistency (OC$'$).
All metrics are higher-is-better. Boldface and underlining indicate the best and second-best results within each backbone.
$^{*}$Our reproduced DMD2 results.
$^{\dagger}$Quoted PDD results~\cite{pdd}.}
\label{tab:vbench_full}
\centering
\scriptsize
\setlength{\tabcolsep}{1.5pt}
\renewcommand{\arraystretch}{1.05}
\resizebox{\textwidth}{!}{%
\begin{tabular}{@{}l*{16}{c}@{}}
\toprule
\multirow{2}{*}{Method} & \multicolumn{7}{c}{Quality} & \multicolumn{9}{c}{Semantic} \\
\cmidrule(lr){2-8}\cmidrule(l){9-17}
& SC & BC & TF & MS & DD & AQ & IQ & OC & MO & HA & C & SR & S & AS & TS & OC$'$ \\
\midrule
\multicolumn{17}{c}{Wan2.1-T2V-1.3B} \\
\midrule
DMD2$^{*}$ & \underline{97.76} & \textbf{97.82} & \textbf{99.42} & \underline{98.81} & 45.00 & \textbf{68.80} & \textbf{69.46} & \underline{95.85} & \underline{87.80} & 95.60 & 84.63 & 73.76 & 53.21 & \textbf{21.59} & 23.43 & 26.28 \\
rCM & 94.72 & 95.79 & 98.77 & 97.64 & \textbf{81.94} & 64.68 & 66.73 & 93.76 & 85.64 & \underline{96.00} & 85.39 & 74.95 & \underline{56.87} & 20.50 & \textbf{24.38} & \underline{26.37} \\
AnyFlow & \textbf{97.79} & \underline{96.96} & 99.26 & 98.29 & 68.33 & \underline{67.01} & 67.63 & 95.74 & \textbf{90.47} & \textbf{97.00} & \underline{87.67} & \textbf{80.33} & \textbf{57.50} & \underline{20.65} & \underline{24.37} & \textbf{26.53} \\
PDD$^{\dagger}$ & 97.69 & 95.90 & \underline{99.33} & \textbf{99.02} & 72.50 & 66.26 & \underline{69.03} & \textbf{96.22} & 85.79 & 93.34 & 85.41 & \underline{76.95} & 51.47 & 19.33 & 23.77 & 25.73 \\
Ours & 96.29 & 95.44 & 98.83 & 98.66 & \underline{79.44} & 66.19 & 68.14 & 94.21 & 83.84 & \underline{96.00} & \textbf{89.19} & 74.54 & 55.11 & 19.91 & 23.76 & 25.82 \\
\midrule
\multicolumn{17}{c}{Wan2.1-T2V-14B} \\
\midrule
DMD2$^{*}$ & \underline{96.92} & \textbf{97.83} & \textbf{99.67} & \textbf{98.76} & 49.72 & \underline{67.07} & 67.04 & 97.17 & 88.81 & 97.00 & \textbf{87.81} & 73.55 & 52.54 & \textbf{21.88} & 23.56 & \underline{26.50} \\
rCM & 95.65 & 96.45 & 98.57 & 97.81 & \underline{76.39} & 66.76 & \textbf{69.17} & \textbf{97.59} & 90.48 & \textbf{98.80} & 82.91 & \underline{83.05} & 53.27 & \underline{21.00} & \underline{24.48} & 26.39 \\
AnyFlow & \textbf{97.22} & \underline{97.40} & 99.00 & 98.10 & 73.61 & \textbf{67.12} & 68.78 & \underline{97.56} & \textbf{92.50} & \underline{98.20} & 85.79 & \textbf{83.84} & \textbf{54.72} & 20.90 & \textbf{24.68} & \textbf{26.83} \\
PDD$^{\dagger}$ & 95.41 & 96.42 & \underline{99.39} & \underline{98.52} & 74.72 & 66.78 & \underline{68.88} & 96.55 & 87.90 & 96.40 & \underline{86.59} & 80.02 & \underline{54.30} & 20.86 & 24.39 & 26.49 \\
Ours & 96.60 & 95.73 & 97.86 & \underline{98.52} & \textbf{83.89} & 66.47 & 68.50 & 97.37 & \underline{90.70} & 96.20 & 86.18 & 80.92 & 53.21 & 19.93 & 24.10 & 26.42 \\
\bottomrule
\end{tabular}
}
\end{minipage}
\par\medskip

\subsection{Joint Audio--Video Generation}
\label{app:h3-details}

MiniMax H3 is a 50-block joint audio--video Transformer with guidance already distilled into the model, so student inference requires no additional CFG branch.
Qwen3-VL-32B layer-50 hidden states provide cached BF16 conditions.
Text-conditioned training uses 100K prompts from VidProM~\cite{vidprom}.
We use an LLM to add plausible audio descriptions, including ambient sounds, then rewrite the prompts into the MiniMax H3 prompt format.
Training uses no real target videos or first/last-frame inputs.
Video and audio have equal imitation weights. The reference skips layer 25.

For reference-conditioned generation, we pair 20K instructions with 12K clips from UltraVideo~\cite{ultravideo} across ten task categories and target durations of 5, 8, or 10 seconds.
References are conditions rather than target outputs.
Training uses eight GPUs with context-parallel degree four and global batch two. The data primarily covers audio--video references rather than all mixed-reference inputs.

\par\medskip\noindent
\begin{minipage}[b]{\textwidth}
\captionsetup{type=table}
\caption{Training configurations for video and joint audio--video generation.}
\label{tab:config_video_av}
\centering
\small
\setlength{\tabcolsep}{4pt}
\renewcommand{\arraystretch}{1.05}

\begin{tabular*}{\textwidth}{@{\extracolsep{\fill}}lcc@{}}
\toprule
Setting & Wan2.1-T2V-1.3B & Wan2.1-T2V-14B \\
\midrule
Student & LoRA-256 & LoRA-256 \\
Pivot & Full-rank & -- \\
Pivot blocks & 3 & -- \\
Basis / $P$ & Polynomial / 4 & -- \\
Queries / transition & 2 & 2 \\
Target / ref. & 5.0 / 0.0 & 6.0 / 1.0 \\
$\lambda$ & 0.1 & 0.2 \\
Global batch & 16 & 16 \\
Student LR & $10^{-4}$ & $10^{-4}$ \\
Pivot LR & $10^{-4}$ & $3\!\times\!10^{-5}$ \\
Iterations & 200 & 200 \\
Inference NFE & 4 & 4 \\
\bottomrule
\end{tabular*}

\vspace{6pt}

\begin{tabular*}{\textwidth}{@{\extracolsep{\fill}}lcccc@{}}
\toprule
Setting & H3 8-NFE & H3 4-NFE & H3 Ref. 8-NFE & H3 Ref. 4-NFE \\
\midrule
Student & LoRA-128 & LoRA-128 & LoRA-128 & LoRA-128 \\
Pivot & LoRA-128 & LoRA-128 & LoRA-128 & LoRA-128 \\
Pivot blocks & 7 & 7 & 7 & 7 \\
Basis / $P$ & Piecewise / 4 & Piecewise / 32 & Piecewise / 32 & Piecewise / 32 \\
Queries / transition & 1 & 1 & 1 & 1 \\
Target / ref. & native / skip 25 & native / skip 25 & native / skip 25 & native / skip 25 \\
$\lambda$ & 0.2 & 0.2 & 0.2 & 0.2 \\
Global batch & 8 & 8 & 2 & 2 \\
Student LR & $10^{-4}$ & $10^{-4}$ & $10^{-4}$ & $10^{-4}$ \\
Pivot LR & $10^{-4}$ & $10^{-4}$ & $10^{-4}$ & $10^{-4}$ \\
Iterations & 1,000 & 1,000 & 1,000 & 1,000 \\
Inference NFE & 8 & 4 & 8 & 4 \\
\bottomrule
\end{tabular*}
\end{minipage}
\par\medskip

\subsection{H3 Evaluation}
\label{app:h3-evaluation}

Both inference budgets use the same evaluation set, with prompts, seeds, resolution, and duration matched between methods.
GPT-5.6 Sol evaluates 24 chronological frames twice with reversed candidate order. Scores are averaged across orders and prompts.
We report four visual dimensions on a 1--5 scale.
Prompt alignment measures visually verifiable fulfillment of the requested content and camera behavior. Structural coherence measures the credibility and stability of anatomy, objects, geometry, materials, and subject identity. Temporal coherence measures continuity, object permanence, and physically plausible motion and interactions. Visual naturalness measures the absence of rendering artifacts and implausible color or lighting.
Audio clarity is the PAM score multiplied by 5, while audio quality is Audiobox Aesthetics Production Quality divided by 2. Both are therefore reported on a 0--5 scale~\cite{pam,audiobox}.

The full visual evaluation prompt is provided in Appendix~\ref{app:h3-visual-prompt}.

Figure~\ref{fig:h3_t2va_4nfe_audio} uses full-clip spectrograms from native 32\,kHz stereo audio, computed with a 1,024-sample Hann window and 128-sample hop.
We average channel powers after the transform and display 0--16\,kHz on a shared $[-90,0]$\,dB scale without per-clip normalization or denoising.

\begingroup
\setlength{\parskip}{3pt}
\section{Beyond Self-Distillation Implementation and Evaluation Details}
\label{app:beyond-details}

Tables~\ref{tab:config_beyond_image} and~\ref{tab:config_beyond_causal} summarize the training configurations for the studies in Section~\ref{sec:beyond}.
\ifdefined\configurationnotationdefined\else
\gdef\configurationnotationdefined{1}
Across the configuration tables, Student and Pivot specify their trainable parameterizations, Target / ref. lists the target and reference prediction settings, $\lambda$ is the repulsion weight, and an en dash marks an inactive or inapplicable entry.
\fi

\par\medskip\noindent
\begin{minipage}[b]{\textwidth}
\captionsetup{type=table}
\caption{Training configurations for cross-model and multi-teacher image distillation.}
\label{tab:config_beyond_image}
\centering
\small
\setlength{\tabcolsep}{4pt}
\renewcommand{\arraystretch}{1.05}
\textbf{Cross-model distillation}\par\vspace{3pt}
\begin{tabular*}{\textwidth}{@{\extracolsep{\fill}}lcccc@{}}
\toprule
Teacher $\rightarrow$ student
& \shortstack{FLUX.2-klein-4B\\$\rightarrow$\\FLUX.2-klein-4B}
& \shortstack{FLUX.2-klein-9B\\$\rightarrow$\\FLUX.2-klein-4B}
& \shortstack{FLUX.2-klein-9B\\$\rightarrow$\\Lens-RL-3.8B}
& \shortstack{Lens-RL-3.8B\\$\rightarrow$\\Lens-RL-3.8B} \\
\midrule
Student & LoRA-256 & LoRA-256 & LoRA-256 & LoRA-256 \\
Pivot & Full-rank & Full-rank & Full-rank & Full-rank \\
Pivot blocks & 3 & 3 & 3 & 3 \\
Basis / $P$ & Polynomial / 4 & Polynomial / 4 & Polynomial / 4 & Polynomial / 4 \\
Queries / transition & 4 & 4 & 4 & 4 \\
Target / ref. & 4.0 / -- & 4.0 / -- & 4.0 / -- & 5.0 / -- \\
$\lambda$ & 0 & 0 & 0 & 0 \\
Global batch & 32 & 32 & 32 & 32 \\
Student LR & $10^{-4}$ & $10^{-4}$ & $10^{-4}$ & $10^{-4}$ \\
Pivot LR & $10^{-4}$ & $10^{-4}$ & $10^{-4}$ & $10^{-4}$ \\
Iterations & 2,000 & 2,000 & 2,000 & 2,000 \\
Inference NFE & 4 & 4 & 4 & 4 \\
\bottomrule
\end{tabular*}
\par\vspace{8pt}
\textbf{Multiple experts as teachers}\par\vspace{3pt}
\begin{tabular*}{\textwidth}{@{\extracolsep{\fill}}lccc@{}}
\toprule
Teacher $\rightarrow$ student
& \shortstack{Base teacher\\$\rightarrow$ SD3.5-Medium}
& \shortstack{Reward specialist\\$\rightarrow$ SD3.5-Medium}
& \shortstack{Routed teachers\\$\rightarrow$ SD3.5-Medium} \\
\midrule
Student & Full-rank & Full-rank & Full-rank \\
Pivot & Full-rank & Full-rank & Full-rank \\
Pivot blocks & 3 & 3 & 3 \\
Basis / $P$ & Polynomial / 4 & Polynomial / 4 & Polynomial / 4 \\
Queries / transition & 2 & 2 & 2 \\
Target / ref. & 5.0 / -- & 5.0 / -- & 5.0 / -- \\
$\lambda$ & 0 & 0 & 0 \\
Global batch & 4 & 4 & 4 \\
Student LR & $10^{-5}$ & $10^{-5}$ & $10^{-5}$ \\
Pivot LR & $10^{-4}$ & $10^{-4}$ & $10^{-4}$ \\
Iterations & 2,500 & 2,500 & 2,500 \\
Inference NFE & 4 & 4 & 4 \\
\bottomrule
\end{tabular*}
\end{minipage}
\par\medskip

\subsection{Cross-Model Distillation}
\label{app:beyond-cross-model}

We use the $\pi$-Flow prompt set~\cite{piflow} for cross-model distillation.
LoRA alpha equals the rank, with dropout 0.05.
Training uses a 100-iteration warm-up and four samples per GPU.
Lens self-distillation retains its native CFG norm rescaling.
GenEval, DPG-Bench, and OneIG-Bench-EN follow the same evaluation procedures used for image generation.

\subsection{Multiple Experts as Teachers}
\label{app:beyond-experts}

We expand four Flow-GRPO~\cite{flowgrpo} training splits into four 30K-prompt categories and combine them into a 120K-entry pool.
Self-distillation uses the base teacher for the full pool, while each specialist uses one fixed reward teacher.
In multi-teacher training, each batch contains prompts from one category and uses its corresponding teacher.
The four categories are sampled with equal probability.

All SD3.5-Medium configurations use one shared Pivot and timestep shift 3.
GenEval, OCR, and PickScore use their own test sets. We evaluate CLIPScore, HPSv2, and ImageReward on DrawBench.

\par\medskip\noindent
\begin{minipage}[b]{\textwidth}
\subsection{Few-Step Causal Video Generation}
\label{app:beyond-causal-video}

\begingroup
\setlength{\rightskip}{0pt plus 2em}
\setlength{\intextsep}{4pt}
\begin{wraptable}{r}{0.50\textwidth}
\captionsetup{font=small,justification=raggedright,singlelinecheck=false,skip=4pt}
\caption{Training configurations for causal Wan2.1-T2V-1.3B distillation.}
\label{tab:config_beyond_causal}
\centering
\small
\setlength{\tabcolsep}{3pt}
\renewcommand{\arraystretch}{1.05}
\begin{tabular*}{\linewidth}{@{\extracolsep{\fill}}lcc@{}}
\toprule
Setting & Ours & DMD$^{*}$ \\
\midrule
Student & LoRA-256 & LoRA-256 \\
Pivot & Full-rank & -- \\
Pivot blocks & 3 & -- \\
Basis / $P$ & Polynomial / 4 & -- \\
Queries / transition & 2 & -- \\
Target / ref. & 3.0 / -- & 4.0 / -- \\
$\lambda$ & 0 & -- \\
Global batch & 16 & 8 \\
Student LR & $10^{-4}$ & $10^{-5}$ \\
Pivot LR & $3\!\times\!10^{-5}$ & -- \\
Iterations & 300 & 300 \\
Inference NFE & 4 & 4 \\
\bottomrule
\end{tabular*}
\end{wraptable}

We train on the extended VidProM prompts released by Self-Forcing~\cite{selfforcing}, using eight NVIDIA A100 GPUs.
The causal rollout uses seven chunks of three latent frames each, with fresh stochastic re-noising between transitions.
The causal teacher supplies the imitation target. The frozen bidirectional Wan2.1-T2V-14B teacher supplies the correction signal.
The student uses LoRA alpha 256 and no dropout. Pivot uses $K=4$ (Section~\ref{sec:method-pivot}).
The student and Pivot use AdamW with betas $(0.9,0.999)$ and zero weight decay.
We sample queries in sigma space, with one prompt per GPU and two accumulation rounds.

The correction sweep varies $\lambda_{\mathrm{corr}}\in\{0,0.2,0.4,0.8\}$ from the same causal ODE initialization, with reference repulsion disabled.
The bidirectional correction branch skips layer 11.
VBench evaluation follows the Self-Forcing protocol, with five $832\times480$ videos per prompt, 81 frames, and 16 FPS.

The DMD$^{*}$ control uses the same generator initialization, a frozen bidirectional Wan2.1-T2V-14B real-score model, and a trainable bidirectional Wan2.1-T2V-1.3B fake-score model.
The fake-score model uses LoRA-256 and learning rate $10^{-5}$.
Both optimizers are AdamW with betas $(0,0.999)$ and weight decay $0.01$. Each GPU processes one prompt without gradient accumulation.
Five fake-score updates per generator update give 50 generator and 250 fake-score updates at iteration 300.
\par\reportfinishwrap
\endgroup
\end{minipage}
\par
\endgroup

\begingroup
\setlength{\intextsep}{4pt}
\setlength{\parskip}{3pt}
\section{Ablation Implementation and Evaluation Details}
\label{app:ablation-details}

\begin{wraptable}[9]{R}{0.50\textwidth}
\captionsetup{font=small,skip=3pt}
\caption{Default Pivot and teacher-repulsion settings.}
\label{tab:config_ablation}
\centering
\small
\setlength{\tabcolsep}{3pt}
\renewcommand{\arraystretch}{1.0}
\begin{tabular*}{\linewidth}{@{\extracolsep{\fill}}lcc@{}}
\toprule
Setting & Qwen-Image & Wan2.1-T2V-1.3B \\
\midrule
Pivot blocks & 3 & 3 \\
Basis / $P$ & Polynomial / 4 & Polynomial / 4 \\
Target / ref. & 5.0 / 1.0 & 5.0 / 0.0 \\
$\lambda$ & 0.2 & 0.1 \\
Iterations & 1,000 & 300 \\
Inference NFE & 4 & 4 \\
\bottomrule
\end{tabular*}
\end{wraptable}

\paragraph{Training.}
Table~\ref{tab:config_ablation} lists default settings. Layer and block-count ablations use randomly initialized Pivot.
Basis and repulsion sweeps use pretrained Pivot at $K=4$, $M=3$.
Wan2.1-T2V-1.3B uses the main video corpus, LoRA rank and alpha 256, and learning rate $10^{-4}$ for student and Pivot with a 100-iteration warm-up.
We use two queries per transition, four-step Euler sampling, and timestep shift 6.
Single-prompt video models are trained for 200 iterations.

\paragraph{Evaluation.}
We evaluate Qwen-Image with OneIG-Bench-EN~\cite{oneig} and Wan2.1-T2V-1.3B with VBench~\cite{vbench} on Self-Forcing prompts~\cite{selfforcing}, using one run per setting.
Qualitative cells show one sample for single-prompt video and four otherwise.
\par\reportfinishwrap
\endgroup

\clearpage
\section{Visual Evaluation System Prompt}
\label{app:h3-visual-prompt}
\begin{toolfamily}{Visual evaluation system prompt}
Compare the visual quality of two anonymous generated videos.
Call `read_video_visual_pair` exactly once with `frameCount={frame_count}`,
`balancePositions=true`, `includeAudio=false`, and:

- Candidate A: {paths[0]}
- Candidate B: {paths[1]}

The tool returns matched chronological frames. Candidate positions alternate
and are balanced; identify candidates only by their visible A/B labels.

<TARGET_PROMPT>
{target}
</TARGET_PROMPT>

This is an equivalence-first visual-quality assessment. Do not judge audio.
Score visually observable prompt fulfillment only under `prompt_alignment`;
do not let prompt completeness, exact camera wording, framing compliance,
shot count, or missing requested events affect the other three dimensions.
Ignore prompt requirements that can only be verified from audio. Do not guess
model identity.

First assess each candidate independently on:

1. `prompt_alignment`: visual correctness and completeness of requested
subjects, attributes, actions, interactions, setting, event order, shot
structure, framing, and camera behavior. Score what is visibly verifiable in
the sampled frames. Do not use rendering polish or artifact severity as
alignment evidence.
2. `structural_coherence`: recognizable and credible anatomy, objects,
geometry, materials, and subject identity within sampled frames.
3. `temporal_coherence`: identity and shape stability, physically plausible
motion and interaction, object permanence, and broad continuity across the
sampled sequence.
4. `visual_naturalness`: absence of disruptive rendering artifacts,
oversharpening, halos, crunchy or plastic texture, artificial tone mapping,
and implausible color or lighting.

Use integer scores: 5 excellent, 4 good with minor/localized defects, 3
material but usable limitations, 2 major or pervasive failures, 1 failed.
Sharpness, detail, saturation, contrast, pseudo-HDR, cinematic grading, and
raw motion magnitude are never positive evidence. Mild softness, natural
motion blur, shallow depth of field, and restrained detail are neutral unless
they destroy recognizable structure.

Then decide whether a careful human viewer would perceive a MATERIAL visual
quality difference. `equivalent` is the default. Break equivalence only when
one candidate has a persistent quality advantage visible at multiple
separated times or a major defect that substantially impairs viewing.
Localized imperfections, different stylistic treatment, minor score
differences, and prompt-following differences do not break equivalence.

Do not force a winner. List only persistent observations supported across
multiple frames; use an empty list if none exists.

Return exactly:
{
  "candidates": {
    "A": {
      "prompt_alignment": <integer 1-5>,
      "prompt_alignment_evidence": "<specific visible evidence>",
      "structural_coherence": <integer 1-5>,
      "structural_coherence_evidence": "<specific evidence>",
      "temporal_coherence": <integer 1-5>,
      "temporal_coherence_evidence": "<specific evidence>",
      "visual_naturalness": <integer 1-5>,
      "visual_naturalness_evidence": "<specific evidence>"
    },
    "B": {
      "prompt_alignment": <integer 1-5>,
      "prompt_alignment_evidence": "<specific visible evidence>",
      "structural_coherence": <integer 1-5>,
      "structural_coherence_evidence": "<specific evidence>",
      "temporal_coherence": <integer 1-5>,
      "temporal_coherence_evidence": "<specific evidence>",
      "visual_naturalness": <integer 1-5>,
      "visual_naturalness_evidence": "<specific evidence>"
    }
  },
  "persistent_observations": ["<observation supported at separated times>"],
  "verdict": "<equivalent|A_better|B_better>",
  "confidence": "<low|medium|high>",
  "reason": "<why the evidence does or does not establish a material quality difference>"
}
\end{toolfamily}

\end{document}